\pdfoutput=1

\documentclass[11pt]{article}

\usepackage[]{acl}

\usepackage{times}
\usepackage{latexsym}

\usepackage[T1]{fontenc}

\usepackage[utf8]{inputenc}

\usepackage{microtype}
\usepackage{booktabs}
\usepackage{graphics}
\usepackage{graphicx}
\usepackage{arydshln}
\usepackage{hyperref}
\usepackage{multirow}
\usepackage{footnote}
\usepackage{footmisc}
\usepackage{xcolor}
\usepackage{colortbl}
\usepackage{tabulary}
\usepackage{amsmath}
\usepackage{tablefootnote}
\usepackage{epigraph}
\usepackage{gb4e}
\noautomath
\usepackage{xcolor}
\usepackage{amssymb}
\usepackage[scientific-notation=true]{siunitx}
\usepackage{pifont}

\usepackage[russian,main=english]{babel}
\babelprovide[import]{russian}

\usepackage{CJKutf8} 

\definecolor{darkyellow}{RGB}{191,110,0}
\definecolor{darkblue}{RGB}{0,0,139}
\definecolor{darkorange}{RGB}{255,100,0}

\title{Large language models effectively leverage document-level context\\ for literary translation, but critical errors persist}

\author{Marzena Karpinska \quad Mohit Iyyer \\
        Manning College of Information and Computer Sciences \\ University of Massachusetts Amherst \\
        \texttt{\{mkarpinska, miyyer\}@cs.umass.edu}}

\begin{document}
\maketitle

\begin{abstract}
Large language models (LLMs) are competitive with the state of the art on a wide range of \emph{sentence-level} translation datasets. However, their ability to translate paragraphs and documents remains unexplored because evaluation in these settings is costly and difficult. 
We show through a rigorous human evaluation that asking the \textsc{Gpt-3.5} (\texttt{text-davinci-003}) LLM to translate an entire \emph{literary} paragraph (e.g., from a novel) at once results in higher-quality translations than standard sentence-by-sentence translation across 18 linguistically-diverse language pairs (e.g., translating into and out of Japanese, Polish, and English). Our evaluation, which took approximately 350 hours of effort for annotation and analysis, is conducted by hiring translators fluent in both the source and target language and asking them to provide both span-level error annotations as well as preference judgments of which system's translations are better. We observe that discourse-level LLM translators commit fewer mistranslations, grammar errors, and stylistic inconsistencies than sentence-level approaches. With that said, critical errors still abound, including occasional content omissions, and a human translator's intervention remains necessary to ensure that the author's voice remains intact.
We publicly release our dataset and error annotations to spur future research on evaluation of document-level literary translation.\footnote{\url{https://github.com/marzenakrp/LiteraryTranslation}}

\end{abstract}
 \section{Introduction}

\begin{figure}[ht!]
\centering
\includegraphics[width=.85 \columnwidth]{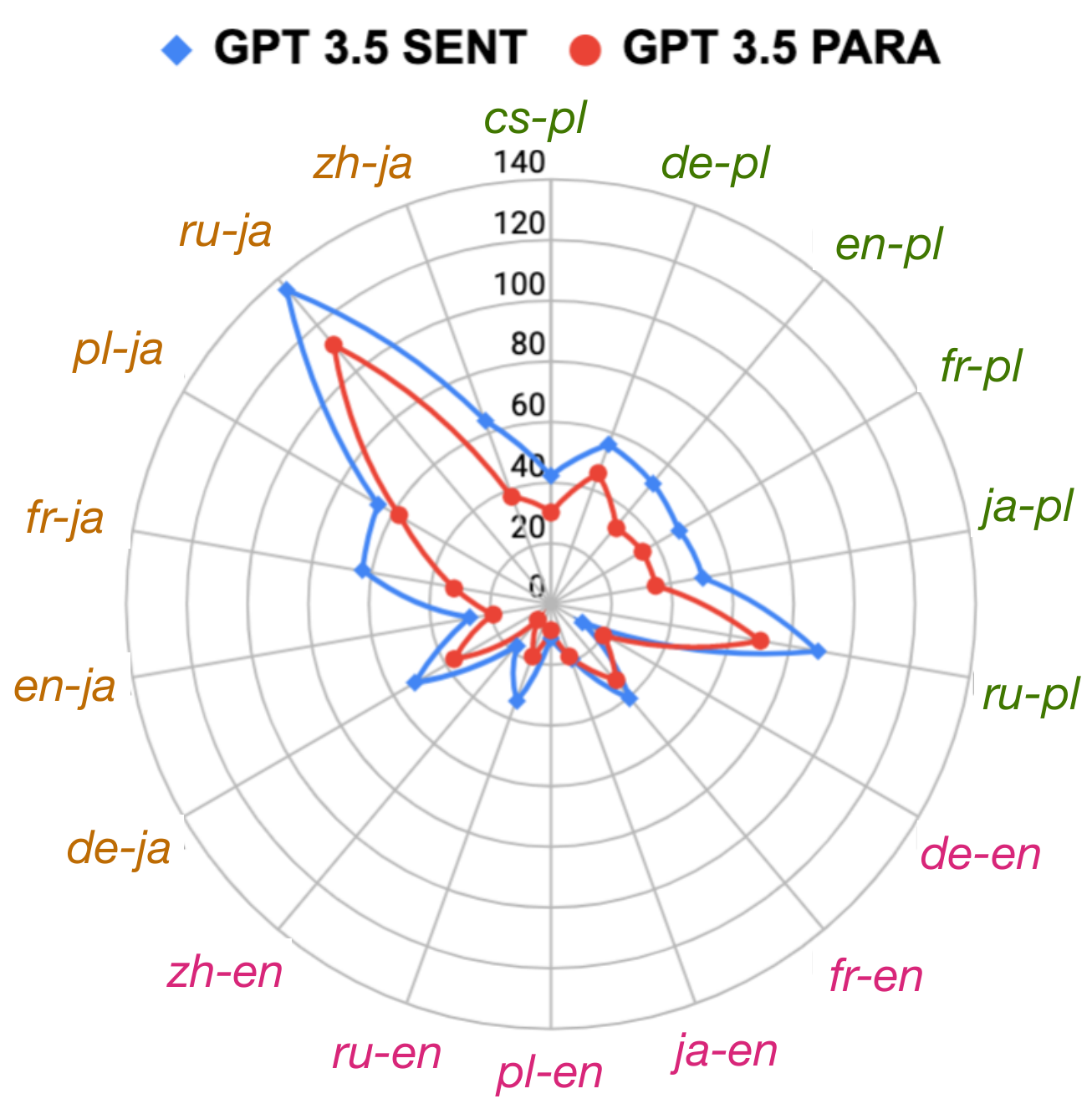}
\caption{A plot of the total number of errors annotated in sentence-level (\textsc{Sent}) and paragraph-level (\textsc{Para}) translations produced by \textsc{Gpt-3.5} across 18 different language pairs. In all cases, \textsc{Para} produces fewer errors than \textsc{Sent}, which demonstrates that \textsc{Gpt-3.5} takes advantage of discourse context during translation.
}
\label{figure:sent_vs_para_in_intro}
\end{figure}

\setlength \epigraphwidth {0.7\linewidth }
\epigraph{Separate text from context and all that remains is a con.}{\emph{Stewart Stafford}}

Large language models (LLMs) such as ChatGPT \cite{chatgpt_intro} demonstrate remarkable performance as stand-alone translation systems, rivaling and sometimes surpassing commercial models on sentence-level benchmarks~\cite{vilar2022prompting, hendy2023good, jiao2023chatgpt}. Furthermore, LLMs are increasingly being deployed for \emph{document-level} translation~\cite{bilingual_book_maker, ATA_trans_2023}, a scenario for which there are currently
no reliable automatic evaluation methods. 
In this paper, we hire human translators to conduct a rigorous fine-grained evaluation of \textsc{Gpt-3.5}'s ability  to translate \textbf{paragraph-level} texts from \textbf{literary works} across 18 different language pairs. Our results (Figure~\ref{figure:sent_vs_para_in_intro}) demonstrate that \textsc{Gpt-3.5}\footnote{We completed our annotations on translations from the \texttt{text-davinci-003} checkpoint obtained prior to the API release of ChatGPT and GPT-4. Nevertheless, we include preliminary analysis of GPT-4's translations in \S\ref{sec:limitations}.} effectively leverages discourse-level context to produce higher-quality translations than when translating sentences in isolation.

\begin{figure*}[t]
\centering
\includegraphics[width=\textwidth]{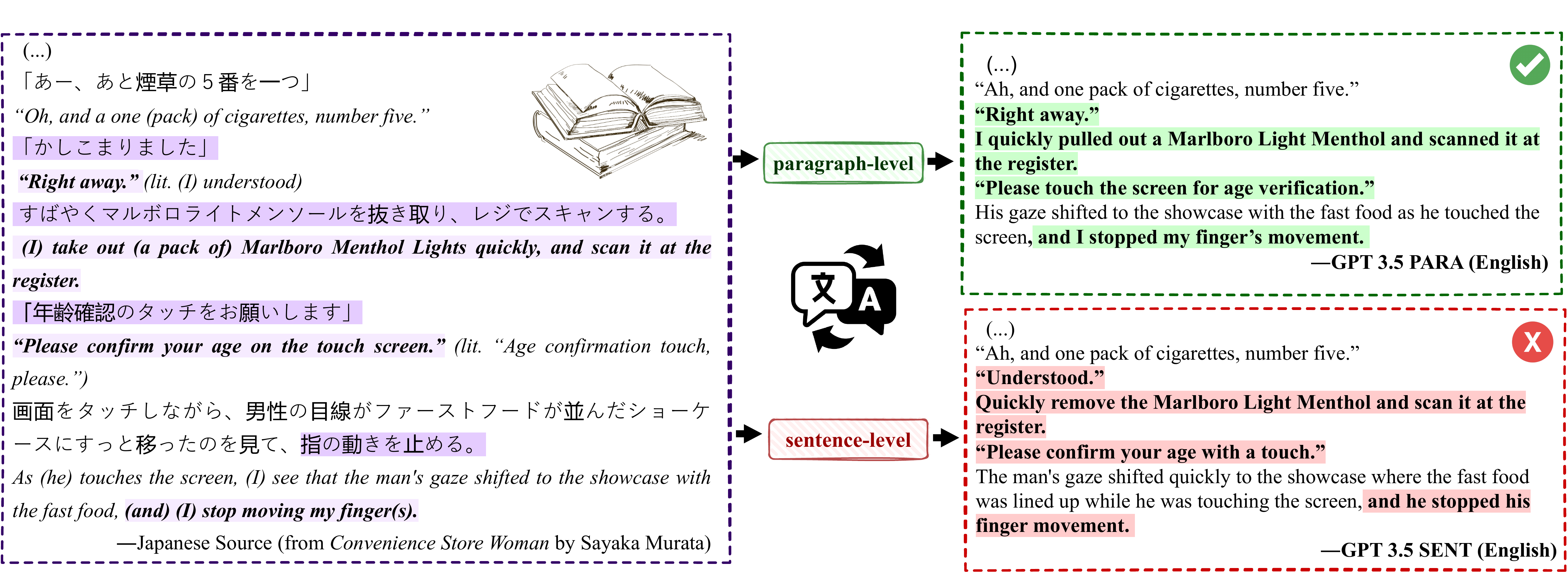}
\caption{An example of paragraph-level (\textsc{Para}) and sentence-level (\textsc{Sent}) translations of the same Japanese paragraph into English. Sentence-level translation results in a range of erroneous translations, from worse word choice (``understood'' vs ``right away'') to incorrect pronouns (``he'' vs ``I'').}
\label{figure:trans_ctx_no_ctx_example}
\end{figure*}

\paragraph{Why literary texts?}    
Translating works of literature poses unique challenges due to the intricate nature of creative work and the importance of capturing the author's voice and contextual nuances. 
Translators thus apply a wide range of translation techniques \cite{Chesterman1997-memes, Molina2004-ms}, from simple shifts in grammatical categories to more complex stylistic or content-based rearrangements that often cross sentence boundaries. Translators may also merge or split sentences, or even entire paragraphs, which renders the traditional sentence-level pipeline insufficient for capturing the full scope of the original text~\cite{Toral2015, taivalkoski2019free}.\footnote{At least 55\% of the reference target paragraphs used in our study split or merge sentences from the source text (measured with an automatic sentence tokenizer).} Taken together, these properties make literary texts a good testbed for document-level machine translation~\citep{thai-etal-2022-exploring}; in our work, we focus on the \emph{paragraph}\footnote{We broadly define a paragraph as a distinct passage within the novel, focusing on a single theme.} as a minimal discourse-level unit.

\paragraph{Why human evaluation?}
The absence of rigorous document-level evaluations of LLM translators is striking but also somewhat understandable given the unreliability of automatic metrics~\citep{thai-etal-2022-exploring} and the difficulty of properly conducting human evaluations~\cite{castilho-2021-towards}. Furthermore, evaluations of LLM translators are especially difficult due to data contamination \cite{aiyappa2023trust}, as it is unclear whether the models are pretrained on existing benchmarks (e.g., from WMT). We fill this gap by first collecting paragraphs from recently-published literary translations. Then, we provide human translators with two candidate machine translations of a given source paragraph and ask them to (1) mark error \emph{spans} and categorize them based on a predefined schema inspired by MQM~\citep{lommel2014multidimensional,10.1162/tacl_a_00437}, (2) make preference judgments of which of the two translations is of higher quality, and (3) provide free-form justifications of their preference judgments. In total, we collect such annotations on \textbf{720} pairs of translated paragraphs across \textbf{18} different language pairs (using three diverse target languages of English, Japanese, and Polish), which we then leverage for a fine-grained analysis of the behavior of different LLM translation methods.

\paragraph{How do we use LLMs to translate paragraphs?}
We use three strategies to generate the paragraph-level translations for our evaluations that all rely on few-shot prompting with \textsc{Gpt-3.5}: (1) translating each sentence in the paragraph in isolation of the others (\textsc{Sent}); (2) translating each sentence in the paragraph when provided with the rest of the paragraph as context (\textsc{Para\_Sent}); and (3) translating the entire paragraph in at once (\textsc{Para}), \textit{not} sentence-by-sentence. Finally, we also compare these methods to Google Translate (\textsc{GTr}).

\paragraph{LLMs produce better translations when provided with paragraph-level context:}
Our evaluations reveal that using \textsc{Gpt-3.5} to translate complete paragraphs (\textsc{Para}) yields translations of significantly higher quality than both the sentence-by-sentence \textsc{Gpt-3.5} methods as well as Google Translate. 
Our detailed analysis of annotated translation errors and free-form comments show that paragraph-level translations exhibit increased coherence, better preservation of literary style, and improved handling of context-dependent expressions (see \autoref{figure:trans_ctx_no_ctx_example}). That said, we also observe that \textsc{Para} still makes numerous critical mistranslations and other errors across different language pairs, which shows that LLM-based translators still have significant room for improvement, particularly when applied to translating contextually-rich literary texts. 

\section{Background}
Before describing our dataset and evaluation, we first contextualize our work within both the existing body of document-level\footnote{Note that the term ``document-level'' has been used in MT research to denote both multi-sentence passages as well as complete documents.} machine translation as well as recent papers on translation via large language models.

\subsection{Existing approaches to document-level translation}
Before the rise of neural machine translation, several attempts were made to incorporate discourse-level phenomena into statistical machine translation systems~\cite{Hardmeier2012, carpuat-simard-2012-trouble, hardmeier-etal-2013-docent, ding-etal-2014-document}. Neural MT systems condition sentence-by-sentence translation on discourse-level context via concatenation models \cite{tiedemann-scherrer-2017-neural, jean2017does, Agrawal2018ContextualHI, junczys-dowmunt-2019-microsoft, lopes-etal-2020-document}, hierarchical models \cite{miculicich-etal-2018-document, tan-etal-2019-hierarchical, chen-etal-2020-modeling, ijcai2020p551}, multi-pass models \cite{mansimov-etal-2021-capturing}, dynamic context models \cite{kang-etal-2020-dynamic}, multi-source models \cite{zhang-etal-2018-improving, feng-etal-2022-learn}, and transfer learning approaches \cite{zhang-etal-2022-multilingual}. 
Despite sometimes obtaining clear gains from discourse-level context~\cite{voita-etal-2019-good}, the machine translation community has not made much progress on this problem, particularly for non-English language pairs, due largely to the scarcity of parallel document-level corpora \cite{zhang-etal-2022-multilingual}. This problem has been partially addressed by introducing a pivot language \cite{cohn-lapata-2007-machine, utiyama-isahara-2007-comparison}, but this approach can also lead to substantial information loss.

\subsection{Translation with large language models}

Many recent studies explore the potential that LLMs hold for translation, an especially attractive prospect given that training or fine-tuning on large parallel corpora is not necessary.\footnote{That said, parallel data is almost certainly included, at least for high-resource languages, in LLM pretraining data.} These works span paragraph-level post-editing with LLMs \cite{thai-etal-2022-exploring}, translating sentence-level inputs \cite{vilar2022prompting, jiao2023chatgpt}, analyzing hallucinations in LLM-generated translations \cite{guerreiro2023hallucinations}, and employing LLMs to evaluate machine translation \cite{kocmi2023large}.
Studies on prompt engineering for translation conclude that simple sentence-level English prompt templates are effective for paragraph translations \cite{zhang2023prompting}. Other findings reveal that automatically-generated dictionaries assist translation \cite{ghazvininejad2023dictionarybased}, and that example quality outweighs lexico-semantic proximity to input \cite{vilar2022prompting}.
To the best of our knowledge, the only work other than ours that evaluates LLMs for paragraph-level translation is that of \citet{hendy2023good}, which focuses on automatic evaluation of context-aware sentence-by-sentence translation. Unlike~\citet{hendy2023good}, we perform a fine-grained \emph{human} evaluation of paragraph-level translation, which sheds more light on the concrete strengths and weaknesses of LLM translators in this setting.

\section{Data \& methods}
Our work differs from existing research on translating with large language models in two key ways: we focus on translating \emph{literary} text at the \emph{paragraph level}. In this section, we describe and motivate the paragraph-level translation dataset used in our study, which covers eighteen unique language pairs (three target languages) and is sourced from recently-published novels. Then, we outline the different ways in which we leverage \textsc{Gpt-3.5} to translate these paragraphs at both the sentence and paragraph levels.

\begin{table*}[htbp]
\centering
\resizebox{\textwidth}{!}{
\begin{tabular}{llp{4in}p{4in}}
\toprule
\textbf{Book} & \textbf{Lang Pair} & \textbf{Source} & \textbf{Target} \\ 
\midrule
An Inventory of Losses & de-ja & Natürlich hatte ich schon davor andere bemerkenswerte Begräbnisstätten besucht: die Toteninsel San Michele etwa, wie sie mit hohen, roten Backsteinmauern aus dem blaugrünen Wasser der Lagune von Venedig emporragt gleich einer uneinnehmbaren Festung, oder das grelle Jahrmarktstreiben des Hollywood Forever Cemetery am alljährlich von der mexikanischen Bevölkerung begangenen Día de los Muertos mit den orange-gelb geschmückten Gräbern und den von der fortgeschrittenen Verwesung auf ewig zum Grinsen verdammten Totenschädeln aus bunt gefärbtem Zucker und Pappmaché. Doch keine hat mich so berührt wie der Friedhof jener Fischersiedlung, in dessen eigentümlichem Grundriss — einer Art Kompromiss aus Kreis und Quadrat ich nichts anderes als ein Sinnbild der ungeheuerlichen Utopie zu erkennen glaubte, die ich dort verwirklicht sah: mit dem Tod vor Augen zu leben. Lange Zeit war ich überzeugt, an diesem Ort, dessen dänischer Name »kleine Insel« oder »vom Wasser umgeben« bedeutet, sei man dem Leben näher, gerade weil seine Bewohner die Toten wortwörtlich in ihre Mitte geholt hatten, anstatt sie wie sonst in unseren Breitengraden üblich — aus dem Innersten der Gemeinden vor die Stadttore zu verbannen, auch wenn der urbane Raum sich die Gräberstätten durch sein ungehemmtes Anwachsen oft nur wenig später wieder einverleibt hat. & \begin{CJK}{UTF8}{min}もちろんそれ以前にもいくつか特筆すべき墓所を訪れたことはあった。たとえばヴェネツィアの干潟の青緑色の水中から、赤煉瓦の高い壁に囲まれて難攻不落の要塞のようにそびえたつ死者の島、サン・ミシェル。あるいはメキシコ系住民が毎年にぎやかに死者の日を祝う、ハリウッド・フォーエバー墓地。墓はオレンジと黄色の花で飾られ、カラフルな砂糖菓子や張り子細工の頭蓋骨は、腐敗が進んで永遠の笑顔を浮かべているようだ。けれども、この漁師町の墓地ほどに私の心を動かす墓所はなかった。まるで円と四角の間の妥協のようなその独特の輪郭に、私はまさにユートピアの象徴を見たように思った。死を目の前にしつつ生きるというユートピアが、そこに実現されていた。長いこと私は確信していた。デンマーク語で「小さな島」とか「水に囲まれた」という意味の名前を持つこの場所に住む人々は、同じくらいの緯度の国々で通常行われているように、共同体の内部から市門の外へと死者たちを追放する代わりに、死者たちを文字通り町の中心に迎え入れた。だからこそ、より生に近いのだと。もっとも都市空間もまた人口膨張のために、ほどなくして墓地をふたたび内部へと取り込まざるを得なくなるのだけれど。\end{CJK} \\
\midrule
A Children's Bible & en-pl & The lady urinated. \newline
“Oh, poor old thing, she has a nervous bladder!” exclaimed someone’s chubby mother. “Is that a Persian rug?” \newline
Whose mother was it? Unclear. No one would cop to it, of course. We canceled the performance. \newline
“Admit it, that was your mother,” said a kid named Rafe to a kid named Sukey, when the parents had filed out. Some of their goblets, highball glasses, and beer bottles were completely empty. Drained. \newline
Those parents were in a hurry, then. \newline
“No way,” said Sukey firmly, and shook her head. \newline
“Then who is your mother? The one with the big ass? Or the one with the clubfoot?” \newline
“Neither,” said Sukey. “So fuck you.” & Dama się posikała. \newline
– Och, biedactwo, ma wrażliwy pęcherz! – wykrzyknęła czyjaś pulchna matka. – Zaraz, to perski dywan? \newline
Czyją matką była? Nie wiadomo. Oczywiście nikt nie chciał się przyznać. Odwołaliśmy przedstawienie. \newline
– No dawaj, to twoja – powiedział chłopiec imieniem Rafe do dziewczynki imieniem Sukey, kiedy rodzice sobie poszli. Zostawili po sobie kieliszki, wysokie szklanki i butelki po piwie. Niektóre były zupełnie puste. Do ostatniej kropelki. \newline
Tym z rodziców się zatem spieszyło. \newline
– W życiu – odparła Sukey stanowczo i pokręciła głową. \newline
– To która? Ta z wielkim dupskiem? Czy ze szpotawą stopą? \newline
– Ani jedna, ani druga. Spierdalaj.\\
\bottomrule
\end{tabular}
}
\caption{Examples of aligned reference source and target paragraphs from our dataset, including both a narrative (\textit{An Inventory of Losses}) and a dialogue (\textit{A Children's Biblie}). Our \textsc{Para} approach takes as input the entire source paragraph and outputs a paragraph-level translation.}
\label{tab:dataset_examples}
\end{table*}

\subsection{Dataset collection}
Literary texts (e.g., novels or short stories) pose unique challenges for translators due to their complex nature. Translators must interpret and honor the author's voice with no objective reality to measure against, which can result in several equally valid translations \cite{Sager1998}. For machine translation systems, these challenges exacerbate the need for discourse-level context~\cite{thai-etal-2022-exploring}: an author's intended meaning or style is often unclear from just a single sentence.\vspace{-0.7cm}

\begin{center}
\begin{table*}[ht]
\centering
\resizebox{\textwidth}{!}{
\begin{tabular}{lllrrrr}
        \toprule
                               &                      &                      & \multicolumn{2}{c}{\textsc{Language}}  & \multicolumn{2}{c}{\textsc{Year Published}}         \\
        \multirow{-2}{*}{\textsc{Book title}}  & \multirow{-2}{*}{\textsc{Author}}  & \multirow{-2}{*}{\textsc{Translator(s)}} & \textsc{Source} & \textsc{Target}        & \textsc{Translation} & \textsc{Original} \\
        \midrule
        \textbf{A Children's Bible}             & Lydia Millet           &    Aga Zano  & \textit{en}        & \textit{pl}                 & 2022                 & 2020                \\
        \textbf{What Can You See From Here} & Mariana Leky  & Agnieszka Walczy   & \textit{de}        & \textit{pl}                 & 2021                 & 2017                \\
        \textbf{The Years}           & Annie Ernaux           & Krzysztof Jarosz \& & \textit{fr}        & \textit{pl}                 & 2022                 & 2008                \\
        & & Magdalena Budzińska &        &                  &                 &               \\
        \textbf{Manaraga}  & Wladimir Sorokin  & Agnieszka Lubomira Piotrowska  & \textit{ru} & \textit{pl} & 2018  & 2017  \\
        \textbf{Crows}  & Petra Dvorakova &  Mirosław Śmigielski  & \textit{cs} & \textit{pl} & 2020   & 2020                \\
        \textbf{Convenience Store Woman}       & Sayaka Murata          &   Dariusz Latoś   & \textit{ja}                 & \textit{pl}                 & 2019                 & 2016                \\
        \midrule
        \textbf{Sixteen Horses} & Greg Buchanan & Fuji Yoshiko  & \textit{en}  & \textit{ja}  & 2022  & 2021   \\
        \textbf{An Inventory of Losses}       & Judith Schalansky      &    Naoko Hosoi & \textit{de}        & \textit{ja}                 & 2022                 & 2018                \\
        \textbf{Dear Reader}                     & Paul Fournel           &  Kei Takahashi  & \textit{fr}       & \textit{ja}                 & 2022                 & 2011                \\
        \textbf{The Shooting Party}            & Anton Chekhov          &   Takuya Hara   & \textit{ru}        & \textit{ja}                 & 2022                 & 1884                \\
        \textbf{Sword of Destiny}              & Andrzej Sapkowski      &  Yasuko Kawano  & \textit{pl}        & \textit{ja}  & 2022     & 1992 \\
        \textbf{Bare burial}  & Fang Fang & Shin'ichi Watanabe & \textit{zh}  & \textit{ja} & 2022 & 2016 \\
        \midrule
        \textbf{What Can You See From Here} & Mariana Leky           &   Tess Lewis   & \textit{de}        & \textit{en}                 & 2021                 & 2017                \\
        \textbf{The Years}    & Annie Ernaux & Alison L. Strayer & \textit{fr}   & \textit{en}  & 2017 & 2008  \\
        \textbf{The Story of a Life}         & Konstantin Paustovsky  &   Douglas Smith   & \textit{ru}        & \textit{en}                 & 2022                 & 1956                \\
        \textbf{The Books of Jacob}  & Olga Yokarczuk         &  Jennifer Croft   & \textit{pl}        & \textit{en}      & 2022 & 2014                \\
        \textbf{Convenience Store Woman} & Sayaka Murata &  Ginny Tapley Takemori & \textit{ja}        & \textit{en}   & 2018 & 2016 \\
        \textbf{Cocoon} & Zhang Yueran  &   Jeremy Tiang & \textit{zh} & \textit{en}  & 2022 & 2018 \\     
        \bottomrule                    
\end{tabular}}
\caption{Details of the translated novels used in our  study. In cases where the same novel is used for multiple target languages (e.g., ``The Years''), identical source paragraphs are extracted to enable comparisons across language pairs. These novels exhibit distinct differences beyond just their source languages. For instance, ``What Can You See From Here'' presents a philosophical exploration of life and death, while ``Sword of Destiny'' is a fantasy story part of ``The Witcher'' saga.}
\label{table:novel_list}
\end{table*}
\end{center}

\paragraph{Selecting paragraphs from novels:}
How good are machines at translating literary paragraphs? To answer this question, we extract \textbf{20} paragraphs (dialogues and narrative texts) each from \textbf{18} recently-published translations of novels, and we manually align these paragraphs with corresponding paragraphs in the source novel\footnote{In most cases, we purchase the source ebook and its corresponding translation before extracting aligned paragraphs, but for a few books, we utilized Amazon's free preview functionality.} (see \autoref{tab:dataset_examples}). The target language of each translation is English, Polish, or Japanese (6 books for each), and we consider eight different source languages. Almost all of the translations were published after 2021 (see \autoref{table:novel_list}), which is important to avoid data contamination with the pretraining data of large language models. In sum, we obtain \textbf{360} aligned source-target paragraphs, which we use for all of the experiments described in the rest of the paper.

\paragraph{Paragraph length:}
All paragraphs consist of at least two sentences, and the majority of them are between four to nine sentences long (mean=7.45, std=4.14).\footnote{A paragraph with fewer sentences is not necessarily short: for example, in the German novel ``An Inventory of Losses,'' sentences can be as long as 70 to 80 words, with the longest reaching 117 words.} As automatic sentence tokenizers are not always reliable for all of the languages considered in our study, we manually perform sentence tokenization to enable a direct comparison of sentence and paragraph-level translation systems. For more details about the dataset statistics, including token and sentence counts, see \autoref{table:tokens_in_paragraphs} and \autoref{table:sent_auto_in_paragraphs}.

\paragraph{Target language selection:} We select English, Japanese, and Polish as the target languages of our study, as these languages differ considerably in many linguistic aspects. English is an analytic language that is widely spoken and extensively studied in the field of natural language processing, and it serves as the primary pretraining language of most large language models, including \textsc{Gpt-3.5}.\footnote{As of 2020, the reported distribution of languages featured in the present study within the \textsc{Gpt-3} training data was as follows: English -- 92.647\% (1st), French -- 1.818\% (2nd), German -- 1.469\% (3rd), Russian -- 0.188\% (9th), Polish -- 0.155\% (11th), Japanese -- 0.111\% (15th), Chinese -- 	0.099\% (17th), Czech -- 0.071\% (18th) (see \url{https://github.com/openai/gpt-3/blob/master/dataset_statistics/languages_by_word_count.csv}). The current \textsc{GPT-3.5} \texttt{text-davinci-003} model is reported to incorporate data up to June 2021 and it is unclear what texts or languages were added to the original training data \url{https://platform.openai.com/docs/models/gpt-3-5}.} In contrast, both Japanese and Polish are comparatively under-explored. Japanese is an agglutinative language that employs three distinct writing systems: Kanji, Hiragana, and Katakana. As a high-context language, the translation of Japanese texts necessitates a profound comprehension of context and cultural nuances, rendering it a compelling choice for testing the limits of LLMs' translation capabilities. Polish, on the other hand, is a fusional language characterized by a rich morphological system. Its complex word forms, grammatical gender, conjugation, and declension make it an apt choice for testing the accuracy and robustness of LLMs.\footnote{The first author is fluent in all three target languages.}

\paragraph{Source language selection:} As source languages, we select English (\textit{es}), Polish (\textit{pl}), Russian (\textit{ru}), Czech (\textit{cs}), French (\textit{fr}), German (\textit{de}), Japanese (\textit{ja}), and Chinese (\textit{zh}). These languages belong to a diverse array of language families -- Indo-European (Romance, Germanic, Slavic), Sino-Tibetan, and Japonic -- each with distinctive morphological traits -- fusional, agglutinative, and analytic. Moreover, they employ a variety of writing systems such as the Latin alphabet, the Cyrillic alphabet, Hanzi, and Kanji/Hiragana/Katakana (see \autoref{table:language_comparison} in \autoref{app:dataset} for details). Finally, we carefully select source-target language pairs to ensure that our study encompasses both linguistically similar and dissimilar languages. For example, we paired \textit{cs}-\textit{pl}, as these languages are characterized by only 10\% lexical distance\footnote{i.e., the percentage of non-cognates in the language pair.} 
and have similar syntactic structures \cite{Jgrov2023czech}. Conversely, we also include \textit{ja}-\textit{pl}, as the two languages have very little lexical overlap, vastly different grammars, and utilize distinct writing systems.

\subsection{Translation with large language models}
In this paper, we focus on translating the literary paragraphs in our dataset using large language models. More specifically, we use the \textsc{Gpt-3.5} \texttt{text-davinci-003} checkpoint, which has been further tuned to follow instructions based on human feedback~\citep{ouyang2022training}.~\citet{hendy2023good}  demonstrate that \textsc{GPT-3.5} produces translations of reasonable quality, though their focus was mostly at the sentence level.
 Since many LLMs including \textsc{Gpt-3.5} are only accessible via black-box APIs, we adapt the model for translation via in-context learning~\citep{brown2020language}.

\paragraph{Demonstration examples:} 
We use few-shot prompting, in which a model is provided with a prompt consisting of five demonstrations. We manually curate the five demonstrations from literary texts for each of the 18 language pairs, resulting in 90 total demonstration examples. These demonstrations are sourced from novels that are \emph{not} part of our translation dataset, resulting in potential differences in topic and style (see \autoref{tab:prompt_novel_data} in the \autoref{app:dataset} for details). We further ensure that each set of five demonstrations includes both dialogues and narrative texts.

\paragraph{Prompting for translation:}
We consider the following three prompting strategies for \textsc{Gpt-3.5} that allow us to compare the model's abilities to translate with and without discourse-level context (see \autoref{tab:prompt_templates} for templates and \autoref{app:prompt_examples} for the exact prompts):

\begin{itemize}
    \item \textbf{\textsc{Gpt-3.5} sentence-level translation without context (\textsc{Sent}):} Each sentence of the paragraph is translated in isolation of the others. To maintain consistency, we provide the same five \textit{sentence}-level examples\footnote{Sentence-level demonstrations for \textsc{Sent} are sampled from the demonstrations for paragraph-level translation.} in each prompt for the given source-target language pair.\footnote{To ensure consistent quotation mark usage and enable a fair comparison with paragraph-level translations, quotation marks in sentence-level translations were manually adjusted.}

    \item \textbf{\textsc{Gpt-3.5} sentence-level translation \emph{with} context (\textsc{Para\_Sent}):} Each sentence of the paragraph is translated in context. The model is provided with the entire source paragraph as input, where the sentence to be translated is wrapped in \texttt{<translate>} and \texttt{</translate>} tags, in addition to a partially-translated target paragraph. The demonstrations in the prompt also contain  \texttt{<translate>} and \texttt{</translate>} tags wrapped around one sentence per demonstration. For each demonstration in the prompt, a sentence in a different position was chosen (e.g., from the beginning, middle, and end of the paragraph).

    \item \textbf{\textsc{Gpt-3.5} paragraph-level translation (\textsc{Para}):} The entire source paragraph is passed into the model, and the output target paragraph is generated conditioned on this input (i.e., without any sentence tokenization). Demonstrations in the prompt are also \textit{paragraphs}\footnote{
The examples for \textsc{Para} and \textsc{Para\_Sent} configurations are necessarily lengthier. Due to the \textsc{Gpt-3.5} maximum context size, it is not always possible to include all five examples within the prompt. Consequently, around 10\% of the data was translated using four or fewer examples.} of translations from the respective source language into the target language in question.\footnote{Initially, we experimented with \textsc{GPT-3} by translating between two non-English languages using English as a pivot, as it is the primary language of the model. The model had access to the source text and its English translation. After manual evaluation and comparison to translations without a pivot language, we found no significant benefit in using English as the pivot. Consequently, we directly translated paragraphs into the target language. Refer to  \autoref{app:pivot_pilot}. for details and results of this preliminary study.}

\end{itemize}

\begin{table}[t]
\small
    \setlength{\tabcolsep}{4pt}
    \centering
    \resizebox{0.9\columnwidth}{!}{
    \begin{tabular}{c p{8cm}}
    \toprule
    & \multicolumn{1}{c}{\bf \textsc{Sent} Prompt Template} \\
    \midrule
     & \texttt{Original text in [SRC LANG]: } \\
     \noalign{\vskip 1.5mm}
     & \texttt{source sentence} \\
     \noalign{\vskip 3mm}
     & \texttt{Translation into [TRG LANG]:} \\
     \noalign{\vskip 1.5mm}
     & \texttt{target sentence} \\
    \toprule
    & \multicolumn{1}{c}{\bf \textsc{Para\_Sent} Prompt Template} \\
    \midrule
     & \texttt{Original text in [SRC LANG]:} \\
     \noalign{\vskip 1.5mm}
     & \texttt{source prefix}\\
     & \texttt{<translate> src sent </translate>} \\
     & \texttt{source suffix}\\
     \noalign{\vskip 3mm}
     & \texttt{Translation into [TRG LANG]:} \\
     \noalign{\vskip 1.5mm}
     & \texttt{target prefix} \\
     & \texttt{<translated> trg sent </translated>} \\
    \toprule
    & \multicolumn{1}{c}{\bf \textsc{Para} Prompt Template} \\
    \midrule
     & \texttt{Original text in [SRC LANG]: } \\
     \noalign{\vskip 1.5mm}
     & \texttt{source paragraph} \\
     \noalign{\vskip 3mm}
     & \texttt{Translation into [TRG LANG]:} \\
     \noalign{\vskip 1.5mm}
     & \texttt{target paragraph} \\
    \bottomrule
    \end{tabular}
    }
    \caption{Prompt templates for \textsc{Sent}, \textsc{Para\_Sent}, and \textsc{Para}.}
    \label{tab:prompt_templates}
\end{table}

\paragraph{Using Google Translate (\textsc{GTr}) as a baseline:} In order to compare commercial-grade translation systems to LLM translators, we also translate all paragraphs in our dataset using Google Translate.\footnote{All paragraphs were translated in January 2023 using the \texttt{GoogleTranslate} API.} We opt for an off-the-shelf commercial system instead of a state-of-the-art system from, for instance, WMT competitions for two primary reasons. First, our experiments focus on \textit{literary} translations. Given that WMT systems are predominantly evaluated on the news domain, it is uncertain which system would perform best, and some language pairs may not even be supported. Second, our main research question revolves around LLMs' ability to incorporate contextual information, rather than merely comparing their performance with state-of-the-art translation systems. We employ \textsc{GTr} as a reasonably robust baseline to assess the extent to which context can enhance MT quality, rather than asserting that LLMs outperform \textit{all} traditional MT systems.

\section{Evaluating document-level literary translation}
How do we compare the translation quality of the systems described above? Automatic metrics such as BLEURT and COMET are untested on document-level inputs as well as literary texts, and as such we do not consider them reliable, although we do report them in \S\ref{subsec:auto_metrics}.\footnote{Automatic metrics developed specifically for document-level MT are also insufficient as they either work best with one-to-one sentence level alignments \cite{easy_doc_mt, hendy2023good} or are available only for English \cite{jiang-etal-2022-blonde}.} Human evaluation is equally problematic, as direct assessments of translation quality (e.g., ``rate the quality of this translation from 0-100'') suffer from calibration issues that are exacerbated with longer texts~\cite{karpinska-etal-2021-perils}. Thus, we opt for a human evaluation inspired by Multidimensional Quality Metrics~\citep[][\textsc{MQM}]{lommel2014multidimensional}, in which annotators mark and classify error spans within the translation. Specifically, for each of the 18 language pairs studied in this work, we hire translators to identify all span-level errors in two competing translations. For each evaluated pair, the annotators were also asked to choose the better translation and provide a free-form rationale. For each source paragraph, the translators make three binary judgments of which translation is higher quality: \textsc{Sent} vs \textsc{Para}, \textsc{Para\_Sent} vs \textsc{Para}, and \textsc{GTr} vs \textsc{Para}.

\begin{figure}[t]
\centering
\includegraphics[width=\columnwidth]{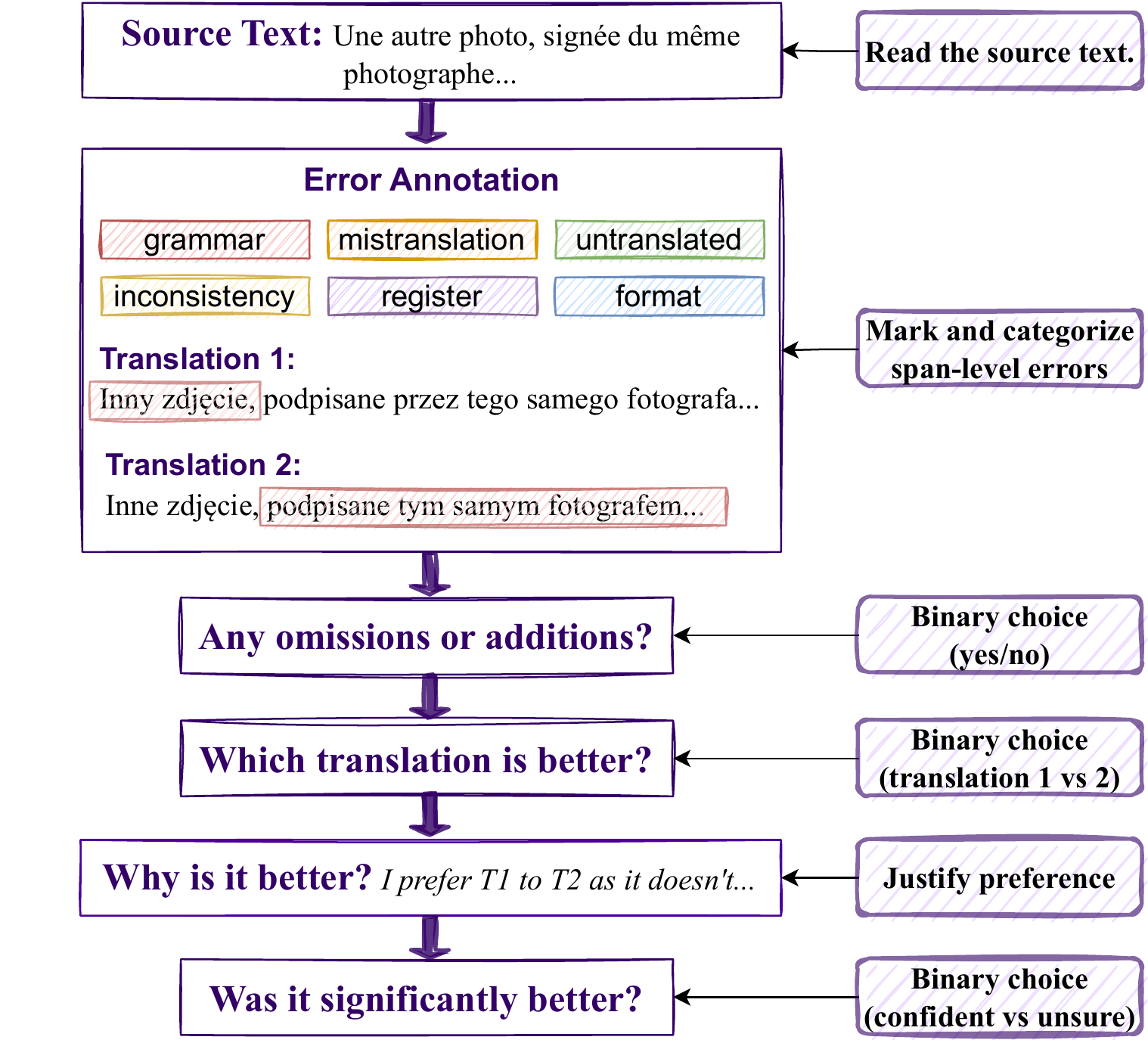}
\caption{A description of the annotation process for a pair of candidate translations given a source paragraph. Note that our hired translators go through this pipeline for \emph{three} different pairs per source paragraph, comparing \textsc{Para} with \textsc{Sent}, \textsc{Para\_Sent}, and \textsc{GTr}.}
\label{figure:annotations_processs_flowchart}
\end{figure}

\label{sec:annotation_schema}
\paragraph{Recruiting annotators:} As our task is complex and requires fluency in both the source and target language, we hire \emph{translators} to provide the annotations. We recruit 13 translators, each of whom is a native speaker of English, Polish, or Japanese\footnote{The annotators for Czech-Polish and Russian-English were both native speakers of the respective source languages and highly proficient in their respective target languages. They collaborated with native speakers of the target languages, who possessed a basic understanding of the source language, to complete their annotations.} 
through the Upwork freelancing platform.\footnote{\url{https://www.upwork.com/}} One translator, hired directly, was a bilingual speaker of English and Polish with advanced knowledge of German; as such, she performed the \textit{pl-en}, \textit{de-en}, and \textit{de-pl} evaluations. Evaluation of \textit{ja-pl}, \textit{pl-ja}, and \textit{pl-en} texts was done by the first author in a collaboration with native speakers of Polish/Japanese to avoid any potential bias.
Each translator was paid \$2 per evaluated pair of candidate translations, with an additional \$5 bonus to cover the time spent familiarizing themselves with the instructions. We asked them to compare three pairs of system translations (\textsc{Para} vs. \textsc{Sent}, \textsc{Para} vs. \textsc{Para\_Sent}, \textsc{Para} vs. \textsc{GTr})  for 10 paragraphs per language pair; as such, 180 total source paragraphs were used in our evaluations. Altogether, we paid approximately \$12 per hour, with a total cost of \$955. 

\paragraph{Annotation task:} First, we tasked the hired translators\footnote{They were presented with guidelines in their native language. The annotation task was performed using the LabelStudio annotation tool \cite{LabelStudio}. See \autoref{figure:annot_interface} for the screenshot of the interface.} with annotating a subset of MQM translation errors identified through a pilot analysis and annotation of the system's outputs. Specifically, we ask them to highlight spans within the candidate translations that contain errors belonging to any of the following error categories: 

\begin{itemize}
    \item \colorbox{orange!6}{\textbf{\color{darkorange}mistranslation:}}\footnote{We note that mistranslations in literary text are often not as grave as, for instance, in news articles. Human translators hold \textit{poetic license}, which allows them to change some details to make the text more enjoyable for the reader. Is changing ``bonito’' into ``tuna’' incorrect? Or can it be perceived as a way to accommodate an English-speaking readership that is likely more familiar with the latter? 
    } accuracy errors that occur when the wrong target word or phrase is chosen to represent content from the source text. In addition to canonical mistranslations, we also include \emph{overly literal} translation errors that occur when systems translate word-by-word into the target language even though the result is nonsensical.
    \item \colorbox{red!6}{\textbf{\color{red}grammar:}} grammatical errors, such as errors in conjugation or declension, wrong prepositions, etc.
    \item \colorbox{teal!6}{\textbf{\color{teal}untranslated:}} words or phrases that should have been translated into the target language but were either left in the source language or just transliterated into the target language.
    \item \colorbox{yellow!10}{\textbf{\color{darkyellow}inconsistency:}} use of different terms to refer to the same entity, or different words where the same word should be used for stylistic reasons (e.g., ``Kasia'' and ``Kate,'' ``coat'' and ``jacket,'' or ``bad'' and ``awful'' );
    \item \colorbox{violet!6}{\textbf{\color{violet}register:}} a clear violation in the use of formal and informal language within the same text, only annotated in Japanese\footnote{We only annotate cases where the level of formality changes abruptly within the same paragraph. It is possible that a given character would be more likely to use formal language but an informal language is being employed. As long as this is consistent we do not consider it an error as this cannot be fully determined from the paragraph context.}
    \item \colorbox{blue!6}{\textbf{\color{darkblue}format:}} incorrect usage of punctuation (e.g., "." instead of \begin{CJK}{UTF8}{min}"。").\end{CJK}
\end{itemize}

After the span-level annotation is complete, we then ask the translators to further identify if any of the candidate translations contains significant content \textbf{additions} or \textbf{omissions} in relation to the source text.\footnote{Note that this task was simplified to a binary choice -- either there were serious omissions/additions or not. We did not ask the annotators to further annotate them due to the time restrictions.} Finally, they are asked to \textbf{choose the better translation} and provide a justification for their choice in two to five sentences. We instruct them to additionally mark whether their chosen translation is significantly superior, or if the decision was difficult because both translations are of roughly comparable quality (see \autoref{figure:annotations_processs_flowchart} and \autoref{app:hum_eval} for details).

\section{Results}
\label{sec:results}
In this section, we compare our different literary translation methodologies using both automatic metrics and aggregate statistics from the human evaluations. Overall, we observe that the \textsc{Para} configuration outperforms competing methods across all evaluations and language pairs. These results demonstrate that \textsc{Gpt-3.5} effectively leverages paragraph-level context to produce better translations than sentence-level methods, and also that the less efficient sentence-by-sentence translation with context is (\textsc{Para\_Sent}) is unnecessary to achieve high translation quality.

\begin{table}[!t]
\centering
\resizebox{\columnwidth}{!}{%
\begin{tabular}{lcccc}
\toprule
\textsc{System} & \textsc{Comet} & \textsc{Bleurt} & \textsc{BertScore} & \textsc{Comet-QE} \\
\midrule
\rowcolor{violet!20} \textsc{Para} & \textbf{0.785}  & \textbf{0.485} & \textbf{0.840} & \textbf{0.038} \\
\textsc{Sent}      & 0.779  & 0.469 & 0.839 & -0.052 \\
\textsc{Para\_Sent} & 0.780  & 0.480 & 0.838 & -0.062 \\
 \textsc{GTr}       & 0.735  & 0.443 & 0.832 & -0.156 \\
\bottomrule
\end{tabular}%
}
\caption{Results of automatic evaluation. A higher number indicates better scores.}
\label{tab:translation_auto_metrics}
\end{table}

\subsection{Automatic metrics favor \textsc{Para}} 
\label{subsec:auto_metrics}
We assess the translation from all four systems using the reference-based \textsc{Comet} \cite{rei-etal-2022-comet}, \textsc{Bleurt} \cite{sellam-etal-2020-bleurt}, and \textsc{BertScore} \cite{zhang2020bertscore} metrics, as well as the reference-free \textsc{Comet-QE} \cite{rei-etal-2021-references}\footnote{We use the newest \texttt{wmt22-comet-da} checkpoints for \textsc{Comet}, \texttt{Bleurt-20} checkpoints for \textsc{Bleurt}, \texttt{wmt20-comet-qe-da} checkpoints for \textsc{Comet-QE}, and the HuggingFace implementation which employs \texttt{roberta-large} for \textsc{BertScore}.} metric.
Although these metrics were not explicitly designed for evaluating paragraph-level outputs and their results should be interpreted with caution, they prove more reliable than string-based metrics like \textsc{Bleu}, especially for literary translations \cite{thai-etal-2022-exploring, karpinska-etal-2022-demetr, gehrmann2022repairing}. 
\autoref{tab:translation_auto_metrics} shows the effectiveness of the \textsc{Para} translation method: a statistical analysis with linear mixed-effects models \cite{Baayen2008_lme} demonstrates that \textsc{Para} significantly outperforms \textsc{Sent} and \textsc{GTr} based on \textsc{Comet}, \textsc{Bleurt}, and \textsc{Comet-QE} scores (\textit{p}<.001), and surpasses \textsc{GTr} based on the \textsc{BertScore} results (\textit{p}<.001).\footnote{We present more details of this analysis in \autoref{app:stats}.}

\subsection{Human evaluation also favors \textsc{Para}} 
\begin{figure}[t]
\centering
\includegraphics[width=1\columnwidth]{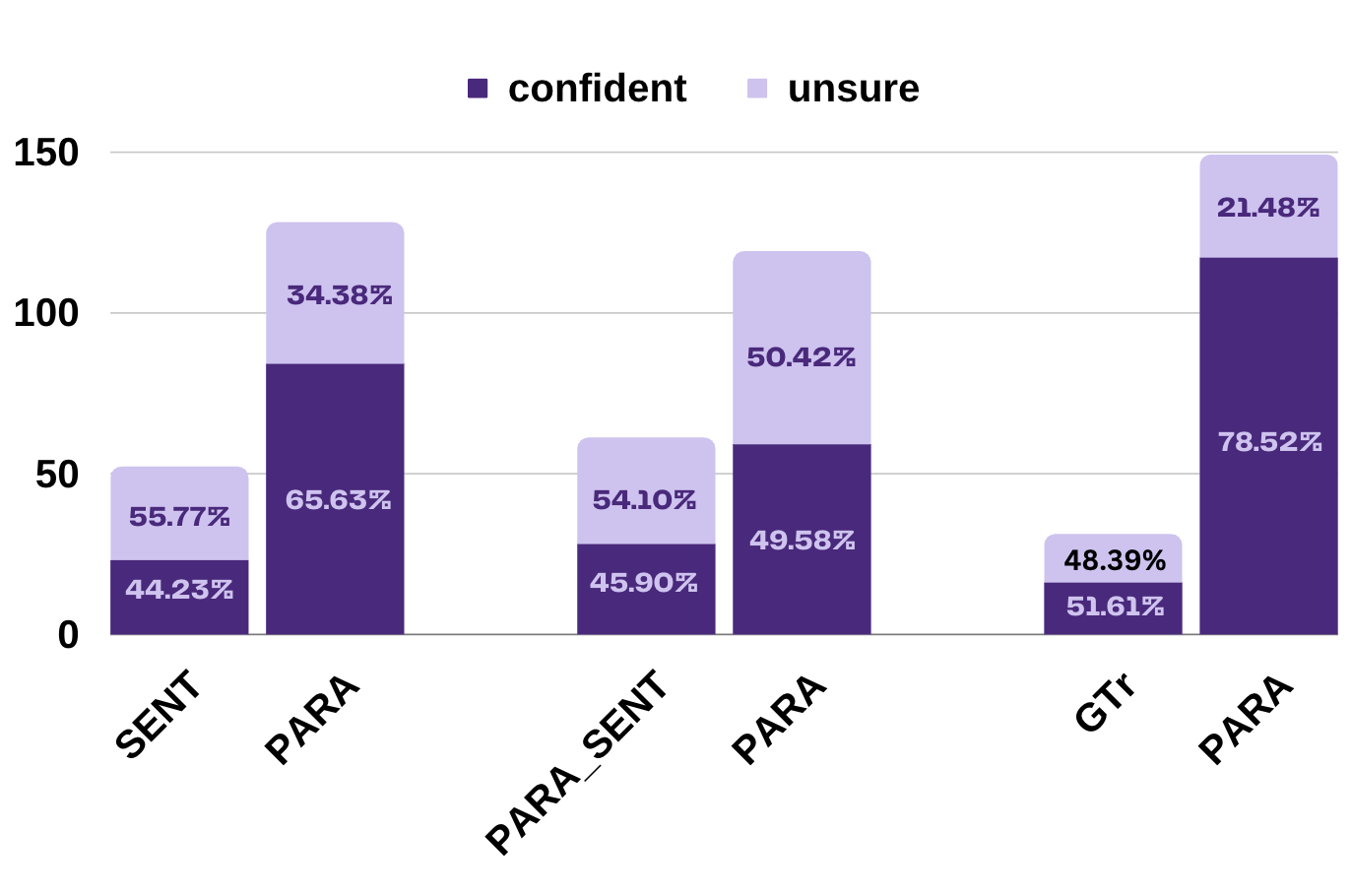}
\caption{The number of votes for \textsc{Sent} vs \textsc{Para}, \textsc{Para\_Sent} vs \textsc{Para}, and \textsc{GTr} vs \textsc{Para} along with rater confidence (\emph{confident} or \emph{unsure}). \textsc{Para} is preferred to all competing methods. All differences are statistically significant at \textit{p}<.001 (binomial test).}
\label{figure:hum_eval_confid_results}
\end{figure}

\autoref{figure:hum_eval_confid_results} contains human preference results comparing \textsc{Para} to \textsc{Sent}, \textsc{Para} to \textsc{Para\_Sent}, and \textsc{Para} to \textsc{GTr}, aggregated across all 18 language pairs studied in this paper (i.e., 180 votes per system comparison). \autoref{table:vote_count_by_lang_pair_humeval} breaks down these results for each language pair, and we observe the same trends for the vast majority of pairs. Overall, the translators significantly favored  \textsc{Para} translations over the alternatives (\textit{p}<.001, binomial test). ~\autoref{tab:error_count_by_trg_lang} contains specific information about grammar and mistranslation errors split across the three target languages (see \autoref{tab:grammar_errors_categories} and \autoref{table:mistrans_type_class_by_trg} for details), which we use to discuss the three comparison settings in more detail below.

\begin{figure}[t!]
\centering
\includegraphics[width=.85 \columnwidth]{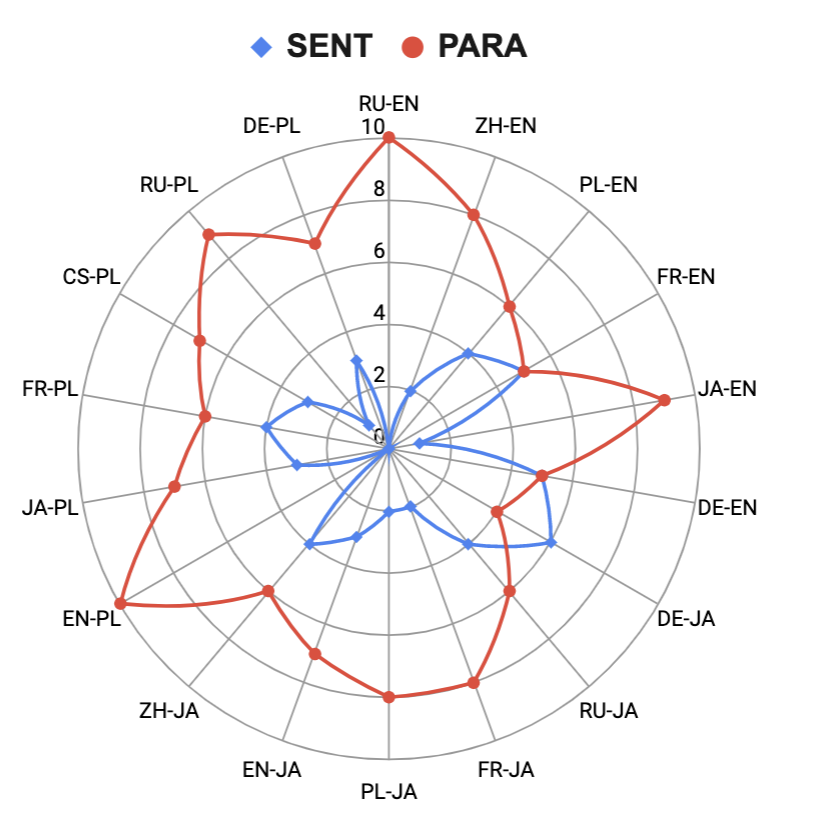}
\caption{The distribution of translator preference judgments between sentence-level translation (\textsc{Sent}) and paragraph-level translation (\textsc{Para}). \textsc{Para} is preferred (i.e., more votes) in every language pair except  \textit{de-ja}, \textit{fr-en} and \textit{de-en}.}
\label{figure:sent_vs_para_in_intro}
\end{figure}


\paragraph{\textsc{Para} is clearly better than \textsc{Sent}:}
\textsc{Para} is preferred by translators over \textsc{Sent} at a rate of 71.1\% (\textit{p}<.001, 95\% CI [0.639, 0.776]). Additionally, when translators preferred \textsc{Para}, they were usually confident in the decision (i.e., it was clearly better than \textsc{Sent}); even if we exclude all ``unsure'' votes, the preference for \textsc{Para} translations remains significant at 78.5\% (\textit{p}<.001, 95\% CI [0.695, 0.859]). The only language pair in which \textsc{Sent} is favored over \textsc{Para} is \textit{de-ja} (see \autoref{figure:sent_vs_para_in_intro}). This result may be attributed to the fact that the German
novel \textit{An Inventory of Losses} by Judith Schalansky, used for this language pair, contains the longest sentences in our dataset (on average 45 tokens per sentence), which means that the intra-sentence context is likely more informative than in other books (see \autoref{table:tokens_in_paragraphs}). 
Overall, \textsc{Sent} translations contain 29.5\% more mistranslations, 65.4\% more grammatical errors, over 12 times more inconsistency errors, and three times more register errors (see \autoref{tab:error_count_by_trg_lang}).

\paragraph{\textsc{Para} is clearly better than \textsc{GTr}:}
\textsc{Para} translations are overwhelmingly preferred over  those from Google Translate (\textsc{GTr}), with an 82.8\% preference rate (\textit{p}<.001, 95\% CI [0.765, 0.880]). Even after removing the ``unsure'' votes, the preference for \textsc{Para} remains significant at 88.0\% (\textit{p}<.001, 95\% CI [0.812, 0.930]). In the \textit{fr-ja}, \textit{pl-ja}, \textit{zh-ja}, and \textit{cs-pl} language pairs, \textsc{Para} received \textit{all} of the ten votes over \textsc{GTr}.
Part of this advantage may be attributed to \textsc{GTr} sometimes using English as a pivot language, which can result in information loss. Our Czech translator observed that mistakes in \textsc{GTr} translations suggest the text was first translated into English.\footnote{For the \textit{cs-pl} language pair, we separately annotated mistranslations arising from pivot translation. These errors accounted for over 50\% of all mistranslations in that language pair. The elimination of the need for parallel data may therefore be beneficial for translating between lower-resource languages where sufficient parallel data is often unavailable necessitating the pivot translation.}
Overall, \textsc{GTr} translations result in 57.7\% more mistranslations, 37.3\% more grammatical errors, over twice as many inconsistency errors, and ten times more register errors (see \autoref{tab:error_count_by_trg_lang}). Additionally, \textsc{GTr} produced 125 format errors while \textsc{Para} produced perfect outputs in this regard. Finally, it is worth noting that \textsc{GTr} left fewer words untranslated, though this is inflated by the fact that in one German text, the word ``Bauer'' (``farmer'') was untranslated 14 times in the \textsc{Para} translation.

\paragraph{\textsc{Para} is slightly preferred over \textsc{Para\_Sent}:}
Our evaluations show that \textsc{Para} is better than \textsc{Para\_Sent}, but the gap is smaller than it is for the other two methods. 
\textsc{Para} is still preferred at a 66.1\% rate (\textit{p}<.001, 95\% CI [0.587, 0.730]).
After removing the ``unsure'' votes, \textsc{Para} remains the preferred option at a rate of 67.8\% (\textit{p}<.001, 95\% CI [0.569, 0.774]). Notably, the error distribution of both translations is more similar than in previous cases. Both \textsc{Para} and \textsc{Para\_Sent} result in a comparable number of mistranslations (\emph{480} vs \textsc{465}), grammar errors (\emph{102} vs \emph{110}), and inconsistencies (\emph{2} vs \emph{3}) (see \autoref{tab:error_count_by_trg_lang}). While \textsc{Para\_Sent} leaves around 22\% more words untranslated, it appears to leverage the contexts and even occasionally selects better equivalents in the target language, as evidenced by translator comments. One major issue with \textsc{Para\_Sent} is that it occasionally repeats  sentences, whereas \textsc{Para} never does so.

\begin{table}[!tbp]
\centering
\resizebox{0.45\textwidth}{!}{%
\begin{tabular}{@{}llcccc@{}}
\toprule
\textsc{Type} & \textsc{Trg Lang} & \textsc{Para} & \textsc{Sent} & \textsc{Para\_Sent} & \textsc{GTr} \\ \midrule
\textsc{Mistranslation} & \textsc{En} & 87 & 104 & 81 & 150 \\
 & \textsc{Ja} & 223 & 294 & 229 & 334 \\
 & \textsc{Pl} & 170 & 224 & 155 & 273 \\
\rowcolor{violet!20} & \textsc{Total} & \emph{480} & \emph{622} & \emph{465} & \emph{757} \\ \addlinespace
\textsc{Grammar} & \textsc{En} & 5 & 20 & 9 & 18 \\
 & \textsc{Ja} & 42 & 50 & 37 & 63 \\
 & \textsc{Pl} & 55 & 86 & 64 & 59 \\
\rowcolor{violet!20} & \textsc{Total} & \emph{102} & \emph{156} & \emph{110} & \emph{140} \\ \addlinespace
\textsc{Inconsistency} & \textsc{En} & 0 & 5 & 0 & 1 \\
 & \textsc{Ja} & 1 & 7 & 2 & 7 \\
 & \textsc{Pl} & 1 & 13 & 1 & 7 \\
\rowcolor{violet!20} & \textsc{Total} & \emph{2} & \emph{25} & \emph{3} & \emph{15} \\ \addlinespace
\textsc{Untranslated} & \textsc{En} & 13 & 5 & 14 & 6 \\
 & \textsc{Ja} & 23 & 31 & 33 & 24 \\
 & \textsc{Pl} & 23 & 16 & 25 & 4 \\
\rowcolor{violet!20} & \textsc{Total} & \emph{59} & \emph{52} & \emph{72} & \emph{34} \\ \addlinespace
\textsc{Register} & \textsc{En} & 0 & 0 & 0 & 0 \\
 & \textsc{Ja} & 7 & 25 & 13 & 71 \\
 & \textsc{Pl} & 0 & 0 & 0 & 0 \\
\rowcolor{violet!20} & \textsc{Total} & \emph{7} & \emph{25} & \emph{13} & \emph{71} \\ \addlinespace
\textsc{Format} & \textsc{En} & 0 & \textit{n/a} & \textit{n/a} & 1 \\
 & \textsc{Ja} & 0 & \textit{n/a} & \textit{n/a} & 116 \\
 & \textsc{Pl} & 0 & \textit{n/a} & \textit{n/a} & 8 \\
\rowcolor{violet!20} & \textsc{Total} & \emph{0} & \textit{n/a} & \textit{n/a} & \emph{125} \\ \bottomrule
\end{tabular}%
}
\caption{Total counts of all of the types of mistakes made by each of the four systems from our annotation. Overall, models with access to paragraph-level context commit fewer translation errors.}
\label{tab:error_count_by_trg_lang}
\end{table}

\paragraph{What do translators think about \textsc{Para}?}
To wrap up this section, we provide a qualitative analysis of the free-form comments written by translators to justify their preference judgments.  Overall, the translators praise \textsc{Para} for its \textit{more skillful use of rhetoric devices}, and \textit{surpas[ing]} \textsc{Sent} \textit{as a literary rendition}. They also mention that \textsc{Para} \textit{uses more of a poetic license but this makes it stylistically much smoother} than \textsc{Sent}. 
Furthermore, translators state that \textsc{Para} \textit{clearly better reflects the content and style of the original} when compared to \textsc{GTr}, and that it \textit{stays consistent within the paragraph}.
Inevitably, translations are not flawless, and there are instances where both compared systems fall short, as highlighted by one of the translators when assessing \textsc{Para} against \textsc{Sent}: \textit{Nightmare, a mistake upon mistake (...) Despite all these mistakes, I can understand the} [\textsc{Para}] \textit{translation better but they are equally miserable.} 
\section{Analyzing translation errors}
\label{sec:analysis}

The aggregate statistics from the previous section confirm that \textsc{Para}-level translation via \textsc{Gpt-3.5} is the strongest literary translator of the methods that we study. Translations produced by \textsc{Para} are favored by both automatic metrics and human translators, and it makes fewer errors than competing methods. In this section, we dive deeper into specific \emph{types} of errors that are made within each high-level category (e.g., grammar, mistranslation), and we present examples of errors associated with lack of context understanding made by \textsc{Sent} and \textsc{GTr} that are fixed by \textsc{Para}.

\subsection{Language-specific grammatical errors}
We begin by analyzing the types of grammatical errors that are made by the studied translation methods in all three target languages.\footnote{There are some differences in the paragraph lengths between the three target languages that should be taken into consideration when analyzing raw numbers. However, the general tendencies remain intact.}

\begin{table}[!t]
\centering
\resizebox{0.48\textwidth}{!}{%
\begin{tabular}{lllrrrr}
\toprule
\textsc{Trg Lang} & \textsc{Type} & \textsc{SubType} &  \textsc{Para} & \textsc{Sents} & \textsc{Para\_Sents} & \textsc{GTr} \\
\midrule
\multirow{9}{*}{\textsc{Japanese}} & \textsc{Particle} & wrong or missing  & 21 & 22 & 13 & 12 \\
 & \textsc{Adjective} & wrong continuative   & 0 & 2 & 3 & 0 \\
 &  & other  & 0 & 0 & 2 & 0 \\
 & \textsc{Verb} & tense & 3 & 7 & 1 & 14 \\
 &  & mood & 2 & 1 & 4 & 5 \\
 &  & finite/non-finite & 5 & 2 & 1 & 3 \\
 &  & other & 2 & 5 & 6 & 0 \\
 & \textsc{Order} & wrong order & 1 & 6 & 1 & 16 \\
 & \textsc{Other} &   & 8 & 5 & 6 & 13 \\
\cmidrule{2-7}
 \rowcolor{violet!20} & \multicolumn{1}{r}{\textsc{Total}} &  & 42 & 50 & 37 & 63 \\
\midrule
\multirow{15}{*}{\textsc{Polish}} & \textsc{Adjective} & gender & 7 & 14 & 8 & 4 \\
 &  & case & 2 & 1 & 1 & 0 \\
 &  & other & 1 & 1 & 1 & 1 \\
 & \textsc{Noun} & case & 9 & 13 & 9 & 1 \\
 &  & other & 3 & 3 & 3 & 2 \\
 & \textsc{Pronoun} & omitted or wrong & 5 & 8 & 3 & 2 \\
 &  & case or gender & 1 & 6 & 4 & 5 \\
 & \textsc{Verb} & aspect & 1 & 5 & 1 & 12 \\
 &  & person or gender & 2 & 8 & 5 & 2 \\
 &  & conjugation & 1 & 0 & 7 & 3 \\
 &  & other  & 2 & 4 & 1 & 13 \\
 & \textsc{Preposition} & omitted or wrong & 14 & 15 & 15 & 4 \\
 & \textsc{Numeral} & case or gender & 2 & 1 & 0 & 1 \\
 & \textsc{Order} & wrong order & 2 & 4 & 2 & 4 \\
 & \textsc{Other} &  & 3 & 3 & 4 & 5 \\
\cmidrule{2-7}
\rowcolor{violet!20} & \multicolumn{1}{r}{\textsc{Total}} & & 55 & 86 & 64 & 59 \\
\midrule
\multirow{4}{*}{\textsc{English}} & \textsc{Article} & omitted or wrong & 1 & 9 & 2 & 8 \\
& \textsc{Preposition} & omitted or wrong & 3 & 7 & 3 & 5 \\
& \textsc{Other} & & 1 & 4 & 4 & 5 \\
\cmidrule{2-7}
\rowcolor{violet!20}& \multicolumn{1}{r}{\textsc{Total}} & & 5 & 20 & 9 & 18 \\
\bottomrule
\end{tabular}
}
\caption{Categorization of grammar errors in each translation configuration, grouped by the target language.}
\label{tab:grammar_errors_categories}
\end{table}

\paragraph{English:}
Perhaps not surprisingly, translations into English contain notably fewer grammatical mistakes than Japanese or Polish (see \autoref{tab:error_count_by_trg_lang}). The most prominent mistakes in English are incorrect articles, which is most frequent in the outputs of \textsc{Sent} and \textsc{GTr}. This is to be expected, as the choice between the definite and indefinite article in English depends heavily on the context. Other mistakes include wrong or omitted prepositions, wrong parts of speech, and incorrect word order (see \autoref{tab:grammar_errors_categories}).

\paragraph{Japanese:} Translations into Japanese contain considerably more mistakes. Most notably, the systems struggle with the correct choice of particle: \textsc{Para} and \textsc{Sent} produce twice as many mistakes in this regard than \textsc{Para\_Sent} and \textsc{GTr} (see \autoref{tab:grammar_errors_categories}). Other mistakes include incorrect tense, verb finite form within the sentence, or incorrect word order, the latter of which is much more frequent in \textsc{GTr} than any of the \textsc{GPT-3.5} translations.

\paragraph{Polish:}
\textsc{GPT-3.5} exhibits more difficulty with Polish, as evidenced by \emph{55} vs \emph{42} errors for \textsc{Para}, \emph{86} vs \emph{50} for \textsc{Sent}, and \emph{64} vs \emph{37} for \textsc{Para\_Sent} (see \autoref{tab:error_count_by_trg_lang}). We notice that \textsc{GPT-3.5} translations frequently generate incorrect gender, case, or prepositions (see \autoref{tab:grammar_errors_categories}). 
We also observe instances in which \textsc{GPT-3.5} alters the gender of a noun, such as producing \textit{grilla}, a non-existent feminine form, in place of the masculine \textit{grill}, while accurately modifying all adjectives and verbs to match the novel feminine noun.\footnote{It is worth noting that \textit{grilla} can also be also the genitive form of the masculine noun \textit{grill}; however, the agreement of surrounding verbs and adjectives with the feminine noun suggests that the system likely treated the word as feminine.}  In contrast, the performance of \textsc{GTr} is comparable for Polish and Japanese in terms of grammar, with \emph{59} and \emph{63} errors respectively. Intriguingly, \textsc{GTr} seems to struggle with Polish aspect, leading to \emph{12} errors, in contrast to \emph{1} error in both \textsc{Para} and \textsc{Para\_Sent}, and \emph{5} errors in \textsc{Sent} within the same category (see \autoref{tab:grammar_errors_categories}).

In summary, although \textsc{GPT-3.5} is primarily trained on English, it is competitive with \textsc{GTr} at Polish and Japanese grammar proficiency. In fact, \textsc{Para} generates the fewest grammatical errors of any system, with a total of \emph{97} for both languages. This is in contrast to \emph{136} errors made by \textsc{Sent}, \emph{101} errors by \textsc{Para\_Sent}, and \emph{122} errors by \textsc{GTr} (see \autoref{tab:error_count_by_trg_lang}). That said, \textit{none} of these systems delivers translations devoid of grammatical inaccuracies, even for English.

\subsection{Context-related errors}
\begin{figure}[!tbp]
\centering
\includegraphics[width=1\columnwidth]{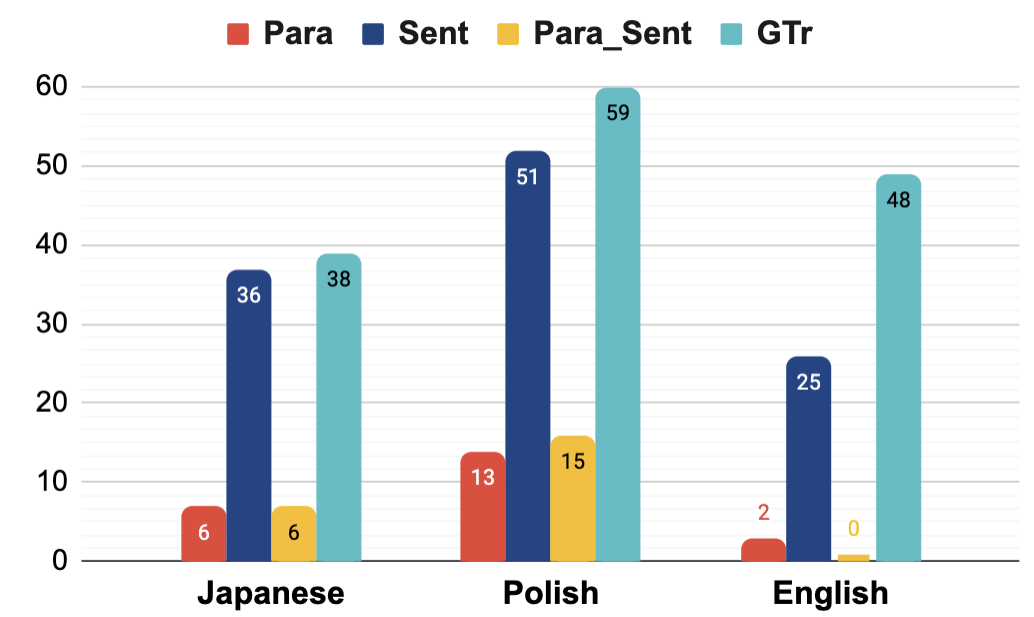}
\caption{Quantification of mistranslations resulting from missing or misinterpreted paragraph-level context in \textsc{Para}, \textsc{Sent}, \textsc{Para\_Sent}, and \textsc{GTr} systems, organized by the target language (Japanese, Polish, and English).}
\label{figure:mistrans_ctx_by_lang}
\end{figure}

We manually classify \textit{all} annotated mistranslations (2,324 instances) into subcategories, several of which  include instances where the absence of discourse-level context is clearly a contributing factor (see \autoref{table:mistrans_type_class_by_trg} for detailed classification). We also further analyze \textit{all} translations in terms of content-related issues. Overall, we observe that context is indeed incorporated into the translations for both \textsc{Para} and \textsc{Para\_Sent} outputs, which results in fewer context-dependent issues (see \autoref{figure:mistrans_ctx_by_lang}).

\paragraph{Pronouns:} Unsurprisingly, the absence of discourse context results in the incorrect translation of pronouns. Consider the following example, with English glosses of important words provided in [brackets]:

\begin{exe}
\small
    \ex\label{pron_rus} \begin{otherlanguage}{russian} И \textcolor{purple}{ветер} \end{otherlanguage} [\textit{wind}] \begin{otherlanguage}{russian}то начинал шуметь в голых деревьях, то замолкал, так же как и я прислушиваясь к течению ночи. Но \textcolor{purple}{он} \end{otherlanguage} [\textit{he}]  \begin{otherlanguage}{russian}не уходил, \textcolor{purple}{он} \end{otherlanguage} [\textit{he}] \begin{otherlanguage}{russian} был здесь.\end{otherlanguage} 
    \begin{flushright}
    \scriptsize
    \vspace{-0.2cm}
    ―\textsc{Russian Source} (from \textit{The Story of a Life})

    \end{flushright}
        \begin{xlist}
            \ex\label{pron_sent}
            The \textcolor{purple}{wind} would start to rustle in the bare trees and then fall silent, just as I listened to the flow of the night. But \textcolor{purple}{he} didn't leave, \textcolor{purple}{he} was here.
            \begin{flushright}
            \scriptsize
            ―\textsc{GPT-3.5 Sent (English)}
            \end{flushright}
            \vspace{0.15cm}
            \ex\label{pron_para} The \textcolor{purple}{wind} would start to rustle in the bare trees, then die down, just like me, listening to the flow of the night. But \textcolor{purple}{it} didn’t go away, \textcolor{purple}{it} was still here.
            \begin{flushright}
            \scriptsize
            ―\textsc{GPT-3.5 Para (English)}
            \end{flushright}
    \end{xlist}
\end{exe}

In Russian, nouns have grammatical gender. ``Wind'' in the first sentence of the source text is a masculine noun, so it is later referred to as ``he'' in (\ref{pron_rus}). Without access to the context, the \textsc{Sent} model incorrectly translates it as ``he'' into English (\ref{pron_sent}), while the \textsc{Para} translation correctly modifies the pronoun to ``it'' (\ref{pron_para}).

When translating from Russian into Polish, another language with grammatical gender, we observe issues when the gender of Russian and Polish nouns differs. Consider the following example:

\begin{exe}
\small
    \ex\label{pron_rus_topl} \begin{otherlanguage}{russian}Романы, как известно, печатались на разной \textcolor{purple}{бумаге} \end{otherlanguage}  [\textit{paper}]. \begin{otherlanguage}{russian}И гореть \textcolor{purple}{она} \end{otherlanguage} [\textit{she}] 
    \begin{otherlanguage}{russian} может по-разному.\end{otherlanguage} 
    \begin{flushright}
    \scriptsize
    \vspace{-0.2cm}
    ―\textsc{Russian Source} (from \textit{Manaraga})

    \end{flushright}
        \begin{xlist}
            \ex\label{pron_sent_topl} Romany, jak wiadomo, drukowano na różnym \textcolor{purple}{papierze} [\textit{paper}]. I może \textcolor{purple}{ona} [\textit{she}] tęsknić na różne sposoby. 
            \begin{flushright}
            \scriptsize
            ―\textsc{GPT-3.5 Sent (Polish)}
            \end{flushright}
            \vspace{0.15cm}
            \ex\label{pron_para_topl} Jak wiadomo, powieści drukowano na różnym \textcolor{purple}{papierze} [\textit{paper}]. I może \textcolor{purple}{on} [\textit{he}] palić się na różne sposoby.
            \begin{flushright}
            \scriptsize
            ―\textsc{GPT-3.5 Para (Polish)}
            \end{flushright}
    \end{xlist}
\end{exe}

Although both Russian and Polish nouns possess grammatical gender, ``Paper'' in (\ref{pron_rus_topl}) is feminine in Russian and referred to as ``she,'' whereas it is a masculine noun in Polish and should be referred to as ``he,'' as in (\ref{pron_para_topl}). The absence of context in \textsc{Sent} leads to an incorrect translation in (\ref{pron_sent_topl}).

\paragraph{Cultural nuances:} Assigning appropriate pronouns without context becomes even more challenging when translating from languages like Japanese, in which speakers frequently refer to the listener (or themselves) in the third person rather than using second-person personal pronouns such as ``you'' in English. 
Consider the following example:

\textcolor{purple}{}
\begin{exe}
\small
    \ex\label{3per_ja} \begin{CJK}{UTF8}{min} 「気が付かなくてすみません」
    
「いやいや、(...)。 
\textcolor{purple}{古倉さんは毎日勤務} なのに手を抜かないからねー！」\end{CJK}

[lit. \textit{Ms./Mrs./Mr. Furukura works every day}]

    \begin{flushright}
    \scriptsize
    \vspace{-0.2cm}
    ―\textsc{Japanese Source} (from \textit{Convenience Store Woman})

    \end{flushright}
        \begin{xlist}
            \ex\label{3per_sent} “I'm sorry I didn't notice.”
            
            “No, no, (...).
            \textcolor{purple}{Furukura-san works hard every day} without taking any shortcuts!”
            \begin{flushright}
            \scriptsize
            ―\textsc{GPT-3.5 Sent (English)}
            \end{flushright}
            \vspace{0.15cm}
            
            \ex\label{3per_para} “I'm sorry I didn't notice.”
            
            “No, no, (...). 
            \textcolor{purple}{You work every day}, but you never slack off!”
            \begin{flushright}
            \scriptsize
            ―\textsc{GPT-3.5 Para (English)}
            \end{flushright}
    \end{xlist}
\end{exe}

From the context of this conversation, a Japanese listener can easily infer that ``Furukura-san'' or ``Miss Furukura''\footnote{Note that the gender of neither character is apparent from the fragment alone.} in the last source sentence (\ref{3per_ja}) is used instead of the second-person ``you'' as per Japanese convention. Translating this sentence without context into English, a language in which third-person reference is not common,\footnote{While third-person reference can be used in English, it is only used in rare circumstances e.g. when addressing children.} results in a confusing translation (\ref{3per_sent}) that implies that the speaker refers to some other ``Furukura'' rather than their listener. However, when translating the sentence in context, the model correctly changes ``Furukura'' into ``you'' (\ref{3per_para}), which makes it clear whom the speaker refers to in English.


\paragraph{Ellipsis:} Another example where context helps is the translation of elliptical constructions. Consider the following example:

\begin{exe}
\small
    \ex\label{elip_cs} 
    
    „Ne, teď \textcolor{purple}{uděláš nádobí}!“ [\textit{(you) will do the dishes!}]
    
    \textcolor{purple}{„Neudělám!“} [\textit{(I) won't do!}]
    
    \textcolor{purple}{„Uděláš!“} [\textit{(You) will do!}]
    \begin{flushright}
    \scriptsize
    \vspace{-0.2cm}
    ―\textsc{Czech Source} (from \textit{Crows})

    \end{flushright}
        \begin{xlist}
            \ex\label{elip_sent} 

            — Nie, teraz \textcolor{purple}{zrobisz zmywanie}! [\textit{(you) will do the washing}]
    
            — \textcolor{purple}{Nie zrobię}! [\textit{(I) won't do!}]
    
            — \textcolor{purple}{Zrobisz to}! [\textit{(You) will do it!}]
            \begin{flushright}
            \scriptsize
            ―\textsc{GPT-3.5 Sent (Polish)}
            \end{flushright}
            \vspace{0.15cm}
            \ex\label{elip_para}    — Nie, teraz \textcolor{purple}{umyjesz naczynia} [\textit{(You) will wash the dishes}]!
    
            — \textcolor{purple}{Nie umyję} [\textit{(I) won't wash}]!
    
             — \textcolor{purple}{Umyjesz} [\textit{(You) will wash}]!
            \begin{flushright}
            \scriptsize
            ―\textsc{GPT-3.5 Para (Polish)}
            \end{flushright}
    \end{xlist}
\end{exe}


Czech uses the same collocation as English, ``\textit{do} the dishes'' (\ref{elip_cs}), which is invalid in Polish. Hence, the ellipses in the last two sentences in (\ref{elip_cs}) require broader context to be translated correctly. \textsc{Para} does it properly, translating both as ``wash'' (\ref{elip_para}), while \textsc{Sent} unsurprisingly fails to choose the correct collocation (\ref{elip_sent}). 

\paragraph{Subject ellipsis:} Similarly, context may be needed to attribute a state or an action to the correct character due to subject ellipsis. This is an obvious issue for languages like Japanese, which tend to omit the subject of the sentence and do not encode any relevant information in the verb form, but it can also arise in English. Consider the following example:

\begin{exe}
\small
    \ex\label{attr_en} When we were done, the lipstick went back into some mother’s Fendi handbag. We watched her apply it, \textcolor{purple}{unaware}. 
    \begin{flushright}
    \scriptsize
    \vspace{-0.2cm}
    ―\textsc{English Source} (from \textit{A Childrens Bible})

    \end{flushright}
        \begin{xlist}
            \ex\label{attr_sent} Gdy skończyliśmy, szminka wróciła do jakiejś torebki Fendi należącej do matki. Patrzyliśmy, jak to robi, \textcolor{purple}{nieświadomi} [\textit{unaware (we)}] tego.  
            \begin{flushright}
            \scriptsize
            ―\textsc{GPT-3.5 Sent (Polish)}
            \end{flushright}
            \vspace{0.15cm}
            \ex\label{attr_para} Kiedy skończyliśmy, szminka wróciła do torebki Fendi jakiejś matki. Patrzyliśmy, jak ją nakłada, \textcolor{purple}{nieświadoma} [\textit{unaware (she)}] naszych działań. 
            \begin{flushright}
            \scriptsize
            ―\textsc{GPT-3.5 Para (Polish)}
            \end{flushright}
    \end{xlist}
\end{exe}

From the second sentence alone it is not clear who is ``unaware'' (\ref{attr_en}) -- the mother or the ``we'' (referring to children) watching her. Only from the broader context can we confidently deduce that it is in fact the mother, not the children, who is ``unaware.'' \textsc{Para} (\ref{attr_para}) correctly attributes the state of being ``unaware'' to the mother, which is exhibited by its usage of the singular feminine form of the adjective. In contrast, \textsc{Sent} (\ref{attr_sent}) mistranslates it using the plural masculine form of the adjective ``unaware,'' which implies that it refers to ``we'' rather than the ``mother.''

\paragraph{Consistency:} Context is sometimes critical for preserving the overall consistency of the text. The simplest cases include referring to the same entity -- a place or a person -- in the same way. More interesting cases pertain to style and can enhance the reader's experience. Consider the following example:

\begin{exe}
\small
    \ex\label{consist_de} Alles zu vergessen, ist gewiss \textcolor{purple}{schlimm} [\textit{bad}]. Noch \textcolor{purple}{schlimmer} [\textit{worse}] ist, nichts zu vergessen (...). 
    \begin{flushright}
    \scriptsize
    \vspace{-0.2cm}
    ―\textsc{German Source} (from \textit{An Inventory of Losses})

    \end{flushright}
        \begin{xlist}
            \ex\label{consist_sent} \begin{CJK}{UTF8}{min}すべてを忘れることは確かに\textcolor{purple}{悲惨な}[\textit{tragic}]ことです。\textcolor{purple}{さらに悪い}[\textit{worse}]のは、何も忘れないことです。\end{CJK}
            \begin{flushright}
            \scriptsize
            ―\textsc{GPT-3.5 Sent (Japanese)}
            \end{flushright}
            \vspace{0.15cm}
            \ex\label{consist_para}  \begin{CJK}{UTF8}{min}すべてを忘れることは確かに\textcolor{purple}{悪い}[\textit{bad}]ことです。\textcolor{purple}{もっと悪い}[\textit{worse}]ことは、何も忘れないことです。\end{CJK}
            \begin{flushright}
            \scriptsize
            ―\textsc{GPT-3.5 Para (Japanese)}
            \end{flushright}
    \end{xlist}
\end{exe}

The German source in (\ref{consist_de}) translates into English as ``To forget everything is \textit{bad}, certainly. \textit{Worse} still is to forget nothing.''\footnote{Excerpt taken from the official English translation by Jakie Smith (2020).} It is arguably important for the translation to repeat the same word which is an equivalent of the German ``schlimm'' (``bad''). \textsc{Para} does it well, translating both as \begin{CJK}{UTF8}{min}悪い\end{CJK} ``warui,'' or ``bad'' (\ref{consist_para}), in the exact same way as the human Japanese translator. \textsc{Sent}, on the other hand, uses two different words, ``tragic'' and ``bad'' (\ref{consist_sent}),  which while technically correct omits the intentional repetition that is meant to introduce an unexpected conclusion.

\paragraph{Polysemy:} The absence of context makes it difficult to interpret words or expressions that have multiple meanings in the source language. Consider the following example:

\begin{exe}
\small
    \ex\label{multi_mean_rus} \begin{otherlanguage}{russian}Все прошло хорошо. Книга прочитана идеально – не быстро и не медленно, минимум дыма. Классика. Я был \textcolor{purple}{в форме} \end{otherlanguage} [\textit{in shape}].
    \begin{flushright}
    \scriptsize
    \vspace{-0.2cm}
    ―\textsc{Russian Source} (from \textit{Maranaga})

    \end{flushright}
        \begin{xlist}
            \ex\label{multi_mean_sent}  Wszystko poszło dobrze. Książka została przeczytana idealnie – nie szybko i nie wolno, minimalna ilość dymu. Klasyka. Byłem \textcolor{purple}{w mundurze} [\textit{in uniform}]. 
            \begin{flushright}
            \scriptsize
            ―\textsc{GPT-3.5 Sent (Polish)}
            \end{flushright}
            \vspace{0.15cm}
            \ex\label{multi_mean_para}  Wszystko poszło dobrze. Książka przeczytana idealnie – nie szybko i nie wolno, minimalna ilość dymu. Klasyka. Byłem \textcolor{purple}{w formie} [\textit{in shape}].
            \begin{flushright}
            \scriptsize
            ―\textsc{GPT-3.5 Para (Polish)}
            \end{flushright}
    \end{xlist}
\end{exe}

The ambiguity stems here from multiple meanings of the Russian noun \begin{otherlanguage}{russian}форма \end{otherlanguage} ``forma'' (\ref{multi_mean_rus}), which can mean  either ``shape'' or ``uniform.'' Since one can be ``in shape'' as well as ``in a uniform'', it is unclear from the sentence alone which meaning was intended by the author. From the preceding context, it is clear that ``everything went well'' for the narrator, who mastered the art of ``book’n’grill,'' a unique form of expression exclusive to this fictional world. Based on this, we can infer that in this instance, the term ``forma'' signifies ``shape,'' as in (\ref{multi_mean_para}), rather than ``uniform,'' as in (\ref{multi_mean_sent}).

\paragraph{Appropriateness:} Finally, context may help to choose the more appropriate equivalent for the given situation. Consider the following example:

\begin{exe}
\small
    \ex\label{prag_ja}  \begin{CJK}{UTF8}{min}
    「あー、あと煙草の５番を一つ」
    
    \textcolor{purple}{「かしこまりました」} 
    \end{CJK} [lit. \textit{(I) understood}] 
    \begin{flushright}
    \scriptsize
    \vspace{-0.2cm}
    ―\textsc{Japanese Source} (from \textit{Convenience Store Woman})

    \end{flushright}
        \begin{xlist}
    
            \ex\label{prag_en_sent}  "Ah, and one pack of cigarettes, number five."
    
            \textcolor{purple}{"Understood."}
            \begin{flushright}
            \scriptsize
            ―\textsc{GPT-3.5 Sent (English)}
            \end{flushright}
            \vspace{0.15cm}
    
             \ex\label{prag_en_para}  “Ah, and one pack of cigarettes, number five.”
    
            \textcolor{purple}{“Right away.”}
            \begin{flushright}
            \scriptsize
            ―\textsc{GPT-3.5 Para (English)}
            \end{flushright}
    \end{xlist}
\end{exe}

The conversation above is between a clerk and a customer. The Japanese expression \begin{CJK}{UTF8}{min}かしこまりました\end{CJK} ``kashikomarimashita'' (\ref{prag_ja}) is an honorific that literally means ``understood.'' However, when choosing the best equivalent, the translator needs to consider the situation at hand to best reflect its meaning in the target language. ``Understood'' in \textsc{Sent} (\ref{prag_en_sent}) is technically correct, but it is an unfortunate word choice for the clerk to employ. On the other hand, ``right away'' in \textsc{Para} (\ref{prag_en_para}) fits much better in the context of this conversation. Had this been a series of commands (e.g., in a military context) ``understood'' would be the more favorable option.


\section{Limitations}
\label{sec:limitations}

So far, we have shown that \textsc{GPT-3.5} leverages paragraph-level context to produce translations that are better than those produced by sentence-level counterparts (\textsc{Sent} vs \textsc{Para}). However, there are still many issues with \textsc{Para}'s translations. From the annotations and translators' comments, we observe that \textsc{Para} suffers from occasional omissions of content from the source paragraph. \textsc{Sent} and \textsc{GTr} are certainly not free of that problem either, but omission appears to be more prominent for \textsc{Para} translations (see \autoref{app:hum_eval}).

Moreover, 
\textsc{Para} still makes a sizeable number of mistranslations and grammatical errors, though fewer than \textsc{Sent} or \textsc{GTr}. We observe that \textsc{Para} occasionally merges sentences with two distinctive subjects attributing all states and/or actions to one of them. 
Very rarely, we also find cases where context possibly confuses the model, resulting in an incorrect translation. The following example illustrates this issue: 

\begin{exe}
\small
    \ex\label{conf_fr} Le bois du bureau amplifie les battements de mon cœur. Le vieux mobilier Art déco conduit bien les émotions et les fatigues. Ruhlman ? Leleu ? \textcolor{purple}{Il} [\textit{he}] en a tant vu.
    \begin{flushright}
    \scriptsize
    \vspace{-0.2cm}
    ―\textsc{French Source} (from \textit{Dear Reader})
    \end{flushright}
        \begin{xlist}
            \ex\label{conf_para} \begin{CJK}{UTF8}{min} 机の木材が私の心臓の鼓動を増幅している。古いアール・デコ家具は感情や疲労をうまく導いてくれる。ルールマン？レルー？\textcolor{purple}{彼ら} [\textit{they}] はそんなに多くを見てきた。
            \end{CJK} 
            \begin{flushright}
            \scriptsize
            ―\textsc{GPT-3.5 Para (Japanese)}
            \end{flushright}
            \vspace{0.15cm}
    \end{xlist}
\end{exe}

In the French text, the narrator wonders whether the brand of the desk was Ruhlman or Leleu, with both proper nouns possibly referring to a person. In the last sentence, the French text uses ``il'' or ``he'' (\ref{conf_fr}), as a desk is a masculine noun in French (``le bureau''). \textsc{Para}, on the other hand, appears to be confused by the two preceding names and incorrectly translates the singular pronoun as \begin{CJK}{UTF8}{min}彼ら\end{CJK}, or ``they.''

Furthermore, we observe (very few) cases where the paragraph-level translation disregards the context. Most representative of this class of errors is when the model struggles to translate from Japanese in cases where the subject is omitted. The following example illustrates this issue:

\begin{exe}
\small
    \ex\label{ctx_ingnored_ja} \begin{CJK}{UTF8}{min}
    \textcolor{purple}{ミホ} [\textit{Miho}] は、今では結婚して地元に中古の一戸建てを買っていて、そこに友達がよく集まっている。明日もアルバイトなので億劫に思う時もあるが、コンビニ以外の世界との唯一の接点であり、同い年の「普通の三十代女性」と交流する貴重な機会なので、\textcolor{blue}{ミホの} [\textit{Miho's}] 誘いにはなるべく応じるようにしている。
    \end{CJK} 
    \begin{flushright}
    \scriptsize
    \vspace{-0.2cm}
    ―\textsc{Japanese Source} (from \textit{Convenience Store Woman})

    \end{flushright}
        \begin{xlist}
            \ex\label{ctx_ingnored_pl} \textcolor{purple}{Miho} [\textit{Miho}] wyszła za mąż i kupiła stary, jednorodzinny dom w swoim rodzinnym mieście. Przychodzą tam często jej znajomi. Mimo że \textcolor{purple}{Miho ma} [\textit{Miho has}] jutro pracę w konbini, zazwyczaj chętnie \textcolor{purple}{odpowiada} [\textit{(she) responds}] na \textcolor{blue}{jej} [\textit{her}] zaproszenia, bo to jedyna okazja, by spotkać się z innymi kobietami w \textcolor{purple}{jej} [\textit{her}] wieku.   
            \begin{flushright}
            \scriptsize
            ―\textsc{GPT-3.5 Para (Polish)}
            \end{flushright}
            \vspace{0.15cm}
            \ex\label{ctx_ingnored_en} \textcolor{purple}{Miho} is now married and has bought an old house in her hometown, where her friends often gather. Though \textcolor{purple}{she} often finds it a chore to work tomorrow, it is \textcolor{purple}{her} only connection to the world outside the convenience store, and a valuable opportunity to interact with other “normal thirty-something women” \textcolor{purple}{her} age, so \textcolor{purple}{she} tries to accept \textcolor{blue}{Miho’s} invitations as often as possible. 
            \begin{flushright}
            \scriptsize
            ―\textsc{GPT-3.5 Para (English)}
            \end{flushright}
    \end{xlist}
\end{exe}

Both Polish (\ref{ctx_ingnored_pl}) and English (\ref{ctx_ingnored_en}) translations of the same source text (\ref{ctx_ingnored_ja}) share a common issue. The narrator begins the paragraph by talking about Miho and then proceeds to describe her own (the narrator's) feelings about the situation, although the gender of the narrator is never revealed in the Japanese text. The second sentence should be written from a first-person perspective, particularly since it directly references Miho towards the end (\textcolor{blue}{blue text}). However, both the Polish and English translations produced by \textsc{Para} are confused by this: by using the third-person's perspective (``she,'' ``her''), both translations incorrectly imply that Miho is the subject of the second sentence.
\textsc{Sent} and \textsc{GTr} translate this passage accurately, albeit with some clumsy phrasing.

Finally, it is important to acknowledge that the languages covered in the current study are either mid or high-resource. Performance might be much worse when translating from or into one of the low-resource languages, such as Zulu or Armenian.



\paragraph{GPT-4 does not magically solve all of these issues!}
 Our preliminary experiments indicate that \textsc{GPT-4} \cite{openai2023gpt4} sometimes generates better paragraph-level translations than those of \textsc{GPT-3.5}. For instance, it seems to have a better grasp of the inverted word order in German, though no broader conclusions should be made without further testing. Nevertheless, it does not resolve all of the issues discussed in our paper. Mistranslations and grammatical errors are still abundant across many language pairs. GPT-4 produces the following translation when fed the previous example paragraph (\ref{ctx_ingnored_ja}) as input; note that all of the issues still remain:\footnote{Although the given paragraph is already comprehensible for a human reader, we also attempt to enhance the translation by incorporating three additional preceding paragraphs for context. Intriguingly, when provided with this extended context, both \textsc{GPT-3.5} and \textsc{GPT-4} generated accurate translations.}


\begin{exe}
\small
            \ex \textcolor{purple}{Miho} is now married and has bought a used single-family home in her hometown where her friends often gather. Although \textcolor{purple}{she} sometimes finds it a drag to work a part-time job the next day, \textcolor{purple}{she} makes an effort to respond to \textcolor{blue}{Miho's} invitations because it's a valuable opportunity to interact with ``normal'' women in their thirties like \textcolor{purple}{herself}, apart from \textcolor{purple}{her} convenience store job.
            \vspace{-0.2cm}
            \begin{flushright}
            \scriptsize
            ―\textsc{GPT-4 Para (English)}
            \end{flushright}
\end{exe}

\textsc{Para} translations hold the potential to captivate readers, especially if LLMs continue to improve at their current pace. Indeed, some of our translators mentioned that they genuinely enjoyed the task, though integrating these paragraphs into a coherent novel still poses a considerable challenge. With all that said, literary translation involves more than just overall ``correctness'' or mere entertainment value. A translation that is perfectly ``correct'' and enjoyable might still fail to convey the author's intentions or meaning skillfully hidden behind a simple phrase. Our \textit{fr-en} translator shares her thoughts on this matter:

\begin{quote}
    \footnotesize
    \textit{Both translations} [\textsc{Sent} and \textsc{Para}] \textit{translate the words without the feeling; the original author's voice is lost.}
    \begin{flushright}
    \scriptsize
    \vspace{-0.4cm}
    ―\textsc{French to English Translator}
    \end{flushright}
\end{quote}

\section{Conclusion}

In this paper, we demonstrate that LLMs leverage paragraph-level context to produce translations that are more coherent and enjoyable than sentence-by-sentence translation while containing fewer mistranslations and grammatical issues. Our evaluations reveal that professional translators prefer paragraph-level translations over both sentence-level translations produced by the same language model, and also to those generated by an off-the-shelf commercial system (\textsc{GTr}). We release our dataset and error annotations to help facilitate the development of new evaluation methodologies and automatic metrics for document-level machine translation.
Finally, a full-length novel extends far beyond the confines of paragraph-level translation. In future work, we will focus on integrating individual paragraphs into cohesive chapters, which can then be expanded to encompass the entire novel.
\section*{Ethical considerations}

\textbf{Translating with LLMs:} The rise of large language models has also brought many ethical concerns to the forefront of NLP research~\cite{blodgett-etal-2020-language,bender-etal-2021-parrots}. LLMs encode biases and exhibit toxicity, and these behaviors can be exacerbated by unconstrained prompting \cite{gehman-etal-2020-realtoxicityprompts, costajussa2022toxicity}. Further ethical concerns arise in the context of machine translation, particularly \emph{literary} translation, where multiple stakeholders -- the author, the translator, and the audience -- are involved \cite{ethics_trans_2019}. Low-quality output can influence the perception of the author's work, impair the reader's linguistic abilities, and hinder the transfer of ideas to the target language, while overrelying on machine translation can possibly threaten the role of human translators  \cite{Drugan2013-rb, Ning2016ComparativeLA, ethics_trans_2019}. On the other hand, machine translation employed \emph{responsibly} as an auxiliary tool holds the potential to alleviate the translator's cognitive burden \cite{OBrien2012} and make the author's work accessible to a broader audience more swiftly \cite{besacier-2014-machine}. Contrary to the predictions in \citet{eloundou2023gpts_market}, we do not view large language models as a substitute for human translators, but rather as a means to assist translators in their work. 

\textbf{Human Evaluation:} The experiments involving human translators were IRB-approved, and all involved translators gave their consent to disclose their annotations, comments, and preference choices. In recognizing contributions, our acknowledgments only include the names of those translators who explicitly gave their consent to be acknowledged by their full name in this publication.

\section*{Acknowledgements}

First and foremost, we would like to express our gratitude to the translators hired mostly on Upwork: Malgorzata Szymczak (\textit{fr-pl}), Kinga Przekota (\textit{ru-pl}), Michal Sikora (\textit{cs-pl}), Paula Kurzawska (\textit{de-pl}, \textit{de-en}, \textit{pl-en}), Kristy Darling Finder (\textit{fr-en}), Timothy Shostak (\textit{ja-en}), Shun Enoki (\textit{zh-ja}), Takanori Kurokawa (\textit{fr-ja}), Yoshiko Kikawa (\textit{en-ja}), Shinnosuke Kasahara (\textit{ru-ja}), and all those who wish to remain anonymous. We encourage any machine translation researchers working on these language pairs to contact these translators for human evaluations. 

We would also like to show our appreciation to Jan Wislicki, Tom Gally, Nader Akoury, Kalpesh Krishna, Simeng Sun, Katherine Thai, and the entire UMass NLP group for insightful discussion, which helped to shape this project. 

Finally, we would like to thank Sergiusz Rzepkowski (\textit{pl}), Paula Kurzawska (\textit{pl, en}), Hiroshi Iida (\textit{ja}), Grégory Fleurot (\textit{fr}), Peyton Bowman (\textit{en}), Simeng Sun (\textit{zh}), Igor Zapala (\textit{pl, de}), Marvin Hoffmann (\textit{de}), Kinga Przekota (\textit{pl, ru}), and Yuki Mori (\textit{ja}) for further consultations on their respective native languages.

This project was partially supported by awards IIS-1955567 and IIS-2046248 from the National Science Foundation (NSF) as well as an award from Open Philanthropy.

\bibliography{anthology,custom}

\begin{thebibliography}{74}
\expandafter\ifx\csname natexlab\endcsname\relax\def\natexlab#1{#1}\fi

\bibitem[{Agrawal et~al.(2018)Agrawal, Turchi, and
  Negri}]{Agrawal2018ContextualHI}
Ruchit Agrawal, Marco Turchi, and Matteo Negri. 2018.
\newblock \href {http://hdl.handle.net/10045/76016} {Contextual {H}andling in
  {N}eural machine {T}ranslation: {L}ook {B}ehind, {A}head and on {B}oth
  {S}ides}.
\newblock In \emph{21st Annual Conference of the European Association for
  Machine Translation}, pages 11--20.

\bibitem[{Aiyappa et~al.(2023)Aiyappa, An, Kwak, and Ahn}]{aiyappa2023trust}
Rachith Aiyappa, Jisun An, Haewoon Kwak, and Yong-Yeol Ahn. 2023.
\newblock \href {http://arxiv.org/abs/2303.12767} {Can we trust the evaluation
  on {C}hat{GPT}?}

\bibitem[{Baayen et~al.(2008)Baayen, Davidson, and Bates}]{Baayen2008_lme}
R.H. Baayen, D.J. Davidson, and D.M. Bates. 2008.
\newblock \href {https://doi.org/10.1016/j.jml.2007.12.005} {Mixed-effects
  modeling with crossed random effects for subjects and items}.
\newblock \emph{Journal of Memory and Language}, 59(4):390--412.

\bibitem[{Bates et~al.(2015)Bates, Mächler, Bolker, and Walker}]{lme4}
Douglas Bates, Martin Mächler, Ben Bolker, and Steve Walker. 2015.
\newblock \href {https://doi.org/10.18637/jss.v067.i01} {Fitting {L}inear
  {M}ixed-{E}ffects {M}odels {U}sing lme4}.
\newblock \emph{Journal of Statistical Software}, 67(1):1–48.

\bibitem[{Bender et~al.(2021)Bender, Gebru, McMillan-Major, and
  Shmitchell}]{bender-etal-2021-parrots}
Emily~M. Bender, Timnit Gebru, Angelina McMillan-Major, and Shmargaret
  Shmitchell. 2021.
\newblock \href {https://doi.org/10.1145/3442188.3445922} {On the {D}angers of
  {S}tochastic {P}arrots: {C}an {L}anguage {M}odels {B}e {T}oo {B}ig?}
\newblock In \emph{Proceedings of the 2021 ACM Conference on Fairness,
  Accountability, and Transparency}, FAccT '21, page 610–623, New York, NY,
  USA. Association for Computing Machinery.

\bibitem[{Besacier(2014)}]{besacier-2014-machine}
Laurent Besacier. 2014.
\newblock \href {https://aclanthology.org/F14-2001} {Machine translation for
  litterature: a pilot study (traduction automatis{\'e}e d{'}une oeuvre
  litt{\'e}raire: une {\'e}tude pilote) [in {F}rench]}.
\newblock In \emph{Proceedings of TALN 2014 (Volume 2: Short Papers)}, pages
  389--394, Marseille, France. Association pour le Traitement Automatique des
  Langues.

\bibitem[{Blodgett et~al.(2020)Blodgett, Barocas, Daum{\'e}~III, and
  Wallach}]{blodgett-etal-2020-language}
Su~Lin Blodgett, Solon Barocas, Hal Daum{\'e}~III, and Hanna Wallach. 2020.
\newblock \href {https://doi.org/10.18653/v1/2020.acl-main.485} {Language
  ({T}echnology) is {P}ower: {A} {C}ritical {S}urvey of {``}{B}ias{''} in
  {NLP}}.
\newblock In \emph{Proceedings of the 58th Annual Meeting of the Association
  for Computational Linguistics}, pages 5454--5476, Online. Association for
  Computational Linguistics.

\bibitem[{{B}ook {M}aker(2023)}]{bilingual_book_maker}
Bilingual {B}ook {M}aker. 2023.
\newblock {M}ake bilingual epub books {U}sing {AI} translate ({G}it{H}ub).
\newblock \url{https://github.com/yihong0618/bilingual_book_maker}.
\newblock [Accessed 05-Apr-2023].

\bibitem[{Brown et~al.(2020)Brown, Mann, Ryder, Subbiah, Kaplan, Dhariwal,
  Neelakantan, Shyam, Sastry, Askell et~al.}]{brown2020language}
Tom Brown, Benjamin Mann, Nick Ryder, Melanie Subbiah, Jared~D Kaplan, Prafulla
  Dhariwal, Arvind Neelakantan, Pranav Shyam, Girish Sastry, Amanda Askell,
  et~al. 2020.
\newblock \href
  {https://proceedings.neurips.cc/paper/2020/file/1457c0d6bfcb4967418bfb8ac142f64a-Paper.pdf}
  {Language models are few-shot learners}.
\newblock \emph{Advances in neural information processing systems},
  33:1877--1901.

\bibitem[{Carpuat and Simard(2012)}]{carpuat-simard-2012-trouble}
Marine Carpuat and Michel Simard. 2012.
\newblock \href {https://aclanthology.org/W12-3156} {{T}he {T}rouble with {SMT}
  {C}onsistency}.
\newblock In \emph{Proceedings of the Seventh Workshop on Statistical Machine
  Translation}, pages 442--449, Montr{\'e}al, Canada. Association for
  Computational Linguistics.

\bibitem[{Castilho(2021)}]{castilho-2021-towards}
Sheila Castilho. 2021.
\newblock \href {https://aclanthology.org/2021.humeval-1.4} {Towards
  {D}ocument-{L}evel human {MT} {E}valuation: {O}n the {I}ssues of {A}nnotator
  {A}greement, {E}ffort and {M}isevaluation}.
\newblock In \emph{Proceedings of the Workshop on Human Evaluation of NLP
  Systems (HumEval)}, pages 34--45, Online. Association for Computational
  Linguistics.

\bibitem[{Chen et~al.(2020)Chen, Li, Zhang, Zhou, Cui, Wang, and
  Su}]{chen-etal-2020-modeling}
Junxuan Chen, Xiang Li, Jiarui Zhang, Chulun Zhou, Jianwei Cui, Bin Wang, and
  Jinsong Su. 2020.
\newblock \href {https://doi.org/10.18653/v1/2020.autosimtrans-1.5} {Modeling
  {D}iscourse {S}tructure for {D}ocument-level {N}eural {M}achine
  {T}ranslation}.
\newblock In \emph{Proceedings of the First Workshop on Automatic Simultaneous
  Translation}, pages 30--36, Seattle, Washington. Association for
  Computational Linguistics.

\bibitem[{Chesterman(1997)}]{Chesterman1997-memes}
Andrew Chesterman. 1997.
\newblock \href {https://doi.org/10.1075/btl.22} {\emph{{M}emes of
  {T}ranslation}}.
\newblock Benjamins Translation Library. Benjamins (John) North America,
  Amsterdam, Netherlands.

\bibitem[{Cohn and Lapata(2007)}]{cohn-lapata-2007-machine}
Trevor Cohn and Mirella Lapata. 2007.
\newblock \href {https://aclanthology.org/P07-1092} {Machine translation by
  triangulation: Making effective use of multi-parallel corpora}.
\newblock In \emph{Proceedings of the 45th Annual Meeting of the Association of
  Computational Linguistics}, pages 728--735, Prague, Czech Republic.
  Association for Computational Linguistics.

\bibitem[{Costa-jussà et~al.(2022)Costa-jussà, Smith, Ropers, Licht,
  Ferrando, and Escolano}]{costajussa2022toxicity}
Marta~R. Costa-jussà, Eric Smith, Christophe Ropers, Daniel Licht, Javier
  Ferrando, and Carlos Escolano. 2022.
\newblock \href {http://arxiv.org/abs/2210.03070} {Toxicity in {M}ultilingual
  {M}achine {T}ranslation at {S}cale}.

\bibitem[{Ding et~al.(2014)Ding, Utiyama, and Sumita}]{ding-etal-2014-document}
Chenchen Ding, Masao Utiyama, and Eiichiro Sumita. 2014.
\newblock \href {https://aclanthology.org/2014.amta-researchers.9}
  {Document-level re-ranking with soft lexical and semantic features for
  statistical machine translation}.
\newblock In \emph{Proceedings of the 11th Conference of the Association for
  Machine Translation in the Americas: MT Researchers Track}, pages 110--123,
  Vancouver, Canada. Association for Machine Translation in the Americas.

\bibitem[{Drugan(2013)}]{Drugan2013-rb}
Joanna Drugan. 2013.
\newblock \href
  {https://www.bloomsbury.com/us/quality-in-professional-translation-9781441149541/}
  {\emph{Quality in professional translation}}.
\newblock Bloomsbury Advances in Translation. Continuum Publishing Corporation,
  New York, NY.

\bibitem[{Eloundou et~al.(2023)Eloundou, Manning, Mishkin, and
  Rock}]{eloundou2023gpts_market}
Tyna Eloundou, Sam Manning, Pamela Mishkin, and Daniel Rock. 2023.
\newblock \href {http://arxiv.org/abs/2303.10130} {{GPT}s are {GPT}s: {A}n
  {E}arly {L}ook at the {L}abor {M}arket {I}mpact {P}otential of {L}arge
  {L}anguage {M}odels}.

\bibitem[{Feng et~al.(2022)Feng, Li, Song, Zheng, and
  Koehn}]{feng-etal-2022-learn}
Yukun Feng, Feng Li, Ziang Song, Boyuan Zheng, and Philipp Koehn. 2022.
\newblock \href {https://doi.org/10.18653/v1/2022.findings-naacl.105} {Learn to
  {R}emember: {T}ransformer with {R}ecurrent {M}emory for {D}ocument-level
  {M}achine {T}ranslation}.
\newblock In \emph{Findings of the Association for Computational Linguistics:
  NAACL 2022}, pages 1409--1420, Seattle, United States. Association for
  Computational Linguistics.

\bibitem[{Freitag et~al.(2021)Freitag, Foster, Grangier, Ratnakar, Tan, and
  Macherey}]{10.1162/tacl_a_00437}
Markus Freitag, George Foster, David Grangier, Viresh Ratnakar, Qijun Tan, and
  Wolfgang Macherey. 2021.
\newblock \href {https://doi.org/10.1162/tacl_a_00437} {{Experts, Errors, and
  Context: A Large-Scale Study of Human Evaluation for Machine Translation}}.
\newblock \emph{Transactions of the Association for Computational Linguistics},
  9:1460--1474.

\bibitem[{Gehman et~al.(2020)Gehman, Gururangan, Sap, Choi, and
  Smith}]{gehman-etal-2020-realtoxicityprompts}
Samuel Gehman, Suchin Gururangan, Maarten Sap, Yejin Choi, and Noah~A. Smith.
  2020.
\newblock \href {https://doi.org/10.18653/v1/2020.findings-emnlp.301}
  {{R}eal{T}oxicity{P}rompts: {E}valuating {N}eural {T}oxic {D}egeneration in
  {L}anguage {M}odels}.
\newblock In \emph{Findings of the Association for Computational Linguistics:
  EMNLP 2020}, pages 3356--3369, Online. Association for Computational
  Linguistics.

\bibitem[{Gehrmann et~al.(2022)Gehrmann, Clark, and
  Sellam}]{gehrmann2022repairing}
Sebastian Gehrmann, Elizabeth Clark, and Thibault Sellam. 2022.
\newblock \href {http://arxiv.org/abs/2202.06935} {Repairing the {C}racked
  {F}oundation: {A} {S}urvey of {O}bstacles in {E}valuation {P}ractices for
  {G}enerated {T}ext}.

\bibitem[{Ghazvininejad et~al.(2023)Ghazvininejad, Gonen, and
  Zettlemoyer}]{ghazvininejad2023dictionarybased}
Marjan Ghazvininejad, Hila Gonen, and Luke Zettlemoyer. 2023.
\newblock \href {http://arxiv.org/abs/2302.07856} {{D}ictionary-based
  {P}hrase-level {P}rompting of {L}arge {L}anguage {M}odels for {M}achine
  {T}ranslation}.

\bibitem[{Guerreiro et~al.(2023)Guerreiro, Alves, Waldendorf, Haddow, Birch,
  Colombo, and Martins}]{guerreiro2023hallucinations}
Nuno~M. Guerreiro, Duarte Alves, Jonas Waldendorf, Barry Haddow, Alexandra
  Birch, Pierre Colombo, and André F.~T. Martins. 2023.
\newblock \href {http://arxiv.org/abs/2303.16104} {Hallucinations in {L}arge
  {M}ultilingual {T}ranslation {M}odels}.

\bibitem[{Han(2020)}]{tqa_han_2020}
Chao Han. 2020.
\newblock \href {https://doi.org/10.1080/13556509.2020.1834751} {Translation
  quality assessment: a critical methodological review}.
\newblock \emph{The Translator}, 26(3):257--273.

\bibitem[{Hardmeier(2012)}]{Hardmeier2012}
Christian Hardmeier. 2012.
\newblock \href {https://doi.org/10.4000/discours.8726} {{D}iscourse in
  {S}tatistical {M}achine {T}ranslation}.
\newblock \emph{Discours}, (11).

\bibitem[{Hardmeier et~al.(2013)Hardmeier, Stymne, Tiedemann, and
  Nivre}]{hardmeier-etal-2013-docent}
Christian Hardmeier, Sara Stymne, J{\"o}rg Tiedemann, and Joakim Nivre. 2013.
\newblock \href {https://aclanthology.org/P13-4033} {{D}ocent: {A}
  {D}ocument-level {D}ecoder for {P}hrase-{B}ased {S}tatistical {M}achine
  {T}ranslation}.
\newblock In \emph{Proceedings of the 51st Annual Meeting of the Association
  for Computational Linguistics: System Demonstrations}, pages 193--198, Sofia,
  Bulgaria. Association for Computational Linguistics.

\bibitem[{Hendy et~al.(2023)Hendy, Abdelrehim, Sharaf, Raunak, Gabr,
  Matsushita, Kim, Afify, and Awadalla}]{hendy2023good}
Amr Hendy, Mohamed Abdelrehim, Amr Sharaf, Vikas Raunak, Mohamed Gabr, Hitokazu
  Matsushita, Young~Jin Kim, Mohamed Afify, and Hany~Hassan Awadalla. 2023.
\newblock \href {http://arxiv.org/abs/2302.09210} {How {G}ood {A}re {GPT}
  {M}odels at {M}achine {T}ranslation? {A} {C}omprehensive {E}valuation}.

\bibitem[{J{\'{a}}grov{\'{a}} and Avgustinova(2023)}]{Jgrov2023czech}
Kl{\'{a}}ra J{\'{a}}grov{\'{a}} and Tania Avgustinova. 2023.
\newblock \href {https://doi.org/10.1007/978-3-031-24337-0_9} {Intelligibility
  of highly predictable polish target words in sentences presented to czech
  readers}.
\newblock In \emph{Computational Linguistics and Intelligent Text Processing},
  pages 110--125. Springer Nature Switzerland.

\bibitem[{Jean et~al.(2017)Jean, Lauly, Firat, and Cho}]{jean2017does}
Sebastien Jean, Stanislas Lauly, Orhan Firat, and Kyunghyun Cho. 2017.
\newblock \href {http://arxiv.org/abs/1704.05135} {Does neural machine
  translation benefit from larger context?}

\bibitem[{Jiang et~al.(2022)Jiang, Liu, Ma, Zhang, Yang, Huang, Sennrich,
  Cotterell, Sachan, and Zhou}]{jiang-etal-2022-blonde}
Yuchen Jiang, Tianyu Liu, Shuming Ma, Dongdong Zhang, Jian Yang, Haoyang Huang,
  Rico Sennrich, Ryan Cotterell, Mrinmaya Sachan, and Ming Zhou. 2022.
\newblock \href {https://doi.org/10.18653/v1/2022.naacl-main.111} {{BlonDe}:
  {A}n {A}utomatic {E}valuation {M}etric for {D}ocument-level {M}achine
  {T}ranslation}.
\newblock In \emph{Proceedings of the 2022 Conference of the North American
  Chapter of the Association for Computational Linguistics: Human Language
  Technologies}, pages 1550--1565, Seattle, United States. Association for
  Computational Linguistics.

\bibitem[{Jiao et~al.(2023)Jiao, Wang, tse Huang, Wang, and
  Tu}]{jiao2023chatgpt}
Wenxiang Jiao, Wenxuan Wang, Jen tse Huang, Xing Wang, and Zhaopeng Tu. 2023.
\newblock \href {http://arxiv.org/abs/2301.08745} {Is {C}hat{GPT} {A} {G}ood
  {T}ranslator? {Y}es with {GPT}-4 {A}s {T}he {E}ngine}.

\bibitem[{Junczys-Dowmunt(2019)}]{junczys-dowmunt-2019-microsoft}
Marcin Junczys-Dowmunt. 2019.
\newblock \href {https://doi.org/10.18653/v1/W19-5321} {{M}icrosoft translator
  at {WMT} 2019: Towards large-scale document-level neural machine
  translation}.
\newblock In \emph{Proceedings of the Fourth Conference on Machine Translation
  (Volume 2: Shared Task Papers, Day 1)}, pages 225--233, Florence, Italy.
  Association for Computational Linguistics.

\bibitem[{Kang et~al.(2020)Kang, Zhao, Zhang, and
  Zong}]{kang-etal-2020-dynamic}
Xiaomian Kang, Yang Zhao, Jiajun Zhang, and Chengqing Zong. 2020.
\newblock \href {https://doi.org/10.18653/v1/2020.emnlp-main.175} {Dynamic
  context selection for document-level neural machine translation via
  reinforcement learning}.
\newblock In \emph{Proceedings of the 2020 Conference on Empirical Methods in
  Natural Language Processing (EMNLP)}, pages 2242--2254, Online. Association
  for Computational Linguistics.

\bibitem[{Kaplansky(2004)}]{Kaplansky2004stranger}
Jonathan Kaplansky. 2004.
\newblock \href {https://doi.org/10.4000/palimpsestes.1583} {{O}utside {T}he
  {S}tranger? {E}nglish {R}etranslations of {C}amus'}.
\newblock \emph{Palimpsestes}, (15):187--198.

\bibitem[{Karpinska et~al.(2021)Karpinska, Akoury, and
  Iyyer}]{karpinska-etal-2021-perils}
Marzena Karpinska, Nader Akoury, and Mohit Iyyer. 2021.
\newblock \href {https://doi.org/10.18653/v1/2021.emnlp-main.97} {The {P}erils
  of {U}sing {M}echanical {T}urk to {E}valuate {O}pen-ended {T}ext
  {G}eneration}.
\newblock In \emph{Proceedings of the 2021 Conference on Empirical Methods in
  Natural Language Processing}, pages 1265--1285, Online and Punta Cana,
  Dominican Republic. Association for Computational Linguistics.

\bibitem[{Karpinska et~al.(2022)Karpinska, Raj, Thai, Song, Gupta, and
  Iyyer}]{karpinska-etal-2022-demetr}
Marzena Karpinska, Nishant Raj, Katherine Thai, Yixiao Song, Ankita Gupta, and
  Mohit Iyyer. 2022.
\newblock \href {https://aclanthology.org/2022.emnlp-main.649} {{DEMETR}:
  {D}iagnosing {E}valuation {M}etrics for {T}ranslation}.
\newblock In \emph{Proceedings of the 2022 Conference on Empirical Methods in
  Natural Language Processing}, pages 9540--9561, Abu Dhabi, United Arab
  Emirates. Association for Computational Linguistics.

\bibitem[{Kocmi and Federmann(2023)}]{kocmi2023large}
Tom Kocmi and Christian Federmann. 2023.
\newblock \href {http://arxiv.org/abs/2302.14520} {Large {L}anguage {M}odels
  {A}re {S}tate-of-the-{A}rt {E}valuators of {T}ranslation {Q}uality}.

\bibitem[{Kuznetsova et~al.(2017)Kuznetsova, Brockhoff, and
  Christensen}]{lmertest}
Alexandra Kuznetsova, Per~B. Brockhoff, and Rune H.~B. Christensen. 2017.
\newblock \href {https://doi.org/10.18637/jss.v082.i13} {lmer{T}est {P}ackage:
  {T}ests in {L}inear {M}ixed {E}ffects {M}odels}.
\newblock \emph{Journal of Statistical Software}, 82(13):1–26.

\bibitem[{Lenth(2023)}]{emmeans}
Russell~V. Lenth. 2023.
\newblock \href {https://CRAN.R-project.org/package=emmeans} {\emph{emmeans:
  {E}stimated {M}arginal {M}eans, aka {L}east-{S}quares {M}eans}}.
\newblock R package version 1.8.5.

\bibitem[{Lommel et~al.(2014{\natexlab{a}})Lommel, Popovic, and
  Burchardt}]{Lommel2014AssessingIA}
Arle Lommel, Maja Popovic, and Aljoscha Burchardt. 2014{\natexlab{a}}.
\newblock Assessing inter-annotator agreement for translation error annotation.
\newblock Reykjavik, Iceland. Proceedings of the 9th {I}nternational
  {C}onference on {L}anguage {R}esources and {E}valuation ({LREC} 14).

\bibitem[{Lommel et~al.(2014{\natexlab{b}})Lommel, Uszkoreit, and
  Burchardt}]{lommel2014multidimensional}
Arle Lommel, Hans Uszkoreit, and Aljoscha Burchardt. 2014{\natexlab{b}}.
\newblock \href {https://ddd.uab.cat/record/130144} {{{M}ultidimensional
  {Q}uality {M}etrics ({MQM}) : {A} {F}ramework for {D}eclaring and
  {D}escribing {T}ranslation {Q}uality {M}etrics}}.
\newblock \emph{Tradum{\`a}tica}, pages 0455--463.

\bibitem[{Lopes et~al.(2020)Lopes, Farajian, Bawden, Zhang, and
  Martins}]{lopes-etal-2020-document}
Ant{\'o}nio Lopes, M.~Amin Farajian, Rachel Bawden, Michael Zhang, and
  Andr{\'e} F.~T. Martins. 2020.
\newblock \href {https://aclanthology.org/2020.eamt-1.24} {Document-level
  neural {MT}: {A} {S}ystematic {C}omparison}.
\newblock In \emph{Proceedings of the 22nd Annual Conference of the European
  Association for Machine Translation}, pages 225--234, Lisboa, Portugal.
  European Association for Machine Translation.

\bibitem[{Mansimov et~al.(2021)Mansimov, Melis, and
  Yu}]{mansimov-etal-2021-capturing}
Elman Mansimov, G{\'a}bor Melis, and Lei Yu. 2021.
\newblock \href {https://doi.org/10.18653/v1/2021.codi-main.14} {Capturing
  document context inside sentence-level neural machine translation models with
  self-training}.
\newblock In \emph{Proceedings of the 2nd Workshop on Computational Approaches
  to Discourse}, pages 143--153, Punta Cana, Dominican Republic and Online.
  Association for Computational Linguistics.

\bibitem[{Miculicich et~al.(2018)Miculicich, Ram, Pappas, and
  Henderson}]{miculicich-etal-2018-document}
Lesly Miculicich, Dhananjay Ram, Nikolaos Pappas, and James Henderson. 2018.
\newblock \href {https://doi.org/10.18653/v1/D18-1325} {Document-level neural
  machine translation with hierarchical attention networks}.
\newblock In \emph{Proceedings of the 2018 Conference on Empirical Methods in
  Natural Language Processing}, pages 2947--2954, Brussels, Belgium.
  Association for Computational Linguistics.

\bibitem[{Molina and Hurtado~Albir(2004)}]{Molina2004-ms}
Luc{\'\i}a Molina and Amparo Hurtado~Albir. 2004.
\newblock {T}ranslation techniques revisited: {A} dynamic and functionalist
  approach.
\newblock \emph{Meta}, 47(4):498--512.

\bibitem[{Ning and Dom{\'i}nguez(2016)}]{Ning2016ComparativeLA}
Wang Ning and C{\'e}sar Dom{\'i}nguez. 2016.
\newblock Comparative literature and translation: A cross-cultural and
  interdisciplinary perspective.
\newblock In Yves Gambier and Luc van Doorslaer, editors, \emph{Border
  crossings. {T}ranslation studies and other disciplines}, pages 287--308. John
  Benjamins.

\bibitem[{O{\textquotesingle}Brien(2012)}]{OBrien2012}
Sharon O{\textquotesingle}Brien. 2012.
\newblock \href {https://doi.org/10.1075/ts.1.05obr} {Translation as
  human{\textendash}computer interaction}.
\newblock \emph{Translation Spaces}, 1:101--122.

\bibitem[{OpenAI(2022)}]{chatgpt_intro}
OpenAI. 2022.
\newblock \href {https://openai.com/blog/chatgpt} {Introducing {C}hat{GPT}}.

\bibitem[{OpenAI(2023)}]{openai2023gpt4}
OpenAI. 2023.
\newblock \href {http://arxiv.org/abs/2303.08774} {{GPT}-4 technical report}.

\bibitem[{Ouyang et~al.(2022)Ouyang, Wu, Jiang, Almeida, Wainwright, Mishkin,
  Zhang, Agarwal, Slama, Ray et~al.}]{ouyang2022training}
Long Ouyang, Jeff Wu, Xu~Jiang, Diogo Almeida, Carroll~L Wainwright, Pamela
  Mishkin, Chong Zhang, Sandhini Agarwal, Katarina Slama, Alex Ray, et~al.
  2022.
\newblock Training language models to follow instructions with human feedback.
\newblock \emph{arXiv preprint arXiv:2203.02155}.

\bibitem[{Pawlak(2023)}]{ATA_trans_2023}
Dorota Pawlak. 2023.
\newblock \href
  {https://www.atanet.org/starting-your-career/chatgpt-for-translators-how-to-use-the-tool-to-work-more-efficiently/}
  {Chat{GPT} for {T}ranslators: {H}ow to {U}se the {T}ool to {W}ork {M}ore
  {E}fficiently?}

\bibitem[{Qi et~al.(2020)Qi, Zhang, Zhang, Bolton, and Manning}]{qi2020stanza}
Peng Qi, Yuhao Zhang, Yuhui Zhang, Jason Bolton, and Christopher~D. Manning.
  2020.
\newblock \href {https://nlp.stanford.edu/pubs/qi2020stanza.pdf} {Stanza: {A}
  {Python} {N}atural {L}anguage {P}rocessing {T}oolkit for {M}any {H}uman
  {L}anguages}.
\newblock In \emph{Proceedings of the 58th Annual Meeting of the Association
  for Computational Linguistics: System Demonstrations}.

\bibitem[{Rei et~al.(2022)Rei, C.~de Souza, Alves, Zerva, Farinha, Glushkova,
  Lavie, Coheur, and Martins}]{rei-etal-2022-comet}
Ricardo Rei, Jos{\'e}~G. C.~de Souza, Duarte Alves, Chrysoula Zerva, Ana~C
  Farinha, Taisiya Glushkova, Alon Lavie, Luisa Coheur, and Andr{\'e} F.~T.
  Martins. 2022.
\newblock \href {https://aclanthology.org/2022.wmt-1.52} {{COMET}-22:
  {U}nbabel-{IST} 2022 {S}ubmission for the {M}etrics {S}hared {T}ask}.
\newblock In \emph{Proceedings of the Seventh Conference on Machine Translation
  (WMT)}, pages 578--585, Abu Dhabi, United Arab Emirates (Hybrid). Association
  for Computational Linguistics.

\bibitem[{Rei et~al.(2021)Rei, Farinha, Zerva, van Stigt, Stewart, Ramos,
  Glushkova, Martins, and Lavie}]{rei-etal-2021-references}
Ricardo Rei, Ana~C Farinha, Chrysoula Zerva, Daan van Stigt, Craig Stewart,
  Pedro Ramos, Taisiya Glushkova, Andr{\'e} F.~T. Martins, and Alon Lavie.
  2021.
\newblock \href {https://aclanthology.org/2021.wmt-1.111} {Are {R}eferences
  {R}eally {N}eeded? {U}nbabel-{IST} 2021 {S}ubmission for the {M}etrics
  {S}hared {T}ask}.
\newblock In \emph{Proceedings of the Sixth Conference on Machine Translation},
  pages 1030--1040, Online. Association for Computational Linguistics.

\bibitem[{Sager(1998)}]{Sager1998}
Juan~C. Sager. 1998.
\newblock \href {https://doi.org/10.1080/13556509.1998.10799007} {{W}hat
  {D}istinguishes {M}ajor {T}ypes of {T}ranslation?}
\newblock \emph{The Translator}, 4(1):69--89.

\bibitem[{Sellam et~al.(2020)Sellam, Das, and Parikh}]{sellam-etal-2020-bleurt}
Thibault Sellam, Dipanjan Das, and Ankur Parikh. 2020.
\newblock \href {https://doi.org/10.18653/v1/2020.acl-main.704} {{BLEURT}:
  {L}earning {R}obust {M}etrics for {T}ext {G}eneration}.
\newblock In \emph{Proceedings of the 58th Annual Meeting of the Association
  for Computational Linguistics}, pages 7881--7892, Online. Association for
  Computational Linguistics.

\bibitem[{Taivalkoski-Shilov(2019{\natexlab{a}})}]{ethics_trans_2019}
Kristiina Taivalkoski-Shilov. 2019{\natexlab{a}}.
\newblock \href {https://doi.org/10.1080/0907676X.2018.1520907} {Ethical issues
  regarding machine(-assisted) translation of literary texts}.
\newblock \emph{Perspectives}, 27(5):689--703.

\bibitem[{Taivalkoski-Shilov(2019{\natexlab{b}})}]{taivalkoski2019free}
Kristiina Taivalkoski-Shilov. 2019{\natexlab{b}}.
\newblock Free indirect discourse: an insurmountable challenge for literary
  {MT} systems?
\newblock In \emph{Proceedings of the Qualities of Literary Machine
  Translation}, pages 35--39.

\bibitem[{Tan et~al.(2019)Tan, Zhang, Xiong, and
  Zhou}]{tan-etal-2019-hierarchical}
Xin Tan, Longyin Zhang, Deyi Xiong, and Guodong Zhou. 2019.
\newblock \href {https://doi.org/10.18653/v1/D19-1168} {Hierarchical modeling
  of global context for document-level neural machine translation}.
\newblock In \emph{Proceedings of the 2019 Conference on Empirical Methods in
  Natural Language Processing and the 9th International Joint Conference on
  Natural Language Processing (EMNLP-IJCNLP)}, pages 1576--1585, Hong Kong,
  China. Association for Computational Linguistics.

\bibitem[{Thai et~al.(2022)Thai, Karpinska, Krishna, Ray, Inghilleri, Wieting,
  and Iyyer}]{thai-etal-2022-exploring}
Katherine Thai, Marzena Karpinska, Kalpesh Krishna, Bill Ray, Moira Inghilleri,
  John Wieting, and Mohit Iyyer. 2022.
\newblock \href {https://aclanthology.org/2022.emnlp-main.672} {Exploring
  document-level literary machine translation with parallel paragraphs from
  world literature}.
\newblock In \emph{Proceedings of the 2022 Conference on Empirical Methods in
  Natural Language Processing}, pages 9882--9902, Abu Dhabi, United Arab
  Emirates. Association for Computational Linguistics.

\bibitem[{Thomson et~al.(2023)Thomson, Reiter, and
  Sundararajan}]{THOMSON2023101482}
Craig Thomson, Ehud Reiter, and Barkavi Sundararajan. 2023.
\newblock \href {https://doi.org/https://doi.org/10.1016/j.csl.2023.101482}
  {Evaluating factual accuracy in complex data-to-text}.
\newblock \emph{Computer {S}peech \& {L}anguage}, 80:101482.

\bibitem[{Tiedemann and Scherrer(2017)}]{tiedemann-scherrer-2017-neural}
J{\"o}rg Tiedemann and Yves Scherrer. 2017.
\newblock \href {https://doi.org/10.18653/v1/W17-4811} {Neural machine
  translation with extended context}.
\newblock In \emph{Proceedings of the Third Workshop on Discourse in Machine
  Translation}, pages 82--92, Copenhagen, Denmark. Association for
  Computational Linguistics.

\bibitem[{Tkachenko et~al.(2020-2022)Tkachenko, Malyuk, Holmanyuk, and
  Liubimov}]{LabelStudio}
Maxim Tkachenko, Mikhail Malyuk, Andrey Holmanyuk, and Nikolai Liubimov.
  2020-2022.
\newblock \href {https://github.com/heartexlabs/label-studio} {{Label Studio}:
  {D}ata labeling software}.
\newblock Open source software available from
  https://github.com/heartexlabs/label-studio.

\bibitem[{Toral and Way(2015)}]{Toral2015}
Antonio Toral and Andy Way. 2015.
\newblock \href {https://doi.org/10.1075/ts.4.2.04tor} {Machine-assisted
  translation of literary text}.
\newblock \emph{Translation Spaces}, 4(2):240--267.

\bibitem[{Utiyama and Isahara(2007)}]{utiyama-isahara-2007-comparison}
Masao Utiyama and Hitoshi Isahara. 2007.
\newblock \href {https://aclanthology.org/N07-1061} {A comparison of pivot
  methods for phrase-based statistical machine translation}.
\newblock In \emph{Human Language Technologies 2007: The Conference of the
  North {A}merican Chapter of the Association for Computational Linguistics;
  Proceedings of the Main Conference}, pages 484--491, Rochester, New York.
  Association for Computational Linguistics.

\bibitem[{Vernikos et~al.(2022)Vernikos, Thompson, Mathur, and
  Federico}]{easy_doc_mt}
Giorgos Vernikos, Brian Thompson, Prashant Mathur, and Marcello Federico. 2022.
\newblock \href {https://statmt.org/wmt22/pdf/2022.wmt-1.6.pdf}
  {{E}mbarrassingly {E}asy {D}ocument-level {MT} {M}etrics: {H}ow to {C}onvert
  {A}ny {P}retrained {M}etric {I}nto a {D}ocument-{L}evel {M}etric}.
\newblock In \emph{Proceedings of the Seventh Conference on Machine
  Translation}, Abu Dhabi, United Arab Emirates. Association for Computational
  Linguistics.

\bibitem[{Vilar et~al.(2022)Vilar, Freitag, Cherry, Luo, Ratnakar, and
  Foster}]{vilar2022prompting}
David Vilar, Markus Freitag, Colin Cherry, Jiaming Luo, Viresh Ratnakar, and
  George Foster. 2022.
\newblock \href {http://arxiv.org/abs/2211.09102} {Prompting palm for
  translation: Assessing strategies and performance}.

\bibitem[{Voita et~al.(2019)Voita, Sennrich, and Titov}]{voita-etal-2019-good}
Elena Voita, Rico Sennrich, and Ivan Titov. 2019.
\newblock \href {https://doi.org/10.18653/v1/P19-1116} {When a good translation
  is wrong in context: Context-aware machine translation improves on deixis,
  ellipsis, and lexical cohesion}.
\newblock In \emph{Proceedings of the 57th Annual Meeting of the Association
  for Computational Linguistics}, pages 1198--1212, Florence, Italy.
  Association for Computational Linguistics.

\bibitem[{Zhang et~al.(2022)Zhang, Bapna, Johnson, Dabirmoghaddam, Arivazhagan,
  and Firat}]{zhang-etal-2022-multilingual}
Biao Zhang, Ankur Bapna, Melvin Johnson, Ali Dabirmoghaddam, Naveen
  Arivazhagan, and Orhan Firat. 2022.
\newblock \href {https://doi.org/10.18653/v1/2022.acl-long.287} {Multilingual
  document-level translation enables zero-shot transfer from sentences to
  documents}.
\newblock In \emph{Proceedings of the 60th Annual Meeting of the Association
  for Computational Linguistics (Volume 1: Long Papers)}, pages 4176--4192,
  Dublin, Ireland. Association for Computational Linguistics.

\bibitem[{Zhang et~al.(2023)Zhang, Haddow, and Birch}]{zhang2023prompting}
Biao Zhang, Barry Haddow, and Alexandra Birch. 2023.
\newblock \href {http://arxiv.org/abs/2301.07069} {Prompting large language
  model for machine translation: A case study}.

\bibitem[{Zhang et~al.(2018)Zhang, Luan, Sun, Zhai, Xu, Zhang, and
  Liu}]{zhang-etal-2018-improving}
Jiacheng Zhang, Huanbo Luan, Maosong Sun, Feifei Zhai, Jingfang Xu, Min Zhang,
  and Yang Liu. 2018.
\newblock \href {https://doi.org/10.18653/v1/D18-1049} {Improving the
  transformer translation model with document-level context}.
\newblock In \emph{Proceedings of the 2018 Conference on Empirical Methods in
  Natural Language Processing}, pages 533--542, Brussels, Belgium. Association
  for Computational Linguistics.

\bibitem[{Zhang et~al.(2020)Zhang, Kishore, Wu, Weinberger, and
  Artzi}]{zhang2020bertscore}
Tianyi Zhang, Varsha Kishore, Felix Wu, Kilian~Q. Weinberger, and Yoav Artzi.
  2020.
\newblock \href {http://arxiv.org/abs/1904.09675} {{BERTS}core: {E}valuating
  {T}ext {G}eneration with {BERT}}.

\bibitem[{Zheng et~al.(2020)Zheng, Yue, Huang, Chen, and Birch}]{ijcai2020p551}
Zaixiang Zheng, Xiang Yue, Shujian Huang, Jiajun Chen, and Alexandra Birch.
  2020.
\newblock \href {https://doi.org/10.24963/ijcai.2020/551} {Towards making the
  most of context in neural machine translation}.
\newblock In \emph{Proceedings of the Twenty-Ninth International Joint
  Conference on Artificial Intelligence, {IJCAI-20}}, pages 3983--3989.
  International Joint Conferences on Artificial Intelligence Organization.
\newblock Main track.

\end{thebibliography}
\bibliographystyle{acl_natbib}
\newpage

\appendix
\section*{Appendix}
\label{sec:appendix}

\section{The Dataset}
\label{app:dataset}

\paragraph{Choosing paragraphs:} The selection of a particular paragraph was semi-random, with certain considerations in mind during the sampling process. We prioritized the following criteria: (1) for each source language we sample paragraphs so that there is a combination of dialogue and narrative texts; (2) the paragraph should be reasonably intelligible to a human translator without additional context; and (3) alignment between the source paragraph and human translation should be feasible, meaning no major content rearrangement \textit{across} paragraphs.

Nonetheless, meeting all these requirements was not always possible. For instance, the source text of \textit{Convenience Store Woman} (\textit{ja}) is mostly written in the first-person narrative. Since Japanese does not encode the speaker`s gender in the verb forms, it is often impossible to determine whether the narrator is a male or a female. In cases where it was impossible to determine the gender of the character we instructed translators to accept \textit{either} option, provided that the translation remained consistent within the given paragraph (i.e., the gender did not change within the paragraph).

\paragraph{A note on literary translation:} It is important to understand the challenges a human translator faces when translating literary texts. Even a simple sentence may lead to substantial struggles. One of the most notable examples is the first sentence of a French novel \textit{The Stranger} by Albert Camus. The story begins in a seemingly trivial way:

\begin{exe}
\small

    \ex \label{ex:stranger} Aujourd’hui, maman est morte. \\
    \vspace{1.5mm}
        \textit{Today, mother died.} 
    \begin{flushright}
    \scriptsize
    \vspace{-0.2cm}
    ―\textsc{French Source} (from \textit{The Stranger})
    \end{flushright}
\end{exe}

While there is nothing particularly difficult in (\ref{ex:stranger}), five English translations of ``The Stranger'' has already been produced and there is little consensus on what the ideal translation should be \cite{Kaplansky2004stranger}.

This very first sentence is of the utmost importance as it introduces the reader to Meursault, the protagonist of the story, who will later kill an unnamed Arab without any apparent reason. Hence, it is crucial for the storyline that the reader understands, from this very beginning, who Meursault is and what affection he holds for his mother. 

 Stuart Gilbert (1946), Joseph Laredo (1982), and Kate Griffith (1982) all translate the beginning sentence as \textit{Mother died today} but this translation is problematic. English word ``mother,'' while technically correct, is too formal to fully embrace the emotions conveyed by the French ``maman.'' Mathew Ward (1988) opts to leave the French ``maman'' untranslated. An understandable choice, as the English reader is likely to decipher the meaning from the surface similarity, though they may not fully grasp its sentiment. 
 Conversely, Sandra Smith (2012) attempts to capture the intimacy of ``maman'' by rendering it as ``\textit{my} mother,'' which is less formal than a simple ``mother'' but doesn't possess the childlike connotation of the English ``mom.''

Literary translation is clearly a challenge that exceeds simple word equivalence. Professional translators face choices, that current systems are unlikely to solve independently. 
However, they can assist translators in their tasks, in a way similar to how computer-assisted translation (CAT) tools have been doing. This approach holds the potential to make more novels accessible to a wider audience; novels that may have remained untranslated otherwise.


\begin{table*}[h!]
\centering
\begin{tabular}{@{}llcc@{}}
\toprule
\textsc{Language} & \textsc{Language Family} & \textsc{Morphological Features} & \textsc{Writing System} \\ \midrule
\rowcolor{violet!10}
\textsc{English}      & Indo-European (Germanic)                & Analytic                      & Latin Alphabet           \\
\textsc{German}       & Indo-European (Germanic)                & Fusional                      & Latin Alphabet           \\
\rowcolor{violet!10}
\textsc{French}            & Indo-European (Romance)                 & Fusional                      & Latin Alphabet           \\
\textsc{Polish}            & Indo-European (Slavic)                  & Fusional                      & Latin Alphabet           \\
\rowcolor{violet!10}
\textsc{Czech }            & Indo-European (Slavic)                  & Fusional                      & Latin Alphabet           \\
\textsc{Russian}           & Indo-European (Slavic)                  & Fusional                      & Cyrillic                 \\
\rowcolor{violet!10}
\textsc{Japanese}          & Japonic                 & Agglutinative                 & Kanji / Hiragana / Katakana \\
\textsc{Chinese}           & Sino-Tibetan            & Analytic                      & Hanzi                    \\ \bottomrule
\end{tabular}
\caption{Details on languages included for the current study.}
\label{table:language_comparison}
\end{table*}

\begin{table*}[!htbp]
\centering
\resizebox{0.95\width}{!}{
\begin{tabular}{@{} l *{8}{>{\centering\arraybackslash}p{1.5cm}} @{}}
\toprule
\textsc{Lang} & \textsc{\#Sent} & \textsc{Src} & \textsc{Hum} & \textsc{Para} & \textsc{Sent} & \textsc{Para\_Sent} & \textsc{Gtr} \\ 
\midrule
cs-pl & 163 & 2,154 & 2,027 & 2,122 & 2,123 & 2,259 & 2,065 \\
de-pl & 153 & 3,172 & 2,997 & 2,785 & 2,899 & 2,835 & 2,764 \\
ru-pl & 170 & 2,350 & 2,471 & 2,467 & 2,463 & 2,458 & 2,375 \\
ja-pl & 111 &  2,627 & 1,855 & 1,782 & 1,907 & 1,830 & 1,800 \\
en-pl & 127 &  1,702 & 1,526 & 1,444 & 1,513 & 1,483 & 1,462 \\
fr-pl & 119 &  3,253 & 2,789 & 2,641 & 2,673 & 2,654 & 2,543 \\
de-ja & 75  &  3,530 & 5,329 & 4,807 & 5,116 & 4,652 & 4,703 \\
en-ja & 176 &  1,959 & 2,617 & 2,538 & 2,653 & 2,617 & 2,634 \\
zh-ja & 194 &  2,998 & 4,124 & 3,861 & 4,249 & 3,957 & 3,978 \\
ru-ja & 193 &  2,539 & 4,753 & 3,982 & 4,348 & 4,088 & 3,921 \\
fr-ja & 195 &  2,510 & 3,426 & 3,110 & 3,355 & 3,106 & 2,958 \\
pl-ja & 188 &  1,953 & 2,944 & 3,083 & 3,418 & 3,199 & 2,972 \\
ja-en & 111 &  2,622 & 2,293 & 2,062 & 2,322 & 2,257 & 2,140 \\
pl-en & 148 &  2,696 & 3,430 & 3,234 & 3,290 & 3,273 & 3,213 \\
ru-en & 117 &  1,693 & 2,008 & 2,029 & 2,056 & 2,028 & 2,019 \\
fr-en & 120 &  3,253 & 3,123 & 3,067 & 3,150 & 3,064 & 3,098 \\
de-en & 153 &  3,172 & 3,346 & 3,361 & 3,413 & 3,325 & 3,314 \\
zh-en & 127 &  2,235 & 2,002 & 2,427 & 2,396 & 2,351 & 2,360 \\
\midrule
\textbf{Total} & \textbf{2,640} & \textbf{46,418} & \textbf{53,060} & \textbf{50,802} & \textbf{53,344} & \textbf{51,436} & \textbf{50,319} \\
\end{tabular}
}
\caption{Number of sentences in the source text sentencized manually (\textsc{\#Sent}) along with the number of tokens in the human reference (\textsc{Hum}) and different machine translations (\textsc{Para}, \textsc{Sent}, \textsc{Para\_Sent}, \textsc{Gtr}). All translations were tokenized using \textsc{SpaCy}\footnote{\url{https://spacy.io/}} with the large model for each of the three target languages (Polish, Japanese, and English). All source texts were tokenized with \textsc{Stanza} \cite{qi2020stanza} as \textsc{Spacy} does not include models for all target languages.}
\label{table:tokens_in_paragraphs}
\end{table*}

\begin{table*}[htbp]
\centering
\resizebox{0.95\width}{!}{
\begin{tabular}{@{}ccccccc@{}}
\toprule
\textsc{Lang} & \textsc{Source} & \textsc{Target} & \textsc{Para} & \textsc{Sent} & \textsc{Para\_Sent} & \textsc{GTr} \\ 
\midrule
\textit{cs-pl} & 168 & 177 & 167 & 169 & 181 & 168 \\
\textit{de-en} & 155 & 182 & 166 & 167 & 164 & 155 \\
\textit{de-ja} & 69 & 133 & 135 & 121 & 117 & 132 \\
\textit{de-pl} & 155 & 170 & 166 & 167 & 169 & 157 \\
\textit{en-ja} & 169 & 168 & 166 & 161 & 169 & 169 \\
\textit{en-pl} & 131 & 127 & 130 & 132 & 130 & 131 \\
\textit{fr-en} & 122 & 138 & 126 & 122 & 124 & 123 \\
\textit{fr-ja} & 193 & 199 & 207 & 220 & 185 & 201 \\
\textit{fr-pl} & 122 & 125 & 125 & 125 & 126 & 123 \\
\textit{ja-en} & 101 & 120 & 116 & 116 & 116 & 111 \\
\textit{ja-pl} & 101 & 127 & 117 & 115 & 118 & 108 \\
\textit{pl-en} & 148 & 156 & 149 & 145 & 151 & 145 \\
\textit{pl-ja} & 189 & 153 & 174 & 196 & 178 & 191 \\
\textit{ru-en} & 123 & 119 & 121 & 124 & 121 & 123 \\
\textit{ru-ja} & 144 & 155 & 158 & 161 & 164 & 196 \\
\textit{ru-pl} & 168 & 172 & 170 & 171 & 172 & 172 \\
\textit{zh-en} & 127 & 130 & 146 & 141 & 140 & 135 \\
\textit{zh-ja} & 195 & 234 & 225 & 229 & 215 & 202 \\
\bottomrule
\textsc{Total} & \textbf{2,580} & \textbf{2,785} & \textbf{2,764} & \textbf{2,782} & \textbf{2,740} & \textbf{2,742} \\ 
\end{tabular}
}
\caption{Number of sentences in the source text and each translation. The data was sentencized with \textsc{SpaCy}. As evident from the data and manual inspection of translations the translations may result in a very different number of sentences as a result of splits and merges. We observe that about 55\% of the data potentially lacks of one-to-one correspondence.}
\label{table:sent_auto_in_paragraphs}
\end{table*}

\begin{figure}[h!]
\centering
\includegraphics[width=.5 \textwidth]{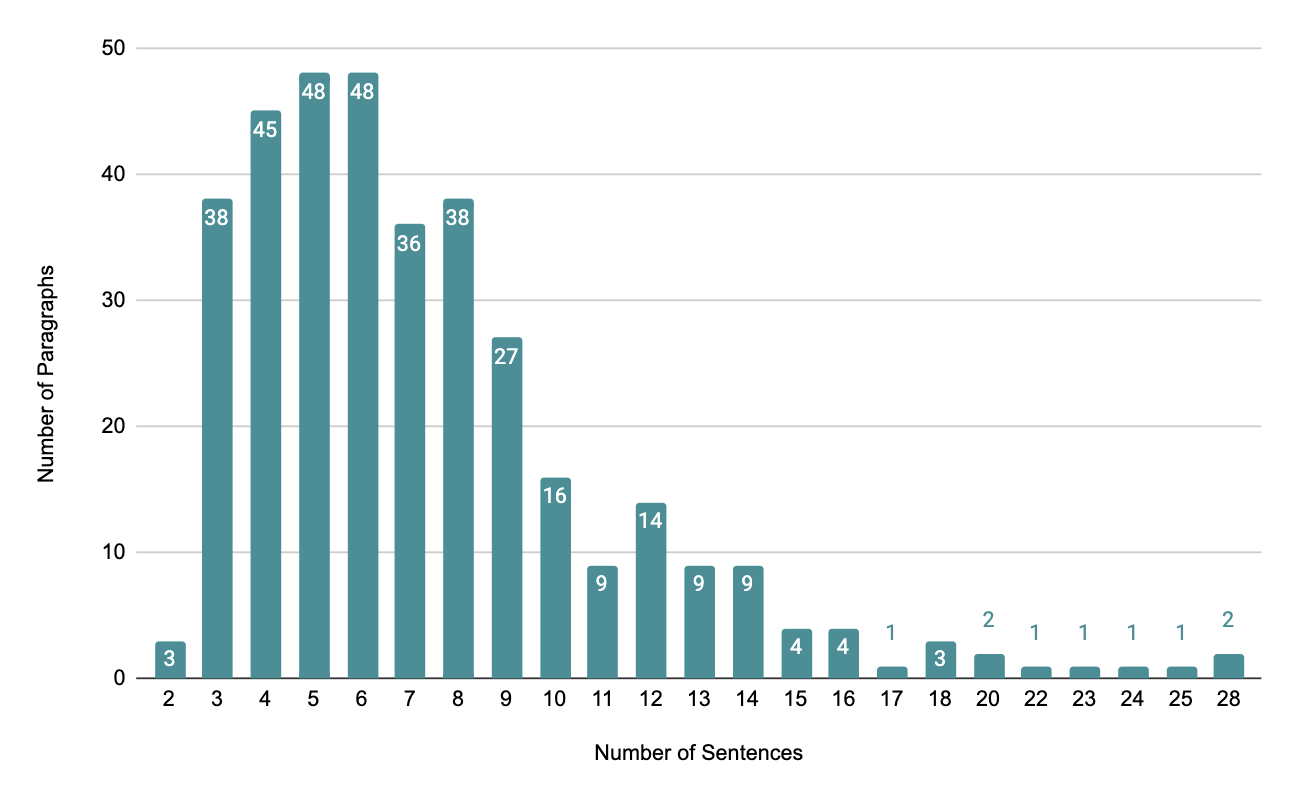}
\caption{Distribution of sentences in the sampled paragraphs. The paragraphs were sentencized manually.}
\label{figure:hum_vs_GTr}
\end{figure}

\begin{table*}
\centering
\resizebox{\textwidth}{!}{
\begin{tabular}{lllllc}
\toprule
                    &                      &                      &   & \multicolumn{2}{c}{\textsc{Year Published}}  \\
 \multirow{-2}{*}{\textsc{Lang Pair}}  & \multirow{-2}{*}{\textsc{Title}} & \multirow{-2}{*}{\textsc{Author}} & \multirow{-2}{*}{\textsc{Translator(s)}} & \textsc{Translation} & \textsc{Original} \\

\midrule
\textit{ja-pl} & Norwegian Wood & Haruki Murakami & Dorota Marczewska \&  & 1987 & 2006 \\
      &                &                 & Anna Zielińska-Elliott &     &      \\
\textit{de-pl} & The Trial & Franz Kafka & Jakub Ekier & 1925 & 2008 \\
\textit{fr-pl} & Les Miserables & Victor Hugo & Krystyna Byczewska & 1862 & 1966 \\
\textit{fr-pl} & The Little Prince & Antoine de Saint-Exupéry & Jan Szwykowski & 1862 & 1967 \\
\textit{en-pl} & The Valley of Fear & Arthur Conan Doyle & Tadeusz Evert & 1915 & 1927 \\
\textit{ru-pl} & War and Peace & Leo Tolstoy & Andrzej Stawar & 1869 & 1958 \\
\textit{cs-pl} & War with Newts & Karel Čapek & Jadwiga Bułakowska & 1936 & 1949 \\
\textit{pl-ja} & Solaris & Stanisław Lem & Mitsuyoshi Numano & 1961 & 2004 \\
\textit{ru-ja} & Anna Karenina & Leo Tolstoy & Hakuyō Nakamura & 1878 & 2004 \\
\textit{de-ja} & Der Steppenwolf & Hermann Hesse & Fujio Nagano & 1927 & 2000 \\
\textit{fr-ja} & Around the World in 80 Days & Jules Verne & Yū Takano & 1873 & 2009 \\
\textit{en-ja} & Animal Farm & George Orwell & Eitarō Sayama & 1945 & 1998 \\
\textit{zh-ja} & Medicine & Lu Xun & Kōbai Inoue & 1919 & 1919 \\
\textit{zh-ja} & The True Story of Ah Q & Lu Xun & Kōbai Inoue & 1921 & 1923 \\
\textit{zh-ja} & Diary of a Madman & Lu Xun & Kōbai Inoue & 1921 & 1923 \\
\textit{ru-en} & Confession & Leo Tolstoy & Peter Carson & 1882 & 2013 \\
\textit{zh-en} & The Day the Sun Died & Yan Lianke & Carlos Rojas & 2015 & 2018 \\
\textit{ja-en} & Kokoro & Natsume Sōseki & Edwin McClelan  & 1914 & 1957 \\
\textit{ja-en} & Kokoro & Natsume Sōseki & Meredith McKinney  & 1914 & 2010 \\
\textit{de-en} & Venus in Furs & Ritter von Leopold Sacher-Masoch & Fernanda Savage & 1870 & \textit{unclear} \\
\textit{fr-en} & The Debacle & Émile Zola & Leonard Tancock & 1870 & 1972 \\
\bottomrule
\end{tabular}
}
\caption{List of novels employed in the prompts.}
\label{tab:prompt_novel_data}
\end{table*}

\section{Prompt Examples}
\label{app:prompt_examples}
\begin{figure*}[!t]
\centering
\resizebox{0.7\textwidth}{!}{
\includegraphics[width=\textwidth]{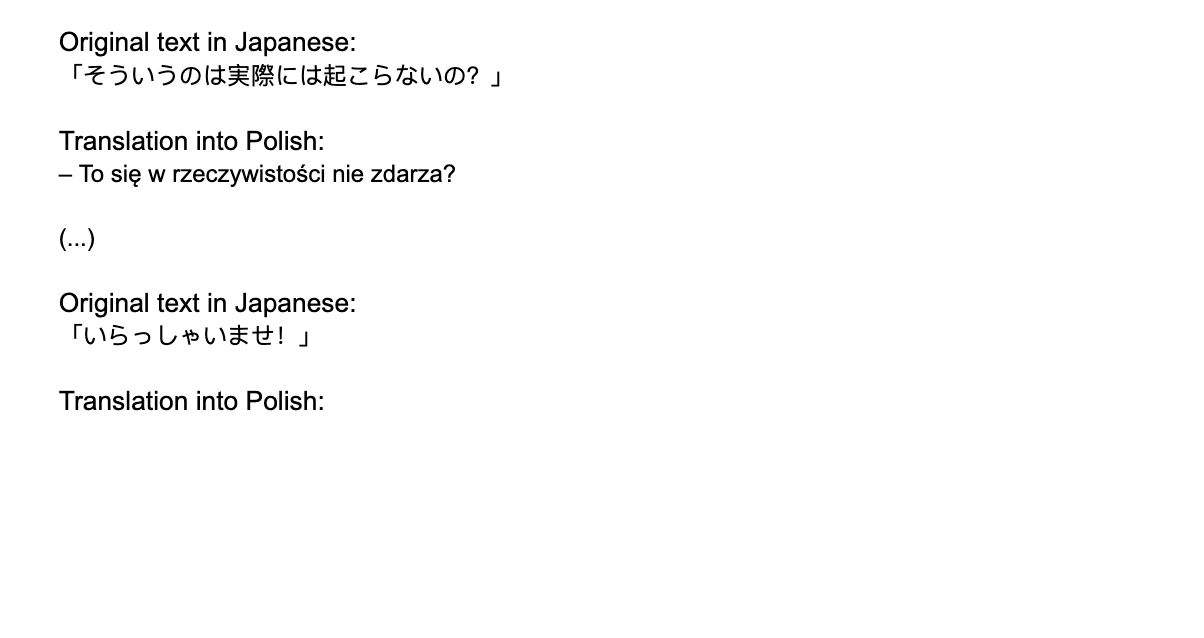}
}
\caption{An example of prompt for \textsc{Sent} translations with one demonstration and a text to translate.}
\label{figure:prompt_sent}
\end{figure*}

\begin{figure*}[!t]
\centering
\resizebox{0.6\textwidth}{!}{
\includegraphics[width=\textwidth]{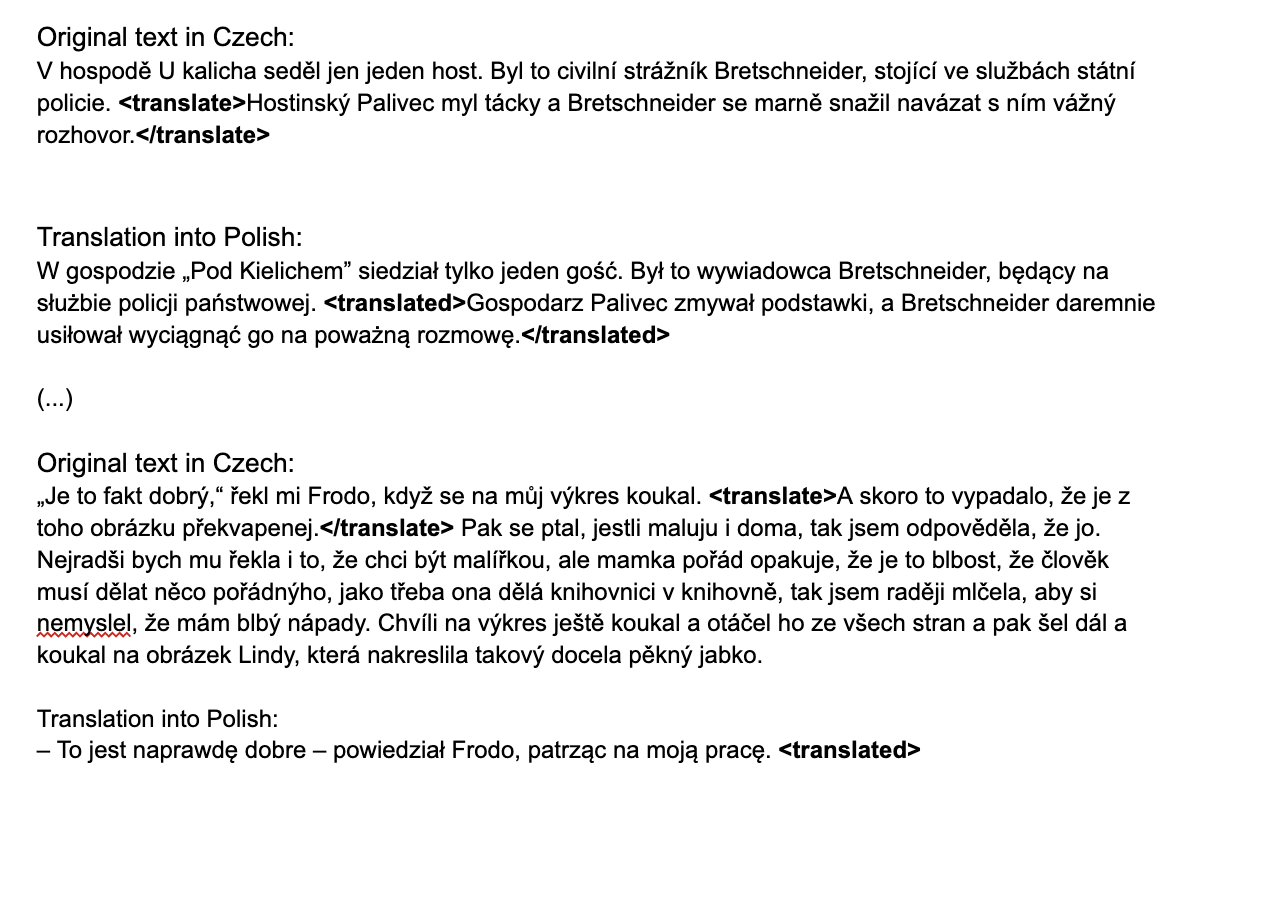}
}
\caption{An example of prompt for \textsc{Para\_Sent} translations with one demonstration and a text to translate.}
\label{figure:prompt_parasent}
\end{figure*}

\begin{figure*}[!t]
\centering
\resizebox{0.6\textwidth}{!}{
\includegraphics[width=\textwidth]{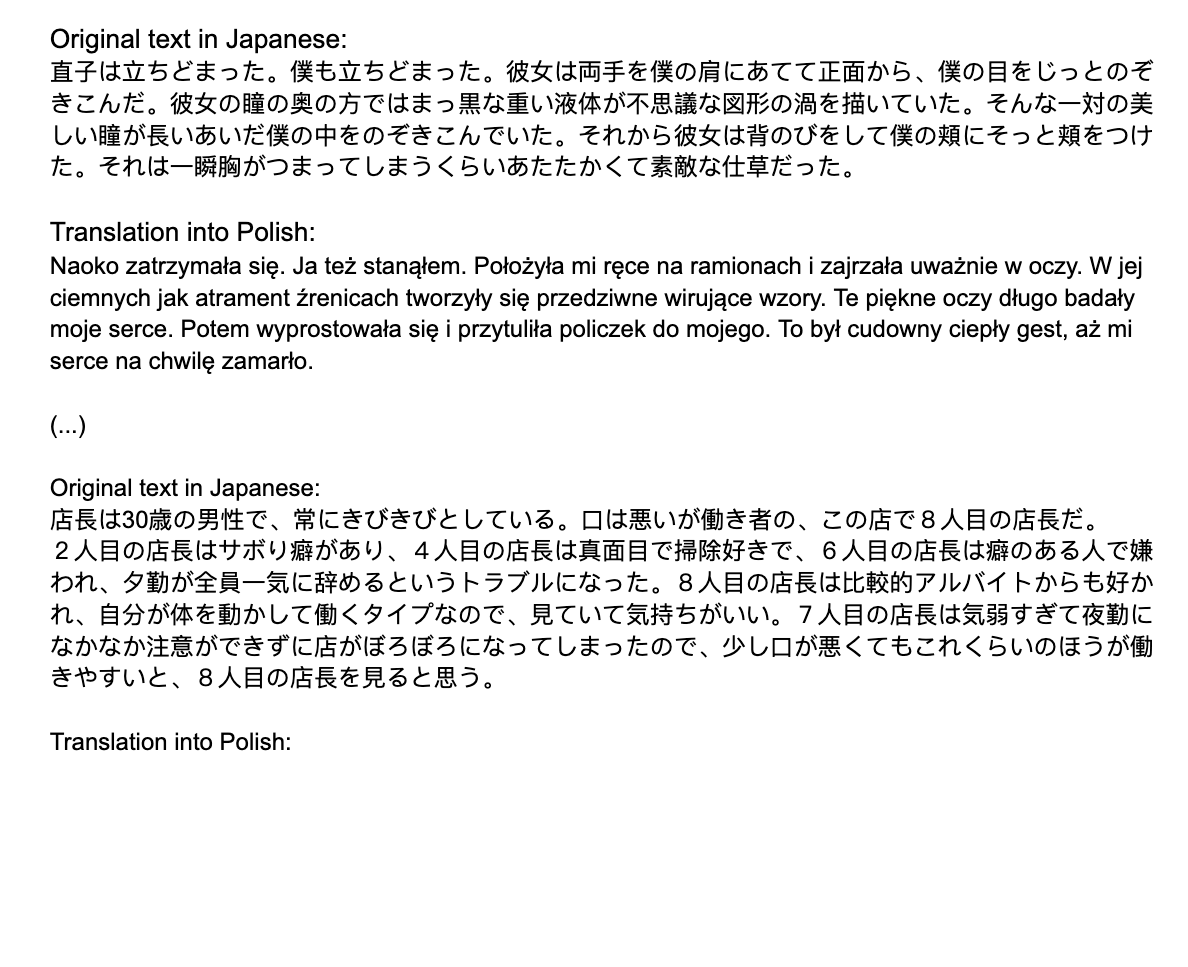}
}
\caption{An example of prompt for \textsc{Para} translations with one demonstration and a text to translate.}
\label{figure:prompt_para}
\end{figure*}

Here we present examples of prompts employed for the translation with \textsc{GPT-3.5}. The prompt wording for \textsc{Sent}, \textsc{Para\_sent}, and \textsc{Para}, with one demonstration each are presented in \autoref{figure:prompt_sent}, \autoref{figure:prompt_parasent}, \autoref{figure:prompt_para} respectively.

\section{Human Evaluation}
\label{app:hum_eval}

In this section, we provide some further details about the human evaluation with a focus on the error annotation. First, discuss the issue of subjectivity in error annotation. Next, we explain some choices we had to make when annotating ``inconsistency'' and ``format'' errors. Then, we discuss the issue of omissions in the produced translations. Finally, we present some details about the translators hired for the evaluation task.

\begin{table*}[ht]
\centering
\begin{tabular}{@{}cccl@{}}
\toprule
\textsc{Lang Pair} & \textsc{Native Lang} & \textsc{Book Familiarity} & \textsc{Gender} \\ \midrule
\textit{zh-en}    & Chinese     & \ding{55}              & Male   \\
\textit{ja-en}    & English     & \ding{55}              & Male   \\
\textit{de-en}    & Polish/English  & \ding{55}          & Female \\
\textit{fr-en}    & English     & \ding{55}              & Female \\
\textit{ru-en}    & Russian     & \ding{55}             & Female \\
\textit{pl-en}    & Polish/English  & \ding{55}          & Female \\
\textit{en-ja}    & Japanese    & \ding{55}              & Female \\
\textit{fr-ja}    & Japanese    & \ding{55}              & Male   \\
\textit{de-ja}    & Japanese    & \ding{55}              & Female \\
\textit{pl-ja}    & Polish (author)      & \ding{55}              & Female \\
\textit{ru-ja}    & Japanese    & \ding{55}              & Male   \\
\textit{zh-ja}    & Japanese    & \ding{55}              & Male   \\
\textit{de-pl}    & Polish/English  & \ding{55}          & Female \\
\textit{en-pl}    & Polish (author)     & \ding{55}              & Female \\
\textit{ru-pl}    & Polish/Russian  & \ding{55}          & Female \\
\textit{cs-pl}    & Czech       & \ding{55}              & Male   \\
\textit{ja-pl}    & Polish (author)     & \ding{55}              & Female \\
\textit{fr-pl}    & Polish      & $\checkmark$             & Female \\ \bottomrule
\end{tabular}
\caption{Details about the translators hired for the current annotation study. We consider a translator bilingual only if they were raised using both languages (e.g., \textit{ru-pl} translator was raised in Poland while speaking Russian at home). In the broader sense of this word, all of the translators are bilingual with some of them being trilingual. For the cases where the hired translator was \textit{not} a native speaker of the target language, the annotations were verified by a native speaker of the target language in consultation with the translator. Only one translator reported familiarity with the source text she was asked to evaluate. All translators were asked to evaluate each passage in isolation allowing for all possible interpretations based on the given part of the source text. Three language pairs (\textit{pl-ja}, \textit{en-pl}, \textit{ja-pl} were annotated by the first author of this paper.}
\label{tab:annotators_details}
\end{table*}

\begin{figure*}[!t]
\centering
\includegraphics[width=\textwidth]{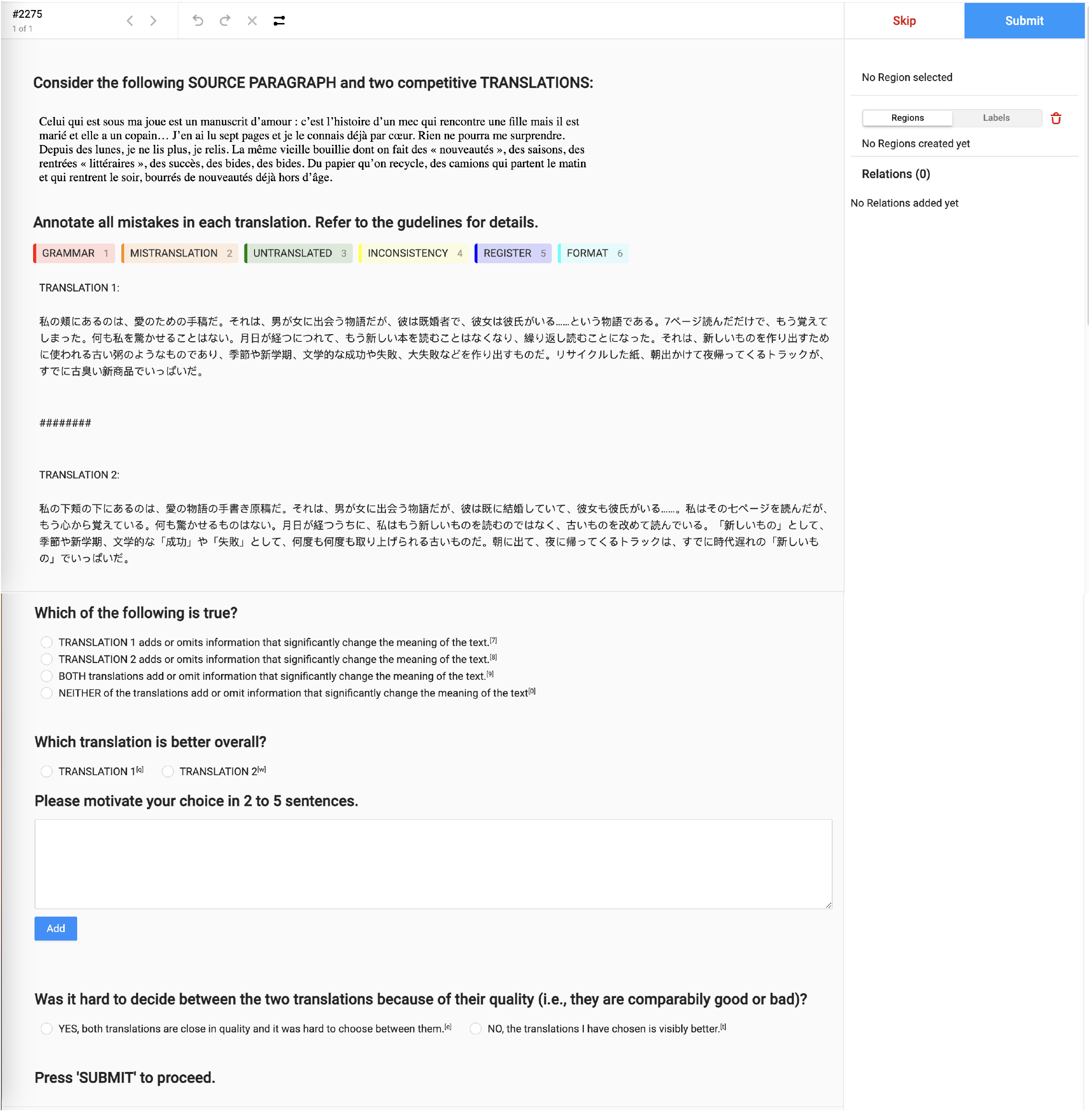}
\caption{The annotation interface used for the error annotation task.}
\label{figure:annot_interface}
\end{figure*}

\paragraph{Error annotation:} Annotating and classifying errors in translations is inherently subjective \cite{Lommel2014AssessingIA, tqa_han_2020}. For instance, translating French ``corsage'' (``bodice'') as a ``blouse'' can be seen as either a mistranslation or a permissible deviation from the original text; this is, in fact, how the ``corsage'' was translated by the human translator in our data. 

Furthermore, sometimes there are multiple ways of annotating errors \cite{THOMSON2023101482}. Consider the following example:

\begin{exe}
    \small
    \ex \label{mistran_exe} We had to hide the running, though, in case our haste betrayed us, so truer to say we slipped out quietly. When one of my parents appeared, my technique was: pretend to catch sight of someone in the next room. \textbf{Move in a natural manner toward this figment of my imagination, making a purposeful face.}
    \begin{flushright}
    \scriptsize
    \vspace{-0.2cm}
    ―\textsc{English Source} (from \textit{A Children's Bible})
    \end{flushright}
\end{exe}

The translation of the last sentence in (\ref{mistran_exe}) into Polish as an imperative can be considered a mistranslation. We would hypothesis that the system misinterpreted the source as an imperative form. However, using the infinitive form of the verb in the translation is less clear and raises questions about whether it is a mistranslation or a grammatical error. The distinction between the two lies in the point at which the mistake was made. If the original sentence was understood correctly but the resulting translation was ungrammatical, then it is a grammatical error. On the other hand, if the use of the infinitive form resulted from interpreting ``move'' as an infinitive, it may be considered a mistranslation as well.


\paragraph{Inconsistency:} For marking the ``inconsistency'' errors we decided to the take \textit{minimal} approach. For instance, is the same person is referred to in the translation as both ``Piotr'' and ``Peter'' we would mark only the one that is less frequent. If ``Piotr'' appears \textit{once} in the paragraph, while ``Peter'' is used \textit{twice}, ``Piotr'' would be annotated as being inconsistent. The same strategy was applied for ``register'' errors, such as when both polite and casual forms were acceptable, but the translation used them randomly.

\paragraph{Format:} We did not label ``format'' errors for the \textsc{Sent} and \textsc{Para\_Sent} translations, as we manually corrected the quotation marks during post-processing of the translations. This manual correction was done to ensure that \textsc{Sent} and \textsc{Para\_Sent} could be compared to \textsc{Para} without relying too heavily on simple heuristic (i.e., incorrect usage of the quotation marks).

\paragraph{Omissions:}
\label{app:omissions}
One thing we ought to discuss is the omission issue. Upon examining translations and annotator feedback, we observe that \textsc{Para} occasionally omits details, which are crucial to the storyline. Preliminary investigation indicates that \textsc{Para} translations are more prone to omissions compared to \textsc{Sent} and \textsc{GTr}. Although \textsc{Para\_Sent} appears to mitigate this problem to some extent, it still results in a higher number of omissions than the sentence-level approach while at the same time introducing some repetition issues.

\paragraph{Translators:} The translators in this study were hired on a freelancing platform, Upwork. All were highly proficient in the source language and most of them were native speakers of the target language. Only one translator reported familiarity with the book, which translation she evaluated. All translators were instructed to evaluate each paragraph in isolation without relying on any prior knowledge about the book. Details about the translators are reported in \autoref{tab:annotators_details}.

\begin{table*}[]
\centering
\resizebox{1.0\textwidth}{!}{
    \begin{tabular}%
          {>{\raggedright\arraybackslash}p{3.5cm}%
           >{\raggedleft\arraybackslash}p{1.5cm}%
           >{\raggedleft\arraybackslash}p{1.5cm}%
           >{\raggedleft\arraybackslash}p{1.5cm}%
           >{\raggedleft\arraybackslash}p{1.5cm}%
           >{\raggedleft\arraybackslash}p{1.5cm}%
           >{\raggedleft\arraybackslash}p{1.5cm}%
          }
        \\\toprule
        \textsc{Language   Pair}  & \textsc{Sent}  & \textsc{Para}  & \textsc{Para\_Sent} & \textsc{Para}  & \textsc{GTr}   & \textsc{Para}  \\
        \midrule
        \textit{Russian - English}  & 0            & 10    & 5        & 5     & 4   & 6     \\
        \textit{Chinese - English}  & 2            & 8     & 4        & 6     & 3   & 7     \\
        \textit{Polish - English}   & 4            &  6    & 4        & 6     & 1   & 9     \\
        \textit{French-English}     & 5            & 5     & 4        & 6     & 2   & 8     \\
        \textit{Japanese-English}   & 1            & 9     & 2        & 8     & 1   & 9     \\
        \textit{German-English}     & 5            & 5    & 3         & 7     & 5   & 5              \\

        \cmidrule{2-7}
        \textsc{Total} & 17	& \cellcolor{purple!30}\textbf{43} & 22 &	\cellcolor{purple!30}\textbf{38} & 16 & \cellcolor{purple!30}\textbf{44} \\
        \noalign{\vskip 2mm}
        \textsc{Percentage} & 28.33\%	& \cellcolor{purple!30}\textbf{71.67\%} & 36.67\% &	\cellcolor{purple!30}\textbf{63.33\%} & 26.67\% & \cellcolor{purple!30}\textbf{73.33\%} \\
        \cmidrule{2-7}
        
        \noalign{\vskip 2mm}
        \textit{German-Japanese}  & 6     & 4     & 3                   & 7     & 1    & 9     \\
        \textit{Russian-Japanese} & 4     & 6     & 4                   & 6     & 2    & 8     \\
        \textit{French-Japanese}  & 2     & 8     & 1                   & 9     & 0    & 10    \\
        \textit{Polish-Japanese}  & 2     & 8     & 4                   & 6     & 0    & 10    \\
        \textit{English-Japanese} & 3     & 7     & 2                   & 8     & 1    & 9     \\
        \textit{Chinese-Japanese} & 4     & 6     & 4                   & 6     & 0    & 10    \\

        \cmidrule{2-7}
        \textsc{Total} & 21	& \cellcolor{purple!30}\textbf{39} & 18 & \cellcolor{purple!30}\textbf{42} & 4 & \cellcolor{purple!30}\textbf{56} \\
        \noalign{\vskip 2mm}
        \textsc{Percentage} & 35\%	& \cellcolor{purple!30}\textbf{65\%} & 30\% &	\cellcolor{purple!30}\textbf{70\%} & 6.67\% & \cellcolor{purple!30}\textbf{93.33\%} \\
        \cmidrule{2-7}
        \noalign{\vskip 2mm}
        \textit{English-Polish}   & 0              & 10    & 3                   & 7     & 4              & 6     \\
        \textit{Japanese-Polish}  & 3              & 7     & 5                   & 5     & 1              & 9     \\
        \textit{French-Polish}    & 4              & 6     & 4                   & 6     & 2              & 8     \\
        \textit{Czech-Polish}     & 3              & 7     & 2                   & 8     & 0              & 10    \\
        \textit{Russian-Polish}   & 1              & 9     & 4                   & 6     & 3              & 7     \\
        \textit{German-Polish}    & 3              & 7     & 3                   & 7     & 1              & 9     \\

        \cmidrule{2-7}
        \textsc{Total} & 14 & \cellcolor{purple!30}\textbf{46} & 21 & \cellcolor{purple!30}\textbf{39} & 11 & \cellcolor{purple!30}\textbf{49} \\
        \noalign{\vskip 2mm}

        \textsc{Percentage} & 23.33\%	& \cellcolor{purple!30}\textbf{76.67\%} & 35\% &	\cellcolor{purple!30}\textbf{65\%} & 18.33\% & \cellcolor{purple!30}\textbf{81.67\%} \\
        \cmidrule{2-7}
        \noalign{\vskip 2mm}
        \midrule
        \textbf{TOTAL}            & 52    & \cellcolor{purple}\textbf{\textcolor{white}{128}}   & 61         & \cellcolor{purple}\textbf{\textcolor{white}{119}}   & 31    & \cellcolor{purple}\textbf{\textcolor{white}{149}}   \\
        \noalign{\vskip 2mm}
        \textbf{PERCENTAGE}       & 28.89\% & \cellcolor{purple}\textbf{\textcolor{white}{71.11\%}} & 33.89\%      & \cellcolor{purple}\textbf{\textcolor{white}{66.11\%}} & 17.22\% & \cellcolor{purple}\textbf{\textcolor{white}{82.78\%}} \\
        \bottomrule
    \end{tabular}}
\caption{The number of votes for \textsc{Sent} vs \textsc{Para}, \textsc{Para\_Sent} vs \textsc{Para}, and \textsc{GTr} vs \textsc{Para} in human evaluation by the language pair.}
\label{table:vote_count_by_lang_pair_humeval}
\end{table*}

\begin{table*}[t]
\centering
\resizebox{\textwidth}{!}{%
\begin{tabular}{lp{12cm}lcccc}
\hline
\noalign{\vskip 1mm}
\textsc{Type} & \textsc{Description} & \textsc{Trg Lang} & \textsc{Para} & \textsc{Sent} & \textsc{Para\_Sent} & \textsc{GTr} \\ 
\noalign{\vskip 1mm}
\hline
\noalign{\vskip 1mm}
\multirow{3}{*}{\textsc{Context (sentence)}} & \multirow{3}{12cm}{A mistranslation that results most likely from lack of ``understanding'' the sentence-level context (e.g., translating ``guide'' as ``doradca,'' or ``adviser'' instead of ``przewodnik,'' or ``guide''). This can include translating a word or a phrase into one that is semantically related but does not convey the intended meaning, or translation which appear to be an outcome of translating a word semantically related to the source word, instead of the source word itself.} & Japanese & 114 & 118 & 107 & 158 \\ \noalign{\vskip 1mm}
 &  & Polish & 64 & 67 & 49 & 82 \\ \noalign{\vskip 1mm}
 &  & English & 30 & 36 & 44 & 59 \\ 
 \noalign{\vskip 14mm}
 \noalign{\vskip 3mm}
\multirow{3}{*}{\textsc{Context (paragraph)}} & \multirow{3}{12cm}{A mistranslation that results from lack of a beyond-sentence context. This include issues such as polysemy, employment of correct pronouns, or translating elliptical expressions.} & Japanese & 6 & 36 & 6 & 38 \\\noalign{\vskip 1mm}
 &  & Polish & 13 & 51 & 15 & 59 \\ \noalign{\vskip 1mm}
 &  & English & 2 & 25 & 0 & 48 \\ \noalign{\vskip 1mm}
 \noalign{\vskip 3mm}
\multirow{3}{*}{\textsc{Minor Issue}} & \multirow{3}{12cm}{A minor issue which does not significantly affect the text and can be disputable, such as translating ``barked'' as ``howl.''} & Japanese & 34 & 25 & 26 & 16 \\ \noalign{\vskip 1mm}
 &  & Polish & 33 & 26 & 16 & 13 \\ \noalign{\vskip 1mm}
 &  & English & 18 & 11 & 12 & 9 \\ \noalign{\vskip 1mm}
\noalign{\vskip 3mm}
\multirow{3}{*}{\textsc{Surface Similarity}} & \multirow{3}{12cm}{A translation by word which is similar to the correct translation on the surface level, but has a different meaning (e.g., ``Wilczak,'' a Polish surname, instead of ``wilczarz,'' a ``wolfhound'').} & Japanese & 8 & 6 & 7 & 2 \\ \noalign{\vskip 1mm}
 &  & Polish & 14 & 13 & 16 & 5 \\ \noalign{\vskip 1mm}
 &  & English & 5 & 5 & 6 & 2 \\ \noalign{\vskip 1mm}
\noalign{\vskip 3mm}
\multirow{3}{*}{\textsc{Word-by-Word}} & \multirow{3}{12cm}{A translation of longer phrase which is overly literal resulting in confusing and incorrect translation.} & Japanese & 15 & 52 & 34 & 84 \\ \noalign{\vskip 1mm}
 &  & Polish & 17 & 23 & 18 & 33 \\ \noalign{\vskip 1mm}
 &  & English & 7 & 13 & 5 & 20 \\ \noalign{\vskip 1mm}
 \noalign{\vskip 3mm}
\multirow{3}{*}{\textsc{Unrelated Word}} & \multirow{3}{12cm}{A translation with unrelated word such as ``klnie'' (``swear'') instead of ``zapuka'' (``knock'') where no apparent semantic relation could be found.} & Japanese & 3 & 2 & 5 & 4 \\ \noalign{\vskip 1mm}
 &  & Polish & 5 & 14 & 10 & 12 \\ \noalign{\vskip 1mm}
 &  & English & 1 & 3 & 1 & 2 \\ \noalign{\vskip 1mm}
 \noalign{\vskip 3mm}
\multirow{3}{*}{\textsc{Subject Changed}} & \multirow{3}{12cm}{Change of subject, which occurs mostly due to merging two sentences with two distinctive subjects where all states and/or actions are then assigned to one of them.} & Japanese & 5 & 2 & 2 & 0 \\ \noalign{\vskip 1mm}
 &  & Polish & 6 & 0 & 5 & 3 \\ \noalign{\vskip 1mm}
 &  & English & 7 & 2 & 5 & 1 \\ \noalign{\vskip 1mm}
 \noalign{\vskip 3mm}
\multirow{3}{*}{\textsc{Factuality}} & \multirow{3}{12cm}{A translation that results in change in factuality, such as translating affirmative sentence as negation or translating word by its antonym.} & Japanese & 4 & 11 & 5 & 7 \\ \noalign{\vskip 1mm}
 &  & Polish & 0 & 2 & 1 & 3 \\ \noalign{\vskip 1mm}
 &  & English & 1 & 2 & 1 & 1 \\ \noalign{\vskip 1mm}
 \noalign{\vskip 3mm}
\multirow{3}{*}{\textsc{Non-word}} & \multirow{3}{12cm}{A translation by a non-existent (made up) word. Some examples include skillfully constructed words like \begin{CJK}{UTF8}{min}火炎棒\end{CJK} which was generated instead of a ``torch.'' While this word does not exist in Japanese (or Chinese) it follows the compositionality rules of these languages and is fully intelligible to a native speaker (\begin{CJK}{UTF8}{min}火炎\end{CJK} ``fire'' and \begin{CJK}{UTF8}{min}棒\end{CJK} ``stick.'')} & Japanese & 1 & 2 & 2 & 0 \\ \noalign{\vskip 1mm}
 &  & Polish & 6 & 8 & 9 & 3 \\ \noalign{\vskip 1mm}
  &  & English & 0 & 0 & 0 & 0 \\ \noalign{\vskip 1mm}
 \noalign{\vskip 10mm}
 \noalign{\vskip 1mm}
\multirow{3}{*}{\textsc{mood}} & \multirow{3}{12cm}{Change in the grammatical mood with regard to the source text. Note that the sentence here is \textit{still} grammatically correct but does not reflect the meaning intended by the author.} & Japanese & 4 & 9 & 1 & 3 \\ \noalign{\vskip 1mm}
 &  & Polish & 1 & 3 & 4 & 2 \\ \noalign{\vskip 1mm}
  &  & English & 0 & 0 & 0 & 0 \\ \noalign{\vskip 1mm}
 \noalign{\vskip 5mm}
\multirow{3}{*}{\textsc{Unnecessary Translation}} & \multirow{3}{12cm}{A translation of text which should be left untranslated such as some proper names.}   & Japanese & 0 & 0 & 0 & 0 \\ \noalign{\vskip 1mm}
& & Polish & 0 & 3 & 0 & 2 \\ \noalign{\vskip 1mm}
 &  & English & 1 & 1 & 1 & 1 \\ \noalign{\vskip 1mm}
 \noalign{\vskip 3mm}
\multirow{2}{*}{\textsc{Language Mismatch}} & \multirow{2}{12cm}{A translation into a language different than the target language (e.g., Chinese instead of Japanese). Note that leaving the word in the source language classifies as an ``untranslated'' error.} & Japanese & 2 & 3 & 3 & 2 \\ \noalign{\vskip 1mm}
 & & Polish & 2 & 0 & 2 & 0 \\ \noalign{\vskip 1mm} 
  &  & English & 0 & 0 & 0 & 0 \\ \noalign{\vskip 1mm}
 \noalign{\vskip 5mm}
\multirow{3}{*}{\textsc{Number/Time}} & \multirow{3}{12cm}{A translation which changes number or time expression, such as translating 1h15min as 1h30min. Note that these rarely affect the overall meaning of the text. We have not observe cases where this would be a critical issue.} & Japanese & 3 & 2 & 4 & 3 \\ \noalign{\vskip 1mm}
 &  & Polish & 0 & 0 & 0 & 0 \\ \noalign{\vskip 1mm}
 &  & English & 5 & 2 & 1 & 3 \\ \noalign{\vskip 1mm}
 \noalign{\vskip 6mm}
\multirow{3}{*}{\textsc{Pivot Translation} (Czech)} & \multirow{3}{12cm}{A mistranslation that stems from pivoting on English (annotated for cs-pl language pair).} & Polish & 0 & 0 & 0 & 43 \\ \noalign{\vskip 1mm}
\noalign{\vskip 5mm}
\multirow{3}{*}{\textsc{other}} & \multirow{3}{12cm}{Other issues which do not fit into any of the above.} & Japanese & 24 & 26 & 27 & 17 \\ \noalign{\vskip 1mm}
 &  & Polish & 9 & 14 & 10 & 13 \\ \noalign{\vskip 1mm}
 &  & English & 10 & 4 & 5 & 4 \\ \noalign{\vskip 3mm}
 \hline
 \noalign{\vskip 1mm}
 & & \textsc{Total} (\textit{Japanese})   & 223 & 294 & 229 & 334 \\ \noalign{\vskip 2mm}
 & & \textsc{Total} (\textit{Polish})  & 170 & 224 & 155 & 273 \\ \noalign{\vskip 2mm}
 & & \textsc{Total} (\textit{English})  & 87 & 104 & 81 & 150 \\ \noalign{\vskip 2mm}
 \rowcolor{violet!20} & & \textsc{Total} (\textit{All})  & \textbf{480} & \textbf{622} & \textbf{465} & \textbf{757} \\ 
 \noalign{\vskip 1mm}
 \hline
\end{tabular}
}
\caption{Classification of mistranslation errors for each system grouped by the target language.}
\label{table:mistrans_type_class_by_trg}
\end{table*}

\section{Pivot Pilot}
\label{app:pivot_pilot}

\autoref{tab:pivot_results} shows the results of the pilot study evaluating the effect of pivoting on English on the quality of paragraph-level translations. 
 The evaluation was done by the first author on \textit{all} 20 passages for every language pair that did not include translation from or into English, as these do not require any pivoting. A total number of 200 pairs was evaluated employing simple preference judgments.

 During the \textsc{Para\_Pivot} translation process, the model utilized both the source text and its corresponding English translation (\texttt{text-davinci-003}, top-p=1.0, temp=0.3). This approach has the potential to mitigate the limitations associated with pivoting translations, where some information may be lost. For example, both Czech and Polish encode the gender information in the past tense form of the verb. English does not, so this information is lost and will most likely result in an erroneous translation. Indeed, we notice that adding the source text helps the model to overcome this shortcoming, however, we do not observe a clear gain from using English as a pivot language. Consider the following example:

\begin{exe}
\small
    \ex\label{} Taťka mě zase zmlátil. Byl hrozně naštvanej, protože mamka řvala. On mě vždycky mlátí, když ona řve. Že prej jsem pořádně neudělala nádobí. Ale já vím, že jsem je udělala dobře. I ten hrnec jsem vydrhla pořádně. A ještě jsem to všechno utřela před koupáním. Ale možná jsem nevyždímala hadr.
    \begin{flushright}
    \scriptsize
    \vspace{-0.2cm}
    ―\textsc{Czech Source} (from \textit{Crows})

    \end{flushright}
        \begin{xlist}
            \ex  Tata mnie znowu zbił. \textcolor{teal}{\textbf{Był wściekły}}, bo mama krzyczała. On zawsze mnie bije, kiedy ona krzyczy. Że niby nie \textcolor{teal}{\textbf{umyłam}} dobrze naczyń. Ale ja wiem, że \textcolor{teal}{\textbf{umyłam}} je dobrze. I garnek też dokładnie \textcolor{teal}{\textbf{wypolerowałam}}. I jeszcze wszystko \textcolor{red}{\textbf{\textbf{wytrzepałam}}} przed kąpielą. Ale może nie \textcolor{red}{\textbf{\textbf{wyciśniłam}}} ręcznika.
            \begin{flushright}
            \scriptsize
            ―\textsc{GPT-3.5 Para (Polish)}
            \end{flushright}
            \vspace{0.15cm}
            \ex Tata mnie znów uderzył. \textcolor{teal}{\textbf{Był wściekły}}, bo mama krzyczała. On zawsze mnie bije, kiedy ona krzyczy. Że niby nie \textcolor{teal}{\textbf{umyłam}} dobrze naczyń. Ale ja wiem, że \textcolor{teal}{\textbf{umyłam}} je dobrze. Nawet garnek dokładnie \textcolor{red}{\textbf{wytrzepałam}}. I jeszcze wszystko przed kąpielą \textcolor{red}{\textbf{wytrzepałam}}. Ale może nie \textcolor{red}{\textbf{wyżągnęłam}} mopa.
            \begin{flushright}
            \scriptsize
            ―\textsc{GPT-3.5 Para\_Pivot (Polish)}
            \end{flushright}
    \end{xlist}
\end{exe}

In each instance, the emphasized verbs could potentially be mistranslated when translated through English as the pivot language, as the speaker's gender information would be lost. For instance, the past tense verb ``washed'' remains unchanged in English regardless of the gender of the speaker, with such details encoded only in the source (\textit{Czech}) and target (\textit{Polish}) languages. In this case, all verbs have been translated accurately with respect to grammatical gender, implying that incorporating the source language into the pivot pipeline does indeed improve the translation. However, \textsc{Para\_Pivot} still selects less suitable verbs (highlighted in red) resulting in slightly more errors in this particular paragraph.
 
 The only pair where pivoting seems to help is \textit{pl-ja}. While it is unclear why this happens, it is possible that this outcome is due to the specifics of the Polish novel employed for the translation. \textit{Sword of Destiny} by Andrzej Sapkowski uses a very distinct language with many archaic expressions. It is possible that translating into English, a language the \textsc{GPT} models were trained on, helps the model deal with these difficult phrases. 

 Since we do not observe any apparent gains from performing the translation via English as a pivot language (\textit{p}=0.62, 95\% [0.448, 0.591]) and doing so reduces the number of examples one can fit into the prompt, we continue our experiments with a direct translation.

 \begin{table*}
    \centering
    \begin{tabular}{c c c c}
        \toprule
        \textsc{Source} & \textsc{Target} & \textsc{Para} & \textsc{Para\_Pivot} \\
        \midrule
        \textsc{Czech} & \textsc{Polish} & 11 & 9 \\
        \noalign{\vskip 2mm}
        \textsc{German} & \textsc{Japanese} & 13 & 7 \\
        \noalign{\vskip 2mm}
        \textsc{German}& \textsc{Polish} & 12 & 8 \\
        \noalign{\vskip 2mm}
        \textsc{French} & \textsc{Japanese} & 9 & 11 \\
        \noalign{\vskip 2mm}
        \textsc{French} & \textsc{Polish} & 11 & 9 \\
        \noalign{\vskip 2mm}
        \textsc{Japanese} & \textsc{Polish} & 10 & 10 \\
        \noalign{\vskip 2mm}
        \textsc{Polish} & \textsc{Japanese} & 3 & 17 \\
        \noalign{\vskip 2mm}
        \textsc{Russian} & \textsc{Japanese} & 10 & 10 \\
        \noalign{\vskip 2mm}
        \textsc{Russian} & \textsc{Polish} & 8 & 12 \\
        \noalign{\vskip 2mm}
        \textsc{Chinese} & \textsc{Japanese} & 9 & 11 \\
        \noalign{\vskip 2mm}
        \cmidrule{3-4}
        & \textsc{Total} & \emph{96} & \emph{104} \\
        \bottomrule
    \end{tabular}
    \caption{The results of pairwise comparison for the paragraph-level translations with (\textsc{Para\_Pivot}) and without (\textsc{Para}) English as a pivot language.}
    \label{tab:pivot_results}
\end{table*}

\section{Automatic Metrics}
\label{app:stats}

\begin{table*}[hbtp!]
\centering
\resizebox{0.8\textwidth}{!}{%
\begin{tabular}{lcccc}
\toprule
\textsc{Metric}  & \textsc{Acc} & $\tau$ & \textsc{Acc} (\textit{conf}) & $\tau$ (\textit{conf}) \\ 
\midrule
\textsc{Comet}      & 67.41\%  & 0.348         & 72.78\%              & 0.456  \\
\textsc{Comet-QE}   & 64.44\%  & 0.289         & 70.64\%              & 0.413  \\
\textsc{Bleurt}     & 61.30\%  & 0.226          & 66.36\%              & 0.327 \\
\textsc{BARTScore}  & 58.52\%  & 0.170         & 63.91\%              & 0.278  \\
\bottomrule
\end{tabular}
}
\caption{Correlation of automatic metrics with human judgments from our human evaluation. We evaluate the metrics performance on \textit{all} human judgments as well as on the \textit{subset} of judgments were the translator indicated that the chosen translation was visibly better (\textit{conf}). We report both the percentage of agreement (\textsc{Acc}) and Kendall's Tau ($\tau$)}
\label{tab:corr_w_humjudgments}
\end{table*}

\paragraph{Correlation with Human Judgements:} We investigate the correlation of automatic metrics with human judgements in our evaluation. We consider (1) all the judgments, as well as (2) a subset of all judgments where the annotator stated that they were sure that one translation is \textit{clearly} better than the other. We compute both \textit{accuracy} (i.e., the percentage of cases where the metric agrees with human judgment), and a correlation coefficient Kendall's Tau which is defined as follows:

\begin{center}
\footnotesize
\[\tau = \frac{\mbox{Concordant} - \mbox{Discordant}}{\mbox{Concordant} + \mbox{Discordant}}\]
\end{center}

\autoref{tab:corr_w_humjudgments} shows the correlation of automatic metrics with the human judgments obtained in this study. \textsc{Comet} exhibits the highest agreement with human judgments both in terms of the \textit{accuracy} (64.04\% for all data, 72.78\% for confident votes only) and Kendall's Tau (0.341 for all data, 0.456 for confident votes only).

\paragraph{Statistical Analysis:} We employ the linear-mixed effect models \cite{Baayen2008_lme} to analyze the scores produced by automatic metrics. We fitted the model in \texttt{R} using the \texttt{lme4} package \cite{lme4}; the \textit{p}-values were obtained with the \texttt{LmerTest} package \cite{lmertest}. Linear-mixed effects models contain both \textit{fixed-effects} and \textit{random-effects} (random \textit{intercept} and/or \textit{slope}). The fixed effect here is the translation setup (\textsc{Para}, \textsc{Sent}, \textsc{Para\_Sent}, \textsc{GTr}) with the source paragraph being coded as the random effect. We inspect the residual plots to ensure that the variance across the fitted range is relatively constant. The results from the fitted model are presented in \autoref{tab:bleurt_stats} (\textsc{Bleurt}), \autoref{tab:comet_stats} (\textsc{Comet}), \autoref{tab:comet_qe_stats} (\textsc{Comet-QE}), and \autoref{tab:bertscore_stats} (\textsc{BertScore}).

 We further perform a post hoc analysis using the \texttt{emmeans} package \cite{emmeans} to obtain \textit{p}-values for the pairwise comparison. The results of the post hoc analysis are presented in \autoref{tab:bleurt_stats_posthoc} (\textsc{Bleurt}), \autoref{tab:comet_stats_posthoc} (\textsc{Comet}), \autoref{tab:comet_qe_stats_posthoc} (\textsc{Comet-QE}), and \autoref{tab:bertscore_stats_posthoc} (\textsc{BertScore}).


\begin{table*}
\centering
\begin{tabular}{lccc}
\toprule
 & \multicolumn{3}{c}{\textsc{Bleurt}} \\
\cmidrule(lr){2-4}
Predictors & Estimates & CI & \textit{p}-value \\
\midrule
(Intercept) & 0.48 & 0.47--0.50 & \textbf{$<$0.001} \\
\textsc{Para\_Sent} & -0.00 & -0.01--0.00 & 0.130 \\
\textsc{Sent} & -0.02 & -0.02--(-0.01) & \textbf{$<$0.001} \\
\textsc{GTr} & -0.04 & -0.05--(-0.04) & \textbf{$<$0.001} \\
\bottomrule
\end{tabular}
\caption{Results of linear-mixed effects models analysis for \textsc{Bleurt} scores.}
\label{tab:bleurt_stats}
\end{table*}

\begin{table*}
\centering
\begin{tabular}{lccccc}
\toprule
 &  & & \textsc{Bleurt} &  &  \\
\midrule
Contrast & Estimate & SE & df & \textit{t}-ratio & \textit{p}-value \\
\midrule
\textsc{Para} - \textsc{Para\_Sent} & 0.00477 & 0.00315 & 1074 & 1.515 & 0.780 \\
\textsc{Para} - \textsc{Sent} & 0.01641 & 0.00315 & 1074 & 5.215 & \textbf{$<$0.001} \\
\textsc{Para} - \textsc{GTr} & 0.04155 & 0.00315 & 1074 & 13.205 & \textbf{$<$0.001} \\
\textsc{Para\_Sent} - \textsc{Sent} & 0.01164 & 0.00315 & 1074 & 3.700 & \textbf{0.001} \\
\textsc{Para\_Sent} - \textsc{GTr} & 0.03678 & 0.00315 & 1074 & 11.690 & \textbf{$<$0.001} \\
\textsc{Sent} - \textsc{GTr} & 0.02514 & 0.00315 & 1074 & 7.990 & \textbf{$<$0.001} \\
\bottomrule
\end{tabular}
\caption{Result of post hoc analysis with \textit{emmeans} package for \textsc{Bleurt}.}
\label{tab:bleurt_stats_posthoc}
\end{table*}


\begin{table*}
\centering
\begin{tabular}{lccc}
\toprule
 & \multicolumn{3}{c}{\textsc{Comet}} \\
\cmidrule(lr){2-4}
Predictors & Estimates & CI & \textit{p}-value \\
\midrule
(Intercept) & 0.79 & 0.77--0.80 & \textbf{$<$0.001} \\
\textsc{Para\_Sent} & -0.01 & -0.01--(-0.00) & \textbf{0.019} \\
\textsc{Sent} & -0.01 & -0.01--(-0.00) & \textbf{0.004} \\
\textsc{GTr} & -0.05 & -0.05--(-0.05) & \textbf{$<$0.001} \\
\bottomrule
\end{tabular}
\caption{Results of linear-mixed effects models analysis for \textsc{Comet} scores.}
\label{tab:comet_stats}
\end{table*}

\begin{table*}[h]
  \centering
  \begin{tabular}{@{}lrrrrr@{}}
    \toprule
     &  & & \textsc{Comet} &  &  \\
    \midrule
    Contrast & Estimate & SE & df & \textit{t}-ratio & \textit{p}-value \\
    \midrule
    \textsc{Para} - \textsc{Para\_Sent} & 0.00563 & 0.00239 & 1074 & 2.356 & 0.112 \\
    \textsc{Para} - \textsc{Sent} & 0.00691 & 0.00239 & 1074 & 2.893 & \textbf{0.023} \\
    \textsc{Para} - \textsc{GTr} & 0.04998 & 0.00239 & 1074 & 20.928 & \textbf{<.001} \\
    \textsc{Para\_Sent} - \textsc{Sent} & 0.00128 & 0.00239 & 1074 & 0.536 & 1.000 \\
    \textsc{Para\_Sent} - \textsc{GTr} & 0.04435 & 0.00239 & 1074 & 18.571 & \textbf{<.001} \\
    \textsc{Sent} - \textsc{GTr} & 0.04307 & 0.00239 & 1074 & 18.035 & \textbf{<.001} \\
    \bottomrule
  \end{tabular}
  \caption{Result of post hoc analysis with \textit{emmeans} package for \textsc{Comet}.}
  \label{tab:comet_stats_posthoc}
\end{table*}


\begin{table*}[h]
  \centering
  \begin{tabular}{@{}lccc@{}}
  \toprule
 & \multicolumn{3}{c}{\textsc{Comet-QE}} \\
    \cmidrule(lr){2-4}
    Predictors & Estimates & CI & \textit{p}-value \\
    \midrule
    (Intercept) & -0.04 & -0.06 – -0.01 & \textbf{0.004} \\
    \textsc{Para\_Sent} & -0.01 & -0.03 – -0.00 & \textbf{0.026} \\
    \textsc{Sent} & -0.02 & -0.04 – -0.01 & \textbf{<0.001} \\
    \textsc{GTr} & -0.12 & -0.13 – -0.11 & \textbf{<0.001} \\
    \bottomrule
  \end{tabular}
  \caption{Results of linear-mixed effects models analysis for \textsc{Comet-QE} scores.}
  \label{tab:comet_qe_stats}
\end{table*}

\begin{table*}[h]
  \centering
  \begin{tabular}{@{}lrrrrr@{}}
    \toprule
     &  & & \textsc{Comet-QE} &  &  \\
    \midrule
    Contrast & Estimate & SE & df & \textit{t}-ratio & \textit{p}-value \\
    \midrule
    \textsc{Para} - \textsc{Para\_Sent} & 0.01464 & 0.00655 & 1074 & 2.235 & 0.154 \\
    \textsc{Para} - \textsc{Sent} & 0.02376 & 0.00655 & 1074 & 3.628 & \textbf{0.002} \\
    \textsc{Para} - \textsc{GTr} & 0.11848 & 0.00655 & 1074 & 18.092 & \textbf{<.001} \\
    \textsc{Para\_Sent} - \textsc{Sent} & 0.00912 & 0.00655 & 1074 & 1.392 & 0.9844 \\
    \textsc{Para\_Sent} - \textsc{GTr} & 0.10384 & 0.00655 & 1074 & 15.857 & \textbf{<.001} \\
    \textsc{Sent} - \textsc{GTr} & 0.09472 & 0.00655 & 1074 & 14.464 & \textbf{<.001} \\
    \bottomrule
  \end{tabular}
  \caption{Result of post hoc analysis with \textit{emmeans} package for \textsc{Comet-QE}.}
  \label{tab:comet_qe_stats_posthoc}
\end{table*}


\begin{table*}[htbp]
  \centering
    \begin{tabular}{lccc}
    \toprule
    & \multicolumn{3}{c}{\textsc{BertScore}} \\
    \cmidrule(lr){2-4}
    Predictors & Estimates & CI & \textit{p}-value \\
    \midrule
    (Intercept) & 0.84  & 0.83--0.85 & \textbf{$<$0.001} \\
    \textsc{Para\_Sent} & -0.00 & -0.00--0.00 & \textbf{0.037} \\
    \textsc{Sent} & -0.00 & -0.00--0.00 & 0.522 \\
    \textsc{GTr} & -0.01 & -0.01--0.01 & \textbf{$<$0.001} \\
    \bottomrule
    \end{tabular}
  \caption{Results of linear-mixed effects models analysis for \textsc{BertScore} scores.}
  \label{tab:bertscore_stats}
\end{table*}

\begin{table*}[htbp]
  \centering
  \begin{tabular}{@{}lrrrrr@{}}
    \toprule
     &  & & \textsc{BertScore} &  &  \\
    \midrule
    Contrast & Estimate & SE & df & \textit{t}-ratio & \textit{p}-value \\
    \midrule
    \textsc{Para} - \textsc{Para\_Sent} & 0.002422 & 0.00116 & 1074 & 2.082 & 0.225 \\
    \textsc{Para} - \textsc{Sent} & 0.000745 & 0.00116 & 1074 & 0.640 & 1.000 \\
    \textsc{Para} - \textsc{GTr} & 0.007508 & 0.00116 & 1074 & 6.454 & \textbf{$<$0.001} \\
    \textsc{Para\_Sent} - \textsc{Sent} & -0.001678 & 0.00116 & 1074 & -1.442 & 0.897 \\
    \textsc{Para\_Sent} - \textsc{GTr} & 0.005086 & 0.00116 & 1074 & 4.372 & \textbf{$<$0.001} \\
    \textsc{Sent} - \textsc{GTr} & 0.006763 & 0.00116 & 1074 & 5.814 & \textbf{$<$0.001} \\
    \bottomrule
    \end{tabular}
    \caption{Result of post hoc analysis with \textit{emmeans} package for \textsc{BertScore}.}
    \label{tab:bertscore_stats_posthoc}
\end{table*}%

\end{document}